\documentclass[10pt,twocolumn,letterpaper]{article}

\usepackage[pagenumbers]{wacv} 

\usepackage{epsfig}
\usepackage{graphicx}
\usepackage{amsmath}
\usepackage{amssymb}

\usepackage[x11names, table]{xcolor}

\usepackage{stmaryrd}

\usepackage{pgfplots}
\pgfplotsset{compat=1.17}

\usepackage{colortbl}
\usepackage{booktabs}

\usepackage{multirow}
\usepackage{rotating}
\usepackage{array}
\newcolumntype{S}{>{\centering\arraybackslash}m{1.5cm}}
\newcolumntype{L}{>{\centering\arraybackslash}m{5cm}}
\usepackage{rotating, 
            makecell}
\usepackage{tabularray}
\usepackage{adjustbox}

\usepackage[T1]{fontenc}

\usepackage{lipsum}

\usepackage{nccmath}

\usepackage{mathtools}

\usepackage{siunitx}

\usepackage{algorithm}
\usepackage{algpseudocode}

\usepackage{pifont}
\newcommand{\xmark}{\ding{51}}%
\newcommand{\xcross}{\ding{55}}%

\makeatletter
\DeclareRobustCommand\onedot{\futurelet\@let@token\@onedot}
\def\@onedot{\ifx\@let@token.\else.\null\fi\xspace}

\def\eg{\emph{e.g}\onedot} 
\def\ie{\emph{i.e}\onedot} 
\def\cf{\emph{cf}\onedot}

\def\etal{\emph{et al}\onedot}
\makeatother

\newcommand{\papertitle}{Towards Fast and
Scalable
Normal Integration using
Continuous
Components\xspace}

\definecolor{wacvblue}{rgb}{0.21,0.49,0.74}
\usepackage[pagebackref,breaklinks,colorlinks,allcolors=wacvblue]{hyperref}

\def\wacvPaperID{456} 
\def\confName{WACV}
\def\confYear{2026}

\title{\papertitle}

\author{Francesco Milano\textsuperscript{1}\hspace{30pt} Jen Jen Chung\textsuperscript{2}\hspace{30pt} Lionel Ott\textsuperscript{1}\hspace{30pt} Roland Siegwart\textsuperscript{1}\\[5pt]${}^1$ETH Zurich\hspace{30pt} ${}^2$The University of Queensland\hspace{30pt}\\[5pt]}

\begin{document}
\maketitle

\begin{abstract}
Surface normal integration is a fundamental problem in computer vision, dealing with the objective of reconstructing a surface from its corresponding normal map. Existing approaches require an iterative global optimization to jointly estimate the depth of each pixel, which scales poorly
to larger normal maps.
In this paper, we
address
this problem by recasting normal integration as the estimation of relative scales of continuous components. By constraining pixels belonging to the same component to jointly vary their scale, we drastically reduce the number of optimization variables. Our framework includes a heuristic to accurately estimate continuous components from the start, a strategy to rebalance optimization terms, and a technique to iteratively merge components to further reduce the size of the problem. Our method achieves state-of-the-art results on
the
standard normal integration
benchmark
in as little as a few seconds and achieves
one-order-of-magnitude
speedup
over pixel-level approaches
on large-resolution normal maps.
\end{abstract}

\section{Introduction\label{sec:introduction}}
The problem of reconstructing a surface from its normal map, known as normal integration, arises naturally in many applications of computer vision,
particularly in
photometric shape
recovery,
such as
shape
from shading~\cite{Horn1975ShapeFromShading} and from polarization~\cite{Atkinson2006RecoverySurfaceOrientation, Ba2020DeepShapeFromPolarization} and photometric stereo~\cite{Woodham1980PhotometricStereo, Ikehata2023SDMUniPS},
in which the objective is to reconstruct a 3D surface from shading information, using estimated normals as an intermediate step.

The mathematical problem
underlying normal estimation has been extensively studied
and
several
recent methods have been
proposed that achieve sub-millimeter accuracy on object-level normal maps~\cite{Cao2022BiNI, Kim2024DiscontinuityPreserving, Milano2025DiscontinuityAwareNormalIntegration}.
While accurate,
state-of-the-art
approaches are 
predominantly based on a costly 
global optimization that jointly estimates the depth value at each pixel, with the number of optimization variables and constraints thereby scaling
with the number of pixels. 
Additionally,
multiple optimization steps with iterative reweighting
are
necessary to capture surface discontinuities, which
cannot
be correctly estimated
a priori
in the general
case
and would otherwise
cause
the reconstruction to be suboptimal~\cite{Cao2022BiNI}.
As a consequence, these methods have runtimes in the order of 
minutes for relatively low 
scales
(typically in the order of $10^{4}$ valid pixels) and
up to hours for high
resolutions and scales.

To address the above limitation,
we observe
that naturally occurring surfaces are often made of several smooth regions. If each such
region was independently reconstructed, obtaining a single, global surface would reduce to optimally aligning each region to one another. Thus, we propose to recast normal integration into the estimation of the relative scales of continuous surface components. We
introduce
a simple and effective heuristic, based on normal similarity, to identify continuous components, and
independently reconstruct each of them 
using the state-of-the-art formulation of~\cite{Milano2025DiscontinuityAwareNormalIntegration}. We then
jointly
scale the depth
of all pixels in each component by optimizing a single scale parameter for each
component. For this, we
reframe
\cite{Milano2025DiscontinuityAwareNormalIntegration}
and the discontinuity model of BiNI~\cite{Cao2022BiNI} to use relative component scales as optimization 
variables,
and
additionally introduce an outlier reweighting mechanism that rebalances
the optimization terms. Importantly, we find that this reweighting
significantly
speeds up convergence also for previous, pixel-level methods, although 
their scalability remains limited due to the size of their optimization problem.
Through its component-based formulation, our method greatly reduces the number of variables and constraints used in the optimization, 
resulting in a
reduction of one order of magnitude in the execution time for mid-to-high resolution normal maps.
Additionally, our
approach
achieves state-of-the-art
reconstruction accuracy
on the DiLiGenT benchmark~\cite{Shi2016DiLiGenT} for normal integration in as little as a few seconds. 

Thus, our contribution is a framework that recasts normal integration as an estimation of relative scales of continuous components, which
\textit{(i)} achieves state-of-the-art reconstruction accuracy on 
normal integration
benchmarks, and
\textit{(ii)} enables scaling discontinuity-preserving normal integration to larger resolutions with an order of magnitude reduction in the execution
time.
Our code is publicly available\footnote{\url{https://github.com/francescomilano172/normal_integration_continuous_components}.}.
\section{Related work\label{sec:related_work}}
We refer the reader to~\cite{Queau2018NormalIntegrationSurvey, Queau2018VariationMethodsNormalIntegration} for
extensive
surveys
of methods for normal integration.
In the following Section, we focus on the 
most recent,
state-of-the-art approaches.

A key challenge in normal integration is 
that the input normal map might represent a surface with
discontinuities, which naturally arise, for instance due to self-occlusions. Failure to preserve such discontinuities results in overly continuous reconstructions with global distortions~\cite{Zhu2020LeastSquaresSurfaceReconstruction, Cao2021NormalIntegrationInversePlaneFitting}. To address this problem, a number of 
\textit{single-analysis}
methods
have proposed detecting discontinuities as
a
preprocessing
step,
according to different strategies~\cite{Karacali2003ReconstructingDiscontinuous, Wu2006VisibleSurfaceReconstruction, Wang2012DetectingDiscontinuitiesSurfaceReconstruction, Xie2019ARobustDGPBased}. 
These methods, however,
tend not to be robust, since errors in the
initial
discontinuity
detection cannot be later corrected during
the reconstruction process.
Recently, large improvements have been achieved by methods that
estimate discontinuities through an iteratively reweighted optimization problem~\cite{Cao2022BiNI, Karacali2003ReconstructingDiscontinuous, Milano2025DiscontinuityAwareNormalIntegration}.
While accurate, these methods suffer from
long
execution times and scale poorly to larger resolutions, due to the need to jointly
recover a depth value
at all pixels through a single, global optimization problem.

To address this problem, Heep and Zell~\cite{Heep2024AdaptiveScreenSpaceMeshingApproach} have recently proposed a meshing-based approach for normal integration, in which a triangle mesh is precomputed from the input normal map based on estimated curvature and a surface reconstruction is obtained by optimizing the depth of each mesh vertex. While
greatly
reducing
the number of variables and the execution time,
the
single,
fixed
mesh
representation
does not support modeling discontinuities, which is admittedly not 
straightforward and would require 
operations to realign the mesh and adjust its topology~\cite{Heep2024AdaptiveScreenSpaceMeshingApproach}.
In this work, we aim at achieving the best of both discontinuity-preserving 
and computationally efficient
approaches. Our component-based formulation is naturally compatible with state-of-the-art pixel-level methods, but also allows
effectively preserving discontinuities while
significantly improving scalability.
\section{Surface normal integration using continuous components~\label{sec:method}}
An overview of our method and
of
its components is shown in 
\cref{fig:method_overview} and
in the form of pseudocode in~\cref{alg:algorithm_pseudocode} (Supplementary).
\Cref{sec:method_background}
formally introduces the problem 
and reviews state-of-the-art approaches
from a unifying
perspective.
\Cref{sec:method_general_framework} illustrates our general framework and
presents
our reformulation of the
underlying mathematical model. 
\Cref{sec:method_formation_filling_initial_components}
describes our proposed heuristics to form continuous components from an input normal map 
(\cref{fig:method_overview}a), and  
\Cref{sec:method_relative_scale_optimization}
presents our strategy for optimizing their relative scales 
to retrieve a globally optimal reconstruction (\cref{fig:method_overview}b). Finally, \Cref{sec:method_optional_component_merging} describes an optional step to merge multiple components 
(\cref{fig:method_overview}c).

\subsection{Background~\label{sec:method_background}}
Surface normal integration is the problem of reconstructing a
depth
map from an input surface normal map, assuming known camera intrinsic parameters. 
Using
the notation of~\cite{Milano2025DiscontinuityAwareNormalIntegration}, for a generic pair of neighboring pixels $(a, b)$ with pixel coordinates $(u_a, v_a)^\mathsf{T}$ and $(u_b, v_b)^\mathsf{T}$, respectively,
we denote with $\boldsymbol{n_a}=\left(n_{ax}, n_{ay}, n_{az}\right)^\mathsf{T}$ and $\boldsymbol{n_b}=\left(n_{bx}, n_{by}, n_{bz}\right)^\mathsf{T}$ the surface normal vectors at those pixels.
The
values of the depth $z_a$, $z_b$ at these pixels can
then
be modeled through a linear relationship in logarithmic space,
which we refer to as a \emph{model of continuity}:
\begin{equation}
    \tilde{z}_a - \tilde{z}_b - \tilde{\omega}_{b\rightarrow a} = 0,
    \label{eq:generic_model_continuity}
\end{equation}
where $\tilde{z}_a \coloneqq \log(z_a)$, $\tilde{z}_b \coloneqq \log(z_b)$, and
the coefficient
$\tilde{\omega}_{b\rightarrow a}$
assumes
a different
value
depending on the formulation. For instance,
in
BiNI~\cite{Cao2022BiNI}, 
$\tilde{\omega}_{b\rightarrow a}$
has the following form for a pinhole camera of focal lengths $f_x$ and $f_y$ and principal point $(c_x, c_y)$:
\begin{equation}
    \tilde{\omega}_{b\rightarrow a} \coloneqq \frac{\delta_{b\rightarrow a}}{n_{ax}(u_a - c_x) + n_{ay}(v_a - c_y) + n_{az}f},
    \label{eq:model_continuity_bini}
\end{equation}
with $f = f_x,\ \delta_{b\rightarrow a} = \pm n_{ax}$ \emph{i.f.f.} $(u_b, v_b) = (u_a \pm 1, v_a)$ and $f = f_y,\ \delta_{b\rightarrow a} = \pm n_{ay}$ \emph{i.f.f.} $(u_b, v_b) = (u_a, v_a \pm 1)$.

Milano~\etal~\cite{Milano2025DiscontinuityAwareNormalIntegration}
instead
derive
the
following
alternative equation for generic central
cameras:
\begin{equation}
    \tilde{\omega}_{b\rightarrow a} \coloneqq \log\left(\frac{\boldsymbol{n_a}^\mathsf{T}\boldsymbol{\tau_m}\cdot\boldsymbol{n_b}^\mathsf{T} \boldsymbol{\tau_b}}{\boldsymbol{n_a}^\mathsf{T} \boldsymbol{\tau_a}\cdot\boldsymbol{n_b}^\mathsf{T} \boldsymbol{\tau_m}}\right),
    \label{eq:model_continuity_milano}
\end{equation}
where $\boldsymbol{\tau_a}$, $\boldsymbol{\tau_b}$, and $\boldsymbol{\tau_m}$
denote, respectively, the ray direction vectors
at
pixel $a$, pixel $b$, and
at a
subpixel $m$
that interpolates
between $a$ and $b$.

\begin{figure}[t]
    \centering
    \begin{subfigure}[b]{0.48\linewidth}
        \includegraphics[width=\linewidth]{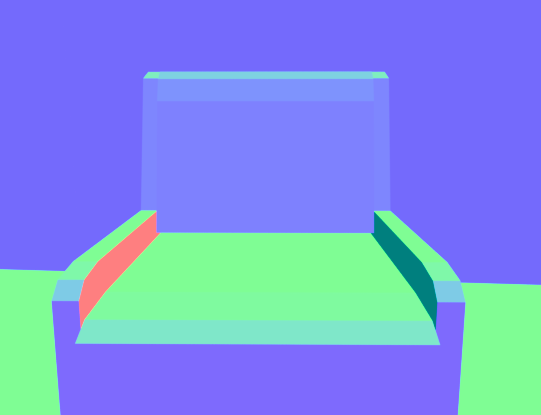}
    \end{subfigure}
    \hfill
    \begin{subfigure}[b]{0.48\linewidth}
        \includegraphics[width=\linewidth]{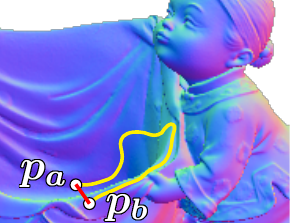}
    \end{subfigure}
    \caption{\textbf{Example of surface discontinuities.} \textit{Left}: The chair in the foreground is separated by a full surface discontinuity from the wall in the background. 
    Infinitely many depth maps can
    describe the input normal map, and solutions of the integration are up to the relative scales
    of
    the chair and the wall. \textit{Right}:
    The
    surface points $\boldsymbol{p_a}$ and $\boldsymbol{p_b}$ are separated by a
    discontinuity along the path in
    red,
    but the yellow path connects them along the surface without discontinuities.
    \textit{Source of object meshes
    }: \cite{MeshOfficePropsLowpoly2021} (left),~\cite{Shi2016DiLiGenT} (right).
    }
    \label{fig:example_discontinuities}
\end{figure}

\begin{figure*}[ht!]
    \centering
    \includegraphics[width=\linewidth]{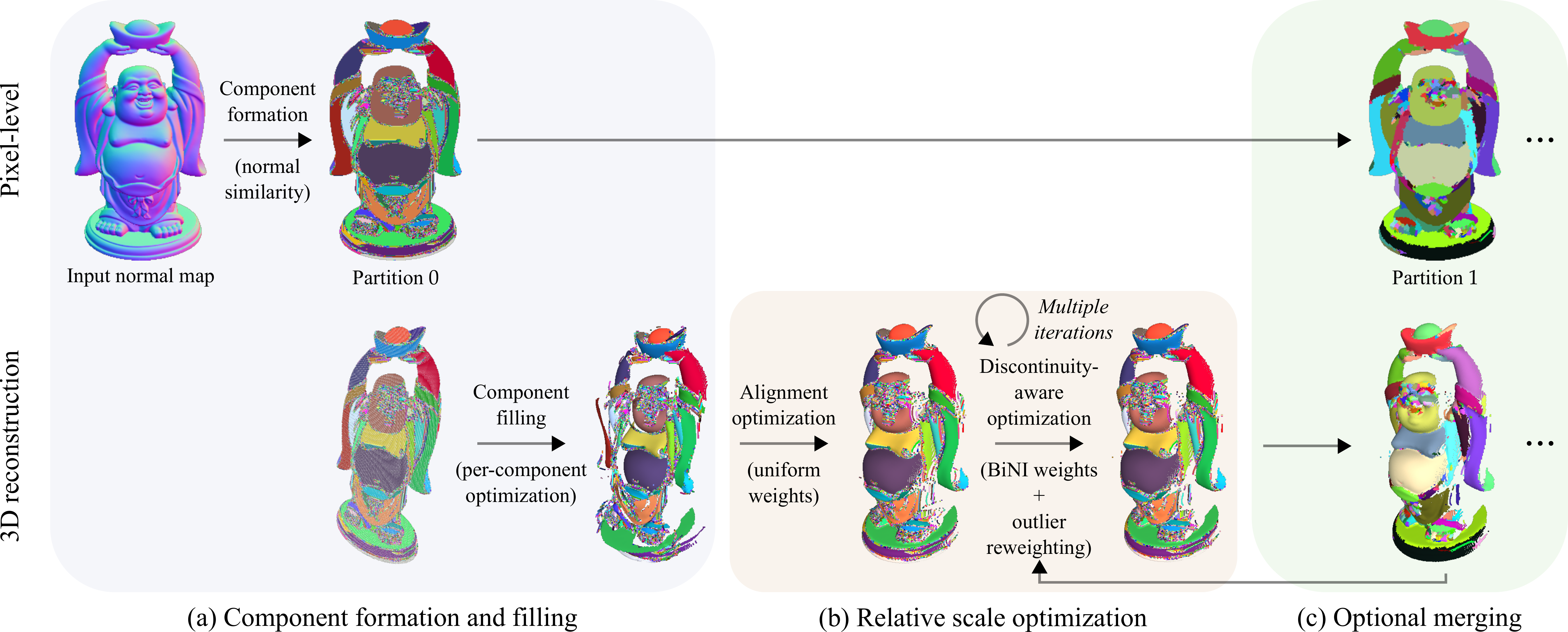}
    \caption{\textbf{Method overview.} We recast normal estimation as the estimation of the relative scale of continuous components. \textit{(a)} Form continuous components based on the similarity of normals at neighboring pixels, independently reconstruct associated surface patches via per-component optimization (\cref{sec:method_formation_filling_initial_components}). \textit{(b)} Align surface patches and recover discontinuities using relative scale optimization based on
    inter-component constraints (\cref{sec:method_relative_scale_optimization}). \textit{(c)} Optionally, during optimization, merge components to 
    further
    reduce complexity (\cref{sec:method_optional_component_merging}).}
    \label{fig:method_overview}
    \vspace{-12pt}
\end{figure*}

Typically, a
global
least-squares
optimization problem is set up to recover the depth map,
jointly enforcing the constraint~\eqref{eq:generic_model_continuity} at all pairs of neighboring pixels.
However,
all
models of continuity 
are 
approximate, since 
the input normal map
only provides a discrete sampling of the surface
normals.
In addition, surface discontinuities
may
be present between two neighboring pixels, which act effectively as unknown constants of the integration problem and
therefore
cannot
be estimated beforehand in the general case. As an example, consider the case of the chair
 in
\cref{fig:example_discontinuities}: while the surface of the chair and that of the wall behind it can be separately reconstructed, the exact depth discontinuity at the boundary 
cannot be estimated from the normals alone, since infinitely many solutions exist. As a result, an unknown residual term $\chi_{b\rightarrow a}$ should be 
introduced in the model of continuity 
to account for discontinuities:
\begin{equation}
    \chi_{b\rightarrow a} \coloneqq \tilde{z}_a - \tilde{z}_b - \tilde{\omega}_{b\rightarrow a}.
    \label{eq:generic_model_continuity_with_discontinuities}
\end{equation}

While in the general case the integration solution is up to these residuals, a number of assumptions can be made about the discontinuities. In particular,
discontinuities often arise from self-occlusions rather than from
fully disconnected surfaces,
for instance,
in normal maps
that capture a single object~\cite{Shi2016DiLiGenT}.
In general,
even in the presence of multiple objects,
while
the residual $\chi_{b\rightarrow a}$
between two neighboring pixels $a$ and $b$
might be significant in magnitude,
there
often
exists
an alternative path along the surface
that connects the two corresponding surface points along a
trajectory with no discontinuities (\cref{fig:example_discontinuities}, right).

State-of-the-art methods
implicitly exploit this fact by 
assigning a lower weight in the optimization objective to
the constraints~\eqref{eq:generic_model_continuity} that, over the course of the optimization, are found to have a large residual magnitude.
The rationale is that
if all constraints were equally weighted, those
associated with discontinuities
would incorrectly
drive
the optimization to assign them a small residual,
causing the reconstruction to converge to an overly continuous surface.
By iteratively adapting the weights of the constraints, these methods instead 
focus
on the more continuous paths along the surface,
and thereby more accurately recover discontinuities, while
still reconstructing a connected surface.
A complementary approach,
recently
proposed by~\cite{Milano2025DiscontinuityAwareNormalIntegration}, consists in dynamically updating the coefficients $\tilde{\omega}_{b\rightarrow a}$ 
in the equations~\eqref{eq:generic_model_continuity} that have large residuals, so as
to explicitly
take into account
the magnitude of the discontinuities. 
This
is achieved through an additive term in the 
logarithm~\eqref{eq:model_continuity_milano}, which
admits a geometrical interpretation.
We refer to the exact mechanism by which 
the above
methods
achieve the reweighting
of the constraints
as a \emph{model of discontinuity}.

The leading model of discontinuity is the bilateral weighting scheme proposed by BiNI~\cite{Cao2022BiNI} which
acts through two different mechanisms.
First,
it relies
on the assumption that
surfaces are
semi-smooth. This implies that, indexing
the two neighboring pixels $b$ and $-b$ on opposite sides of a pixel $a$,
the depth map 
has a discontinuity between at most one of
the sides
$(a, b)$
or
$(a, -b)$.
This is modeled by weighting the constraint~\eqref{eq:generic_model_continuity} between $a$ and $b$ by the factor,
\begin{equation}
    w^{\mathrm{BiNI}}_{b\rightarrow a}=\sigma_k\left(\gamma_{b\rightarrow a}^2\cdot\left((\tilde{z}_{-b} - \tilde{z}_a)^2 - (\tilde{z}_{b} - \tilde{z}_a)^2\right)\right),
\end{equation}
where $\sigma_k(x)$ is the sigmoid function $\left(1+e^{-kx}\right)^{-1}$ with hyperparameter $k$, and $\gamma_{b\rightarrow a}$ is a pixel-specific factor.
If at a given point in the optimization 
it holds that
$(\tilde{z}_{-b} - \tilde{z}_a)^2 \ll (\tilde{z}_{b} - \tilde{z}_a)^2$,
\ie, the surface is estimated to be significantly more continuous on the side $(a, -b)$ than on the side
$(a, b)$,
the value of $w^{\mathrm{BiNI}}_{b\rightarrow a}$ will tend to $0$ and the constraint between $a$ and $b$ will therefore be down-weighted.

The second mechanism is through the factor
$\gamma_{b\rightarrow a}$ itself,
the
square
of which
is
used to scale the term
$w^{\mathrm{BiNI}}_{b\rightarrow a}$, so that the overall weight associated to each constraint~\eqref{eq:generic_model_continuity} is,
\begin{equation}
    W_{b\rightarrow a} = \gamma_{b\rightarrow a}^2\cdot w^{\mathrm{BiNI}}_{b\rightarrow a}.
    \label{eq:total_weight_bini_model_of_discontinuity}
\end{equation}
As noted by~\cite{Milano2025DiscontinuityAwareNormalIntegration},
the factor $\gamma_{b\rightarrow a}$ can be expressed as
\begin{equation}
    \gamma_{b\rightarrow a} = f\cdot \boldsymbol{n_a}^\mathsf{T}\boldsymbol{\tau_a},
\end{equation}
where $f$ is the focal length of the camera,
and is crucial to the optimization convergence, which it affects
through~\eqref{eq:total_weight_bini_model_of_discontinuity}
by
introducing a
multiplicative
factor
$(\boldsymbol{n_a}^\mathsf{T}\boldsymbol{\tau_a})^2$. Since the quantity $\boldsymbol{n_a}^\mathsf{T}\boldsymbol{\tau_a}$ tends to $0$ for pixels $a$ that lie close to an object boundary~\cite{Marr1977AnalysisOccludingContour}, this weighting effectively assigns lower confidence to
constraints involving
pixels that are likely to lie
next
to a depth discontinuity.

These approaches
enforce all constraints in a
global
optimization problem
of the form $(\mathbf{A}\mathbf{\tilde{z}}-\mathbf{b})^\mathsf{T}\mathbf{W}(\mathbf{A}\mathbf{\tilde{z}}-\mathbf{b})$, where
$\mathbf{W}\coloneqq\mathrm{diag}(W_{b\rightarrow a})$.
The optimization variable $\tilde{\mathbf{z}}$ represents the log-depth at each pixel and has 
dimensionality equal to the number of pixels.
Similarly,
the
design
matrix
$\mathbf{A}$,
while sparse,
scales 
with
the number of constraints
and
the number of pixels.
Thus,
the
overall
complexity of the
optimization scales poorly
as the input dimensionality increases,
quickly becoming infeasible for medium-to-large resolution normal maps.
In the next Sections, we present our
method for effective and scalable discontinuity-aware normal integration using continuous components.

\subsection{General framework~\label{sec:method_general_framework}}
Our proposed framework draws inspiration from the
observation that surfaces
typically consist of
large
regions that
are locally
smooth, and
therefore
contain no surface discontinuities.
As a consequence, if each of these surface ``patches'' -- which we refer to as \emph{continuous components} -- was independently described using the model of continuity~\eqref{eq:generic_model_continuity_with_discontinuities}, the residuals $\chi_{b\rightarrow a}$ within the component would all be close to $0$ in
magnitude. As a result,
each of the
component-level
reconstructions,
taken separately,
would
approximate
the corresponding surface
patch
with high accuracy. Obtaining a global reconstruction of the full surface would then reduce to estimating the discontinuities between each pair of neighboring components, which as detailed in the previous Section, could be achieved through a \emph{model of discontinuity}.
Crucially,
since these components separately and accurately describe a fixed surface patch, estimating the discontinuities between them can be reframed as adjusting the \emph{scale} of each component relative to one another (\cref{fig:method_overview}b). For instance, 
in \cref{fig:example_discontinuities},
a different magnitude of discontinuity between the chair and the wall 
in the background
could be achieved by
scaling the depth of
all points on the surface of the chair
by
the same
factor $s_1$; a factor $s_1 > 1$ would allow
rigidly moving
the chair closer to the wall, while a value $s_1 < 1$ would move the chair further away from it and closer to the camera.
While conceptually similar to
optimizing the depth of each chair- and wall
pixel, which relies on all inter-pixel constraints,
estimating the
\emph{relative scales}
of the chair and the wall greatly reduces
the
optimization complexity,
since it only requires
solving for
two values and involves only
constraints on the inter-object boundaries.

\begin{figure}[t]
    \centering
    \begin{subfigure}[b]{0.24\linewidth}
        \includegraphics[width=\linewidth]{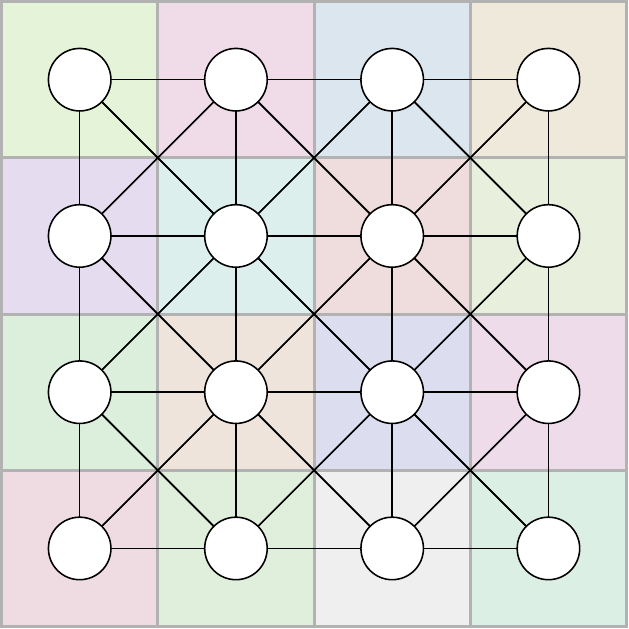}
        \caption{$\mathcal{G}_{0}=(V, E)$}
        \label{fig:graph_example_0}
    \end{subfigure}
    \hfill
    \begin{subfigure}[b]{0.24\linewidth}
        \includegraphics[width=\linewidth]{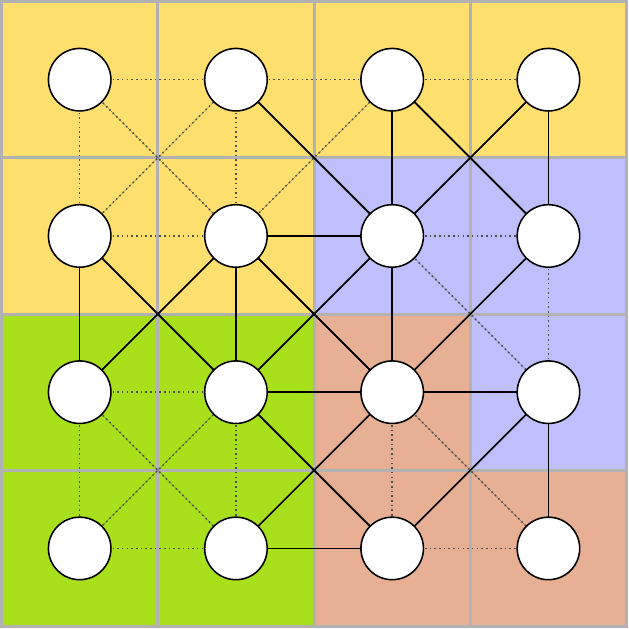}
        \caption{Partition of $V$}
        \label{fig:graph_example_1}
    \end{subfigure}
    \hfill
    \begin{subfigure}[b]{0.24\linewidth}
        \includegraphics[width=\linewidth]{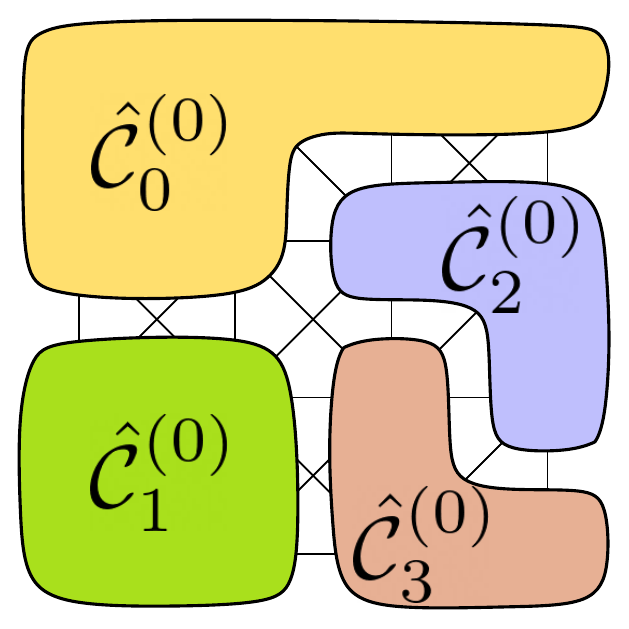}
        \caption{
        $\mathcal{Q}_0$
        }
        \label{fig:graph_example_2}
    \end{subfigure}
    \hfill
    \begin{subfigure}[b]{0.24\linewidth}
        \includegraphics[width=\linewidth]{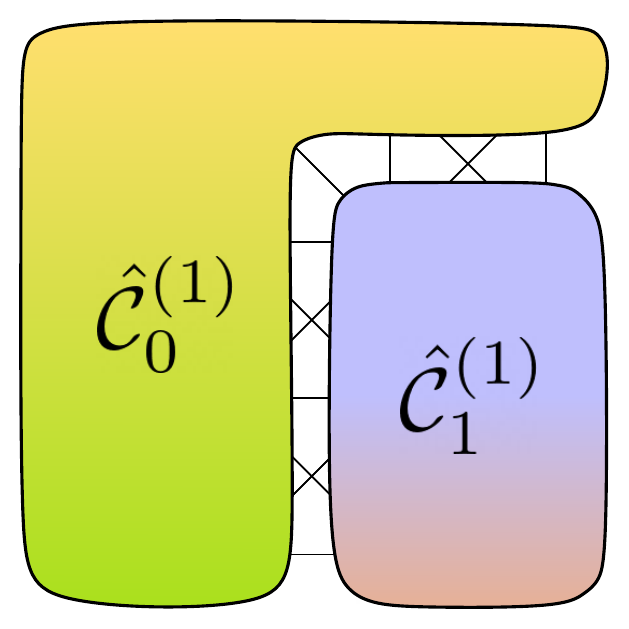}
        \caption{$\mathcal{Q}_1$}
        \label{fig:graph_example_3}
    \end{subfigure}
    \caption{\textbf{Graph-theory interpretation of our framework.} \textit{(a)} Pixel-level optimization can be seen as operating on a dense graph $\mathcal{G}_0=(V, E)$. \textit{(b)} We propose partitioning the per-pixel vertices $V$ into 
    components. \textit{(c)} The resulting optimization operates on a \emph{quotient graph} $\mathcal{Q}_0$~\eqref{eq:quotient_graph_definition}. \textit{(d)} Optionally, components can be merged into one another, forming a reduced quotient graph (\cref{sec:method_optional_component_merging}).}
    \label{fig:graph_example}
\end{figure}

To better appreciate
the difference in complexity,
let us look at the
problem from a graph-theoretic perspective, which also allows us to introduce a notation that will be convenient to describe
our framework's components
in the subsequent Sections.
Existing approaches based on global optimization of per-pixel (log-)depth values are akin to modeling the problem using a dense graph
$\mathcal{G}_{0}=(V, E)$,
where the set of vertices $V$ comprises a node for each valid pixel in the input normal map and the set of edges $E$ is
defined by all pairs of valid neighboring pixels according to the chosen pixel connectivity (\cref{fig:graph_example_0}). The optimization problem then consists of estimating the value of a variable for each node in the graph, with constraints defined by the graph edges.

When operating with our proposed \emph{continuous components}, optimization can
instead
be seen as based
on a meta-graph, or \emph{quotient graph} $Q_m$, obtained from $\mathcal{G}_0$ by partitioning its vertices $V$ into a set of components $\{\mathcal{C}^{(m)}_c\}$, s.t. $V = \bigcup_c\mathcal{C}^{(m)}_c$ (\cref{fig:graph_example_1}), where we index with $m$ a specific version of the partitioning, which can subsequently be updated.
In particular,
since all pixels in a component $\mathcal{C}^{(m)}_c$ belong to the same
surface patch,
we let
their
scale
change jointly
and
define
a single meta-node $\hat{\mathcal{C}}_c^{(m)}$
in the
meta-graph.
Since each surface patch is considered fixed up to scale, all constraints relating pixels in the same component are ignored when optimizing the relative scale of the
components. Therefore,
the edges in the meta-graph only include constraints between pixels in different components.

Formally,
for
each component $\mathcal{C}^{(m)}_c$
let us
denote the set of
its \emph{intra-component} edges 
(with both endpoints in $\mathcal{C}^{(m)}_c$) and the set of its \emph{inter-component} edges 
(with only one endpoint in $\mathcal{C}^{(m)}_c$) as
\begin{align}
\begin{split}
\displaystyle
E(\mathcal{C}^{(m)}_c)&\coloneqq\{(a, b)\in E\ |\ a,b \in\mathcal{C}^{(m)}_c\}\\
\partial \mathcal{C}^{(m)}_c&\coloneqq\{(a, b)\in E\ |\ a\in\mathcal{C}^{(m)}_c\ \land\ b\notin\mathcal{C}^{(m)}_c\}.
\end{split}
\end{align}
Let us
further
refer to
the set of all intra-component edges and the set of all inter-component edges respectively as
\begin{equation}   
E^{(m)}_{\mathrm{intra}}\coloneqq \bigcup_{c}E(\mathcal{C}^{(m)}_c),\ E^{(m)}_{\mathrm{inter}}\coloneqq \bigcup_{c}\partial\mathcal{C}^{(m)}_c,
\end{equation}
and let us
denote
with $\pi_{m}$ the mapping from
a
pixel to the index of its corresponding
node
in the
quotient graph $\mathcal{Q}_m$:

\begin{equation}
    \pi_{m}:V\mapsto\{0, \dots, |C^{(m)}| - 1\},\ \mathrm{s.t.}\ 
    \forall a\in V, a\in\mathcal{C}^{(m)}_{\pi_{m}(a)}.
\end{equation}
With the above notation,
we define the quotient graph as
\begin{equation}
    \mathcal{Q}_m = (C^{(m)}, E^{(m)}_\mathrm{inter}),\  \textrm{with}\ \ \ \ C^{(m)}\coloneqq\{\hat{\mathcal{C}}^{(m)}_c\}.
    \label{eq:quotient_graph_definition}
\end{equation}

At
optimization iteration $t$, with partitioning version $m$, 
the goal of the optimization
is
to
scale, for each component $\mathcal{C}^{(m)}_c$,
the depth values $\mathbf{z_c}^{(t)} \coloneqq (z_a\ |\ a\in\mathcal{C}^{(m)}_c)^{\mathsf{T}}$ of all pixels in $\mathcal{C}^{(m)}_c$ by a multiplicative factor $s_c$, or equivalently to estimate an additive factor $\tilde{s}_c\coloneqq\log(s_c)$ to apply to all corresponding log-depth values $\mathbf{\tilde{z}_c}^{(t)} = (\tilde{z}_a\ |\ a\in\mathcal{C}^{(m)}_c)^{\mathsf{T}}$.
We therefore set as optimization variable
the vector of relative scales to apply to each component,
\begin{equation}
\mathbf{\tilde{s}}^{(t)}\coloneqq\left(\tilde{s}^{(t)}_0, \dots, \tilde{s}^{(t)}_{|C^{(m)} - 1|}\right)^{\mathsf{T}},
\end{equation}
and
we
define the optimization constraints by
imposing
the following
updated
version of the
model of continuity 
uniquely
on the inter-component edges
$(a, b)\in E^{(m)}_{\mathrm{inter}}$:
\begin{equation}
    \overline{\chi}^{(t)}_{b\rightarrow a} \coloneqq \tilde{z}^{(t-1)}_a - \tilde{z}^{(t-1)}_b - \tilde{\omega}_{b\rightarrow a} + \tilde{s}^{(t)}_{\pi_{m}(a)} - \tilde{s}^{(t)}_{\pi_{m}(b)}.
    \label{eq:model_continuity_with_residuals_relative_scales}
\end{equation}
After each optimization iteration, the log-depth at each pixel is efficiently updated in parallel 
by rigidly scaling
all pixels in the same component by the same factor, \ie, $\forall a \in V$,
\begin{equation}
    \tilde{z}^{(t+1)}_a \gets \tilde{z}^{(t)}_a + \tilde{s}^{(t)}_{\pi_{m}(a)},
    \label{eq:log_depth_update_relative_scale_ver_1}
\end{equation}
or equivalently, $\forall c\in\{0, \dots, |C^{(m)}| -1\}$,
\begin{equation}
\mathbf{\tilde{z}_c}^{(t+1)}\gets\mathbf{\tilde{z}_c}^{(t)} + \tilde{s}_c^{(t)}\mathbf{1}.
\label{eq:log_depth_update_relative_scale_ver_2}
\end{equation}

As a final remark,
it is worth noting
that our relative-scale framework
can
also be applied in the limit where each component only contains a single pixel. In this case, the quotient graph
coincides
with the dense graph $\mathcal{G}_0$, hence
there
are
no advantages
over per-pixel log-depth optimization
in terms of number of
variables and constraints. However, as we show in \Cref{sec:experiments}, we find that our formulation converges to a slightly more accurate solution.

\subsection{Formation and filling of the initial components~\label{sec:method_formation_filling_initial_components}}
To exploit the computational advantages of our relative-scale framework, we need a decomposition of the input normal map into regions that are likely to correspond to continuous surface patches. For this, we propose a simple but effective heuristic based on the observation that
if the derivative $f^{\prime}$ of a generic signal $f$ is continuous at a point, then the signal itself is continuous at that point. Since surface normals represent a form of derivative of the depth map to reconstruct (hence the name ``normal \emph{integration}''), we propose to
exploit
the local continuity of surface normals as a proxy for the continuity of the underlying surface. In particular, for each pair of neighboring pixels $(a, b)\in E$, we compute the relative angle, $\theta_{a,b} \coloneqq\arccos\left(\boldsymbol{n_a}^\mathsf{T}\boldsymbol{n_b}\right)$,
between the normals $\boldsymbol{n_a}$ and $\boldsymbol{n_b}$ at each pixel.
We then form continuous components $\{\mathcal{C}^{(0)}_c\}$ by finding the connected components of the subgraph obtained from $\mathcal{G}_0$ by only selecting edges for which $\theta_{a, b} < \theta_{c}$, where $\theta_c$ is a hyperparameter. As we show in \Cref{sec:experiments}, we find that for small values of $\theta_c$, this simple heuristic allows effectively recovering large, approximately continuous surface regions.

Once the continuous components are detected,
we
perform
a separate optimization
for each component $\mathcal{C}^{(0)}_c$,
through which we ``fill'' the log-depth values of its
pixels. We
apply
a
regular log-depth model of continuity~\eqref{eq:generic_model_continuity_with_discontinuities} 
and the
BiNI model of discontinuity~\eqref{eq:total_weight_bini_model_of_discontinuity} 
on its intra-component edges $(a, b)\in E(\mathcal{C}^{(0)}_c)$, expressed in matrix form as:
\begin{align}
    \begin{split}
    \mathbf{A_c} &= \begin{bmatrix}
    1 & -1 & 0 & \cdots\\
    -1 & 1 & 0 & \cdots\\
    \vdots & \vdots & \vdots & \ddots\\
    \end{bmatrix},\ \mathbf{b_c}=\begin{bmatrix}
        \tilde{\omega}_{b\rightarrow a}\\
        \tilde{\omega}_{a\rightarrow b}\\
        \vdots
    \end{bmatrix},\\
    \mathbf{W_c}&=\mathrm{diag}\left(\{W_{b\rightarrow a}\}_{(a,b) \in E(\mathcal{C}^{(0)}_c)}\right).
    \end{split}
    \label{eq:intracomponent_matrices}
\end{align}

Importantly, since no two components share any variables or constraints, the above optimization problem can be solved independently and in parallel for each of the components. Similarly to previous log-depth-based methods~\cite{Cao2022BiNI, Milano2025DiscontinuityAwareNormalIntegration}, we initialize all log-depth values to $0$ and perform optimization using the conjugate-gradient method~\cite{Hestenes1952ConjugateGradient}, denoted as $\mathrm{cg}(\mathbf{A}, \mathbf{b})$, on the normal equations:
\begin{equation}
    \mathbf{\tilde{z}_c}^{(0)} \gets   \texttt{cg}\left(\mathbf{A_c}^\mathsf{T}\mathbf{W_c}\mathbf{A_c}, \mathbf{A_c}^\mathsf{T}\mathbf{W_c}\mathbf{b_c}\right).
\end{equation}
We find a single iteration of the above optimization to be sufficient to
accurately reconstruct
each surface patch.

\subsection{Relative scale optimization~\label{sec:method_relative_scale_optimization}}
Once the continuous components have each been independently reconstructed, an optimization of their relative scales is necessary to recover the full surface. For a given partitioning version $m$ and an optimization iteration $t$, 
we rewrite our updated model of continuity~\eqref{eq:model_continuity_with_residuals_relative_scales} as follows,
\begin{align}
    \begin{split}
    \overline{\mathbf{A}}_m &= \overset{
    \scriptsize
  \begin{array}{cccccc}
    & & \pi_{m}(a) & & \pi_{m}(b) &
  \end{array}
}{\begin{bmatrix}
    \cdots & 0 & 1 & \cdots & -1 & \cdots\\
    \cdots & 0& -1 & \cdots & 1 & \cdots\\
    \vdots & \vdots & \vdots & \vdots & \vdots & \ddots\\
    \end{bmatrix}},\\     \overline{\mathbf{b}}_m&=\begin{bmatrix}
        \tilde{\omega}_{b\rightarrow a} - (\tilde{z}^{(t-1)}_a - \tilde{z}^{(t-1)}_b)\\
        \tilde{\omega}_{a\rightarrow b} - (\tilde{z}^{(t-1)}_b - \tilde{z}^{(t-1)}_a)\\
        \vdots
    \end{bmatrix},
    \end{split}
    \label{eq:intercomponent_matrices}
\end{align}
where the constraints are defined on the inter-component edges $(a, b) \in E^{(m)}_{\mathrm{inter}}$, rearranged so that the non-zero columns of the design matrix correspond to the indices $\pi_{m}(a)$ and $\pi_{m}(b)$ of the connected components.
We then retrieve the logarithmic scale to be applied to each component (\eqref{eq:log_depth_update_relative_scale_ver_1}, \eqref{eq:log_depth_update_relative_scale_ver_2}) through conjugate gradient:
\begin{equation}
\mathbf{\tilde{s}}^{(t)}\gets\texttt{cg}\left(\mathbf{\overline{A}}_m^{\mathsf{T}}\mathbf{\overline{W}}^{(t)}_{m}\mathbf{\overline{A}}_m, \mathbf{\overline{A}}_m^{\mathsf{T}}\mathbf{\overline{W}}^{(t)}_{m}\mathbf{\overline{b}}_m\right).
\end{equation}

Importantly, we note that after the initial filling of the components, the scale of each per-component reconstruction might differ significantly, since each optimization retrieves
log-depth maps only
up to scale (\cref{fig:method_overview}a).
Thus, the inter-component residuals may be arbitrary and hence imbalance the
subsequent
optimization. Therefore, in the first iterations of relative scale optimization (we find $2$ iterations to yield
accurate
results), we weight all constraints~\eqref{eq:model_continuity_with_residuals_relative_scales} equally, to 
align the components in the most continuous way possible; \ie, for $t \le 1$, we set $\overline{\mathbf{W}}^{(t)}_m=\mathrm{diag}(1)$. In subsequent iterations, to retrieve the discontinuities we adopt the BiNI model of discontinuity~\eqref{eq:total_weight_bini_model_of_discontinuity}.

However, we find that directly applying such a model results in suboptimal reconstructions. The reason for this is that the quotient graph $\mathcal{Q}_m$ has in general a much lower number of edges than $\mathcal{G}_0$. As a consequence, constraints with large residuals that might have only partially affected the optimization in a dense graph assume a much larger weight in the smaller component-based
optimization.
To address this,
we introduce an \emph{outlier reweighting} strategy that reduces
the influence of large residuals in subsequent iterations.
In particular, we define two thresholds $L, U$, with $0 < L < U$, such that an equation with residual $\overline{\chi}_{b\rightarrow a}$
is considered
an
outlier if $|\overline{\chi}_{b\rightarrow a}|\ge U$ and an inlier
if $|\overline{\chi}_{b\rightarrow a}|\le L$.
We model this through a soft weight,
\begin{equation}
    W^{{\mathrm{out}}^{(t)}}_{b\rightarrow a}\coloneqq\sigma_1\left(\frac{4}{\tilde{L} - \tilde{U}}(2\log_{10}(|\overline{\chi}^{(t-1)}_{b\rightarrow a}|) - (\tilde{L} + \tilde{U}))\right),
    \label{eq:outlier_reweighting}
\end{equation}   
where
$\tilde{\cdot} \coloneqq \log_{10}(\cdot)$, 
and we set in our experiments $U=10^{-3}$ and $L=10^{-5}$.
This provides an affine mapping in $\mathrm{log}$ space, with $U$ mapped to $-4$ and $L$ mapped to $4$, resulting in an outlier weight of
$\sigma_1(-4)\approx 0.02$
when $|\chi_{b\rightarrow a}| = U$ and of
$\sigma_1(4)\approx 0.98$
when $|\chi_{b\rightarrow a}| = L$.
For all
iterations
$t> 1$, we
multiply~\eqref{eq:outlier_reweighting}
to the BiNI weight, so that the overall weight matrix is,
\begin{equation}
\overline{\mathbf{W}}^{(t)}_m=\mathrm{diag}\left(\{W_{b\rightarrow a}\cdot W_{b\rightarrow a}^{\mathrm{out}}\}_{(a, b)\in E^{(m)}_{\mathrm{inter}}}\right).
\end{equation}

We run relative scale optimization until 
the optimization energy $E_t\coloneqq(\mathbf{\overline{A}}_m\mathbf{\tilde{s}}^{(t)} - \mathbf{\overline{b}}_m )^{\mathsf{T}}\mathbf{\overline{W}}^{(t)}_{m}(\mathbf{\overline{A}}_m\mathbf{\tilde{s}}^{(t)} - \mathbf{\overline{b}}_m)$
changes
in relative terms
by less than a threshold
$\Delta E_{\mathrm{max}}$ or the number of iterations reaches a pre-defined value $T$.

\subsection{Optional merging~\label{sec:method_optional_component_merging}}
Optionally, to further reduce the number of variables and constraints of the optimization, 
components can be iteratively merged into
one another
(\cref{fig:method_overview}c).
After a certain number, $\mathrm{freq}_{\mathrm{merging}}$, of relative-scale optimization iterations
with a specific partitioning
$\{\mathcal{C}^{(m)}_c\}$, 
we can identify a new set of components
as follows: \textit{(i)} For each node $\hat{\mathcal{C}}^{(m)}_{c}$ in the quotient graph $\mathcal{Q}_m$, select the edge among those incident to it that
has
the smallest residual magnitude $|\overline{\chi}_{b\rightarrow a}|$, \ie, $(\hat{a},\hat{b})^{(m)}_c\coloneqq\arg\min_{(a,b)\in\partial\mathcal{C}^{(m)}_{c}} |\overline{\chi}_{b\rightarrow a}|$. \textit{(ii)} Form a subgraph 
$\hat{\mathcal{Q}}_m=(C^{(m)}, \{(\hat{a},\hat{b})^{(m)}_c\})$
from $\mathcal{Q}_m$
using the selected edges.
\textit{(iii)} Find a new set of components $\{\mathcal{C}^{(m+1)}_c\}$ by computing the connected components of
$\hat{\mathcal{Q}}_m$.
The process can then continue by optimizing for the relative scale of the new components (\cref{sec:method_relative_scale_optimization}).
\section{Experiments~\label{sec:experiments}}

\subsection{Experimental settings}
Our implementation relies on a combination of GPU-accelerated tensor operations and CPU-based operations (\eg,
conjugate-gradient-based optimization,
for which we use the 
SciPy
library~\cite{Virtanen2020SciPy}).
We compare our method to the current state-of-the-art approach of~\cite{Milano2025DiscontinuityAwareNormalIntegration},
noting that
we also
use
their model of continuity in our framework.
We reimplement
the
baseline
and integrate
it
into our framework, to
decouple
the effects of GPU acceleration and 
formulation-specific computational aspects.
All
experiments are run on a single NVIDIA RTX $3080$ Laptop
GPU.

\subsection{Benchmark experiments}
We evaluate on the DiLiGenT benchmark~\cite{Shi2016DiLiGenT} for normal integration, which features ground-truth, object-level normal maps of resolution $608\times512$. In these
maps the background is masked out, and a normal vector is defined only
at
the object
pixels,
resulting in
$\num{24706}$ to $\num{56560}$
depth values to be estimated.
For each object and run we report the mean average depth error (MADE) and the total execution time.

For both methods we use $8$-connectivity and convergence parameters $\Delta E_{\mathrm{max}} = 10^{-3}$, $T = 150$. We refer the reader to the Supplementary for ablations on these parameters. 
Additionally, while slight speed-ups of our method can be achieved through our optional merging (see Supplementary), we find that this effect is not
particularly
significant, given the small number of valid pixels in the input normal maps. We therefore do not use merging in our DiLiGenT experiments, and show instead
its
benefit
on larger-resolution maps in the next Section. 
We test our method for different values of the normal similarity threshold $\theta_c$, including the limit case in which each pixel is assigned to a component, which we denote as $\theta_c = \mathrm{None}$. For the baseline, we include both the version with and without discontinuity computation ($\alpha_{b\rightarrow a}$ in~\cite{Milano2025DiscontinuityAwareNormalIntegration}).
Finally, we note that our proposed outlier reweighting mechanism can also be
applied to
the log-depth-based optimization of \cite{Milano2025DiscontinuityAwareNormalIntegration} and
we
therefore evaluate both methods with and without 
it.

The results of our evaluation are shown in \Cref{tab:diligent_results_main}.
We start by noting that our outlier reweighting 
significantly improves
the convergence time for \emph{pixel-level} approaches, not only for our relative scale optimization, but also for the log-depth-based baseline of~\cite{Milano2025DiscontinuityAwareNormalIntegration}. Importantly, for
this baseline,
outlier reweighting also improved accuracy, 
although
with the exception
of some
objects (\cf, \texttt{buddha}, \texttt{cow}, \texttt{pot1}).
Additionally, we note that while the discontinuity computation strategy of~\cite{Milano2025DiscontinuityAwareNormalIntegration} is beneficial for objects with large discontinuities (\texttt{buddha}, \texttt{harvest}, $\mathrm{D}$ vs. $\mathrm{ND}$ in \Cref{tab:diligent_results_main}), its effect is partially reduced by the use of outlier 
reweighting.

Crucially, for our component-level versions, outlier reweighting is
critical
for
achieving
optimal
accuracy, resulting
in significantly better and more stable convergence across the board, particularly for objects with large discontinuities (\texttt{buddha}, \texttt{harvest}, \texttt{reading}, \texttt{goblet}), as can be seen by comparing successive rows in \Cref{tab:diligent_results_main}.
Overall, while with
outlier reweighting the performance of the two methods is comparable
for the most continuous objects, our
component-based variants
achieve
better
accuracy
than log-depth optimization for the more discontinuous objects (\texttt{buddha}, \texttt{harvest}, \texttt{reading}).
This effect
is a function of the chosen 
value
for the threshold
$\theta_c$,
with smaller values generally resulting in lower error at the cost of slightly increased runtime.
We refer to 
the Supplementary for
details on the minimum theoretical error that can be achieved for different values of $\theta_c$.
Notably, our
pixel-level version also
slightly outperforms
the (pixel-level) formulation of~\cite{Milano2025DiscontinuityAwareNormalIntegration}, suggesting that solving for relative scales instead of log-depth may
lead to a
more well-posed optimization problem.

Finally,
we note that, when using outlier reweighting both our component-based formulation and the pixel-level method of~\cite{Milano2025DiscontinuityAwareNormalIntegration} achieve comparable runtime on the DiLiGenT benchmark. As we show in the next Section, however, the gap between the execution times of the two frameworks becomes
significantly larger
as the number of valid pixels in the normal map increases.

\begin{table*}[!ht]
    \centering
    \resizebox{\linewidth}{!}{
    \begin{tabular}{l ll cc cc cc cc cc cc cc cc cc}
    \toprule
    \multirow{2}{*}{Method} & & & \multicolumn{2}{c}{\texttt{bear}} & \multicolumn{2}{c}{\texttt{buddha}} & \multicolumn{2}{c}{\texttt{cat}} & \multicolumn{2}{c}{\texttt{cow}} & \multicolumn{2}{c}{\texttt{harvest}} & \multicolumn{2}{c}{\texttt{pot1}} & \multicolumn{2}{c}{\texttt{pot2}} & \multicolumn{2}{c}{\texttt{reading}} & \multicolumn{2}{c}{\texttt{goblet}\textsuperscript{*}}\\
    \cmidrule(lr){4-5} \cmidrule(lr){6-7} \cmidrule(lr){8-9} \cmidrule(lr){10-11} \cmidrule(lr){12-13} \cmidrule(lr){14-15} \cmidrule(lr){16-17} \cmidrule(lr){18-19} \cmidrule(lr){20-21}
    & $\theta_c$ & $W^{\mathrm{out}}_{b\rightarrow a}$
    & $\mathrm{Err}$ & $t$ & $\mathrm{Err}$ & $t$ & $\mathrm{Err}$ & $t$ & $\mathrm{Err}$ & $t$ & $\mathrm{Err}$ & $t$ & $\mathrm{Err}$ & $t$ & $\mathrm{Err}$ & $t$ & $\mathrm{Err}$ & $t$ & $\mathrm{Err}$ & $t$\\
    \midrule
    \multirow{2}{*}{\cite{Milano2025DiscontinuityAwareNormalIntegration}, $\mathrm{D}$} & N/A & \xmark & $\mathbf{0.02}$ & $2.70$ & $0.62$ & $3.32$ & $0.05$ & $2.98$ & $0.13$ & $1.49$ & $2.07$ & $6.96$ & $0.47$ & $\underline{3.74}$ & $0.15$ & $2.86$ & $0.25$ & $2.02$ & $\mathbf{4.04}$ & $3.45$ \\ 
    & N/A & \xcross & $0.07$ & $5.02$ & $0.44$ & $19.89$ & $\underline{0.04}$ & $32.68$ & $\mathbf{0.06}$ & $15.18$ & $2.80$ & $51.00$ & $\mathbf{0.36}$ & $54.31$ & $\underline{0.14}$ & $24.73$ & $0.92$ & $6.57$ & $7.96$ & $31.52$ \\ 
    \multirow{2}{*}{\cite{Milano2025DiscontinuityAwareNormalIntegration}, $\mathrm{ND}$} & N/A & \xmark & $\mathbf{0.02}$ & $1.84$ & $0.70$ & $3.04$ & $0.05$ & $1.87$ & $0.13$ & $1.15$ & $2.21$ & $5.99$ & $0.47$ & $4.13$ & $\underline{0.14}$ & $\mathbf{1.79}$ & $0.24$ & $\underline{1.62}$ & $\mathbf{4.04}$ & $1.62$ \\ 
    & N/A & \xcross & $\underline{0.05}$ & $20.62$ & $0.49$ & $37.97$ & $0.21$ & $90.13$ & $\underline{0.07}$ & $21.73$ & $2.38$ & $73.30$ & $\underline{0.37}$ & $5.31$ & $\underline{0.14}$ & $6.96$ & $1.31$ & $59.01$ & $9.38$ & $27.67$ \\ 
    \arrayrulecolor{gray!70}\specialrule{0.2pt}{0.2pt}{0.2pt}
    \arrayrulecolor{black}
    \multirow{8}{*}{Ours} & None & \xmark & $\mathbf{0.02}$ & $7.39$ & $0.20$ & $19.56$ & $\mathbf{0.03}$ & $36.98$ & $0.09$ & $3.35$ & $1.31$ & $44.35$ & $\mathbf{0.36}$ & $32.51$ & $\mathbf{0.13}$ & $9.83$ & $0.17$ & $3.40$ & $9.41$ & $4.78$ \\ 
    & None & \xcross & $0.06$ & $143.81$ & $0.46$ & $18.25$ & $0.20$ & $119.53$ & $\mathbf{0.06}$ & $63.15$ & $10.10$ & $181.82$ & $\mathbf{0.36}$ & $81.85$ & $\mathbf{0.13}$ & $40.81$ & $1.24$ & $48.81$ & $9.33$ & $71.70$ \\ 
    & $2.0^{\circ}$ & \xmark & $\mathbf{0.02}$ & $2.55$ & $0.17$ & $19.07$ & $\mathbf{0.03}$ & $3.11$ & $0.09$ & $2.54$ & $\mathbf{1.04}$ & $28.33$ & $\mathbf{0.36}$ & $6.26$ & $\underline{0.14}$ & $5.59$ & $\underline{0.10}$ & $6.54$ & $9.37$ & $2.33$ \\ 
    & $2.0^{\circ}$ & \xcross & $\mathbf{0.02}$ & $7.86$ & $0.42$ & $6.39$ & $\underline{0.04}$ & $3.21$ & $0.10$ & $7.28$ & $1.94$ & $6.19$ & $0.38$ & $28.34$ & $\underline{0.14}$ & $11.51$ & $0.74$ & $3.43$ & $9.56$ & $4.90$ \\ 
    & $3.5^{\circ}$ & \xmark & $\mathbf{0.02}$ & $1.27$ & $\mathbf{0.11}$ & $8.03$ & $\underline{0.04}$ & $1.50$ & $0.09$ & $1.53$ & $\underline{1.07}$ & $18.81$ & $0.38$ & $\mathbf{3.51}$ & $\underline{0.14}$ & $2.66$ & $\mathbf{0.09}$ & $2.68$ & $9.49$ & $1.40$ \\ 
    & $3.5^{\circ}$ & \xcross & $0.20$ & $3.16$ & $0.91$ & $2.91$ & $0.16$ & $1.54$ & $0.10$ & $3.19$ & $2.17$ & $\underline{4.39}$ & $0.46$ & $8.21$ & $\underline{0.14}$ & $6.50$ & $0.87$ & $1.77$ & $\underline{5.37}$ & $\underline{0.80}$ \\ 
    & $5.0^{\circ}$ & \xmark & $\mathbf{0.02}$ & $\underline{0.88}$ & $\underline{0.15}$ & $\underline{2.76}$ & $0.51$ & $\underline{1.24}$ & $0.39$ & $\underline{1.04}$ & $1.75$ & $7.29$ & $0.55$ & $5.80$ & $\mathbf{0.13}$ & $\underline{1.92}$ & $0.16$ & $\mathbf{1.49}$ & $9.62$ & $0.84$ \\ 
    & $5.0^{\circ}$ & \xcross & $0.17$ & $\mathbf{0.82}$ & $1.04$ & $\mathbf{2.23}$ & $0.51$ & $\mathbf{1.10}$ & $0.40$ & $\mathbf{0.97}$ & $2.51$ & $\mathbf{3.51}$ & $0.39$ & $4.89$ & $\underline{0.14}$ & $1.97$ & $0.24$ & $3.40$ & $6.78$ & $\mathbf{0.68}$ \\ 
    \bottomrule
    \end{tabular}
    }
    \caption{\textbf{Mean absolute depth error (MADE, abbreviated as $\mathrm{Err}$) [$\si{mm}$] and total execution time (abbreviated as $t$) [$\si{s}$] on the DiLiGenT benchmark~\cite{Shi2016DiLiGenT}.} For each object, \textbf{bold} and \underline{underlined} denote respectively the best and the second-best result across the methods. All experiments use $\Delta E_{\mathrm{max}}=10^{-3}$, $T=150$, and $8$-connectivity, without merging. $\mathrm{D}$ and $\mathrm{ND}$ denote the version of~\cite{Milano2025DiscontinuityAwareNormalIntegration} with and without discontinuity computation, respectively ($\alpha_{b\rightarrow a}$ in~\cite{Milano2025DiscontinuityAwareNormalIntegration}). \textsuperscript{*}This object contains a full depth discontinuity.}
    \label{tab:diligent_results_main}
\end{table*}

\subsection{Evaluation of scalability}
\begin{figure*}[!ht]
\centering
\def\colwidth{0.14\textwidth}
\def\minicolwidth{0.01\textwidth}
\newcolumntype{M}[1]{>{\centering\arraybackslash}m{#1}}
\addtolength{\tabcolsep}{-4pt}
\begin{tabular}{m{0.7em} m{0.7em} M{\colwidth} M{\colwidth} M{\minicolwidth} M{\colwidth} M{\colwidth} M{\minicolwidth} M{\colwidth} M{\colwidth}}
\begin{turn}{90}{Normal map}\end{turn} & &
\includegraphics[width=\linewidth]{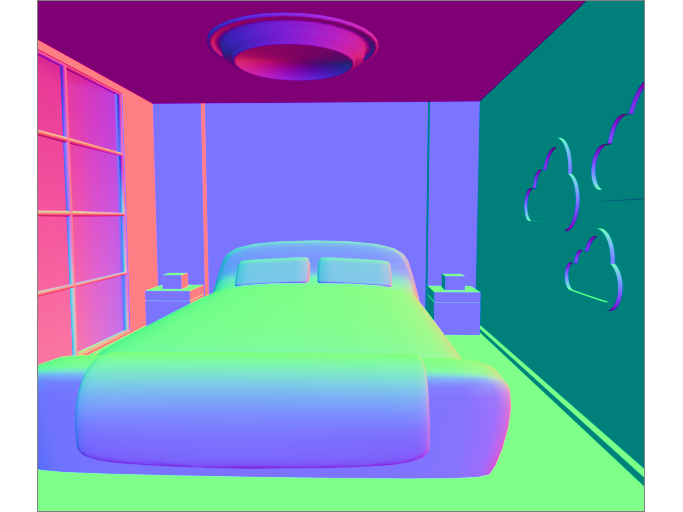} & 
\includegraphics[width=\linewidth]{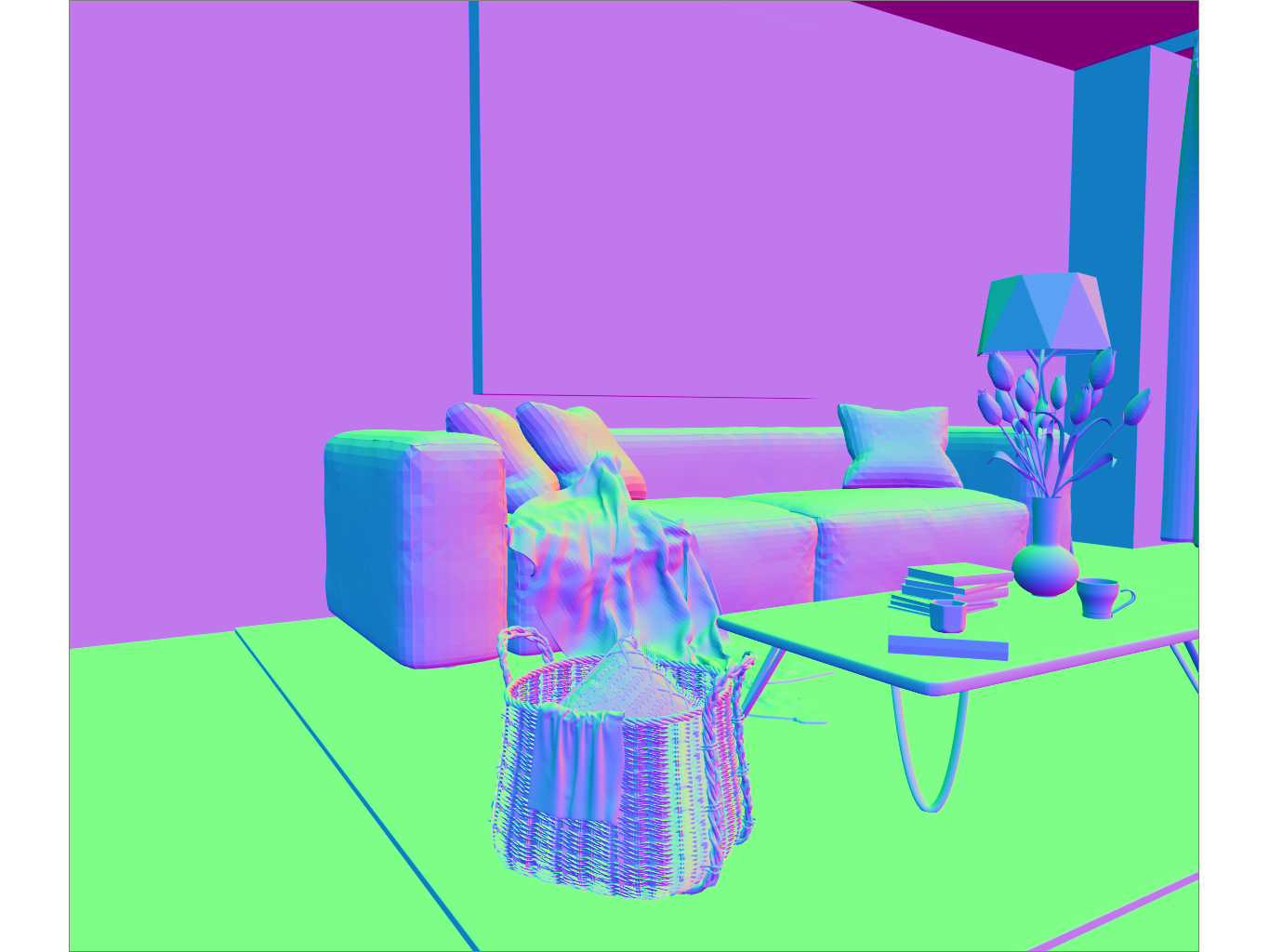} & & \includegraphics[width=\linewidth]{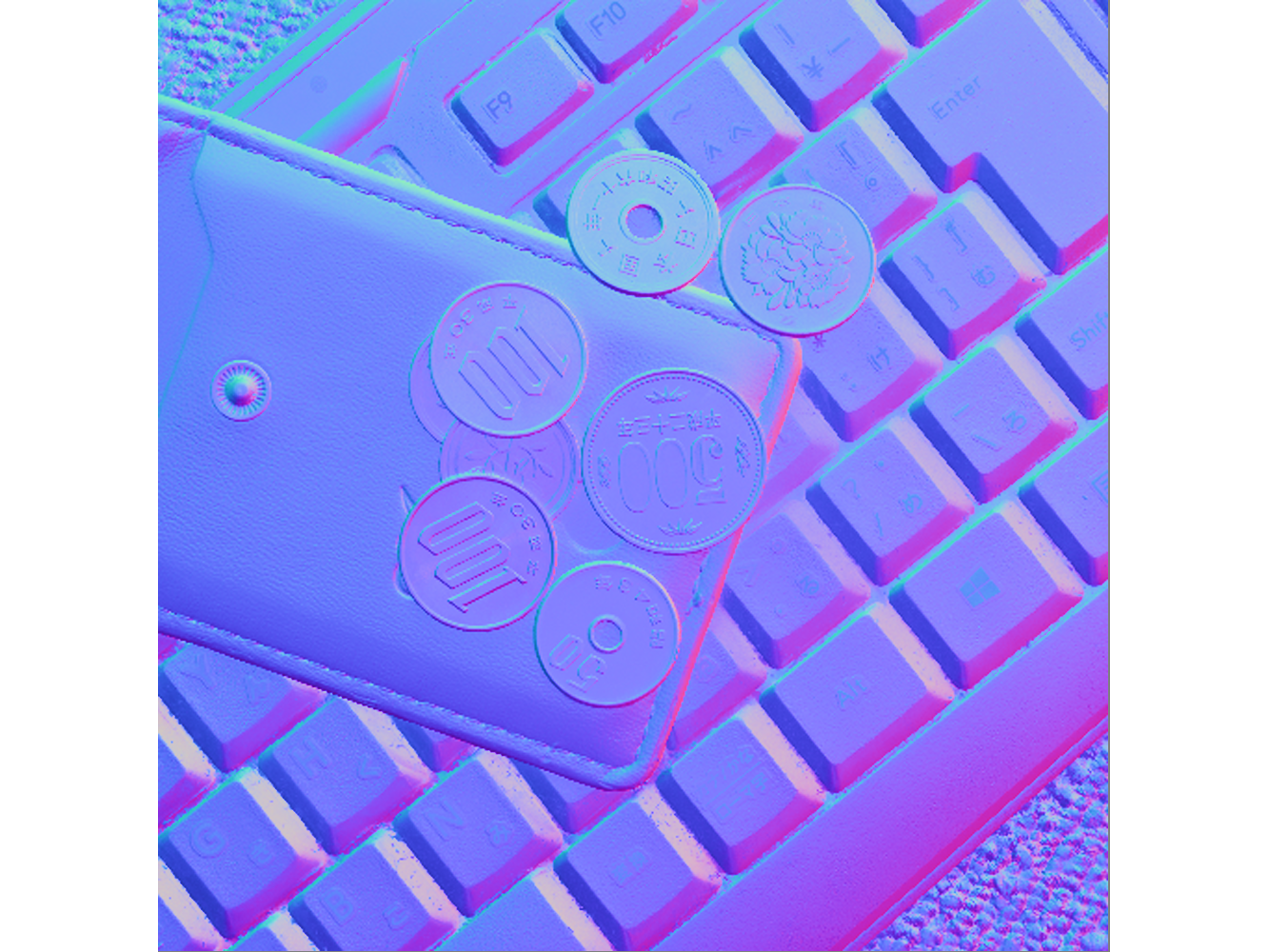} &
\includegraphics[width=\linewidth]{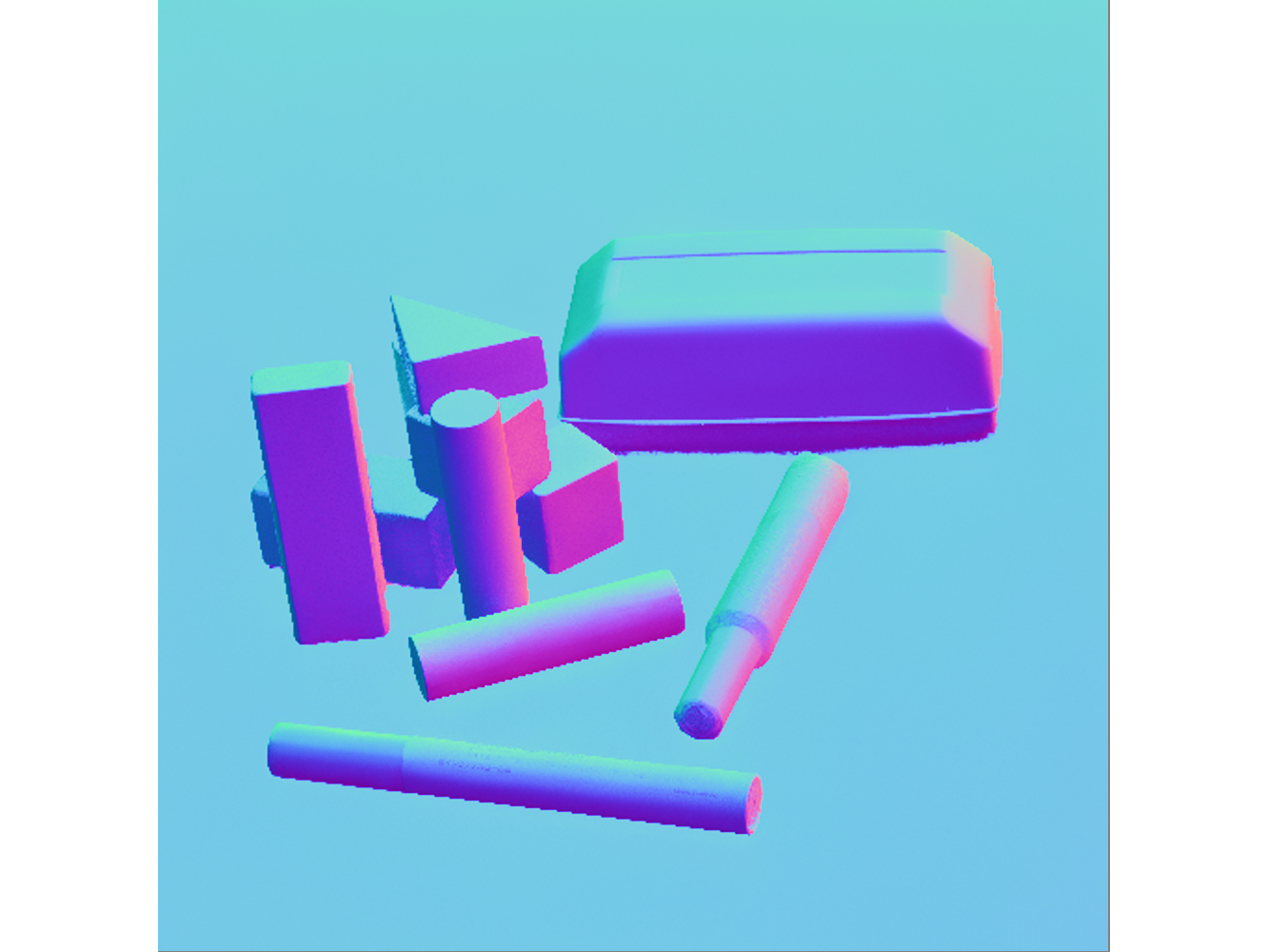} & &
\includegraphics[width=\linewidth]{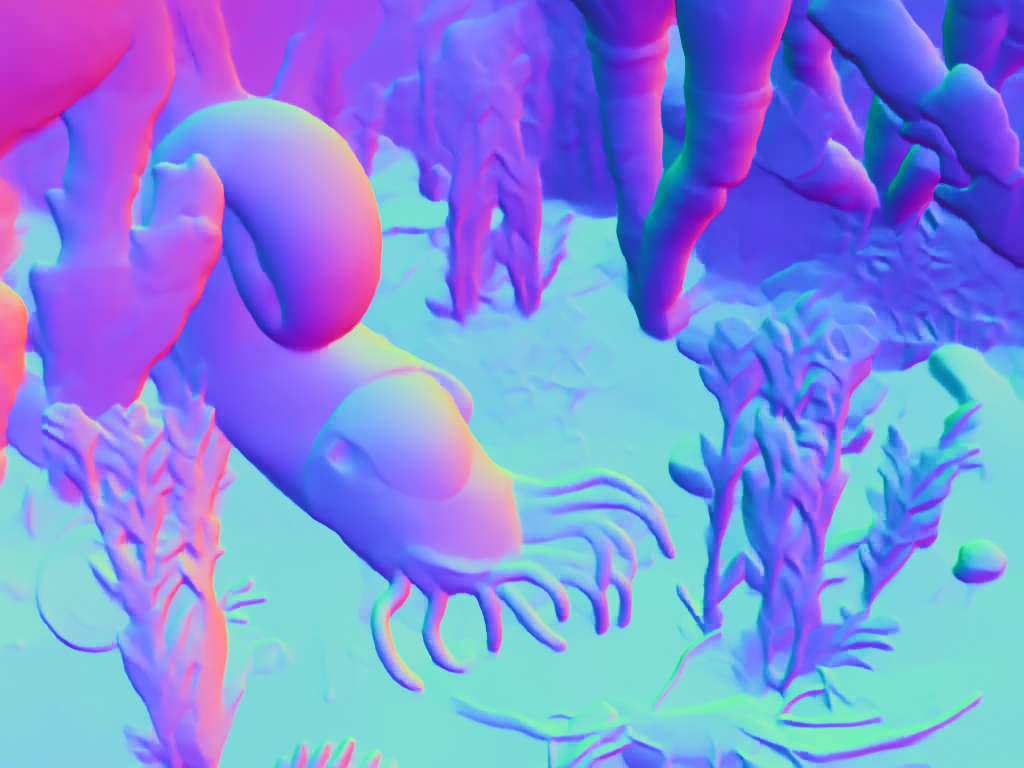} & 
\includegraphics[width=\linewidth]{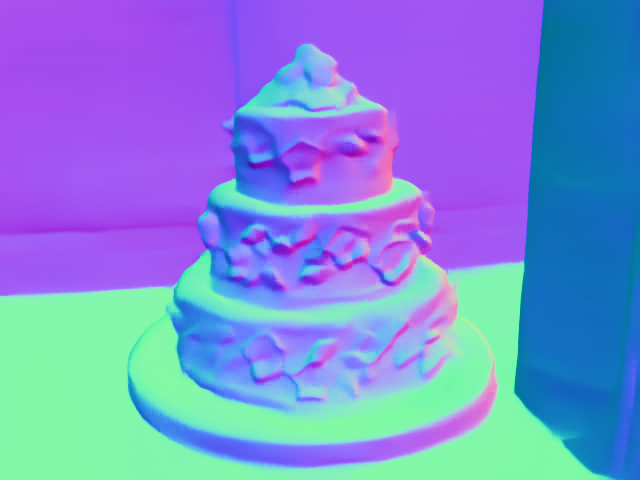}
\\[-5pt]
& & \scriptsize{Resolution: $608\times 512$} & \scriptsize{Resolut.: $1216\times 1024$} & & \scriptsize{Resolut.: $1000\times 1000$} & \scriptsize{Resolut.: $1000\times 1000$} & & \scriptsize{Resolut.: $1024\times 768$} & \scriptsize{Resolution: $640\times 480$}
\\
\noalign{\vskip -0.2em} 
\cline{3-10}
\noalign{\vskip 0.2em}
\multirow{5}{*}{\parbox[c][10ex][c]{\ht\strutbox}{\begin{turn}{90}
{Ours, $\theta_{c}=2.0^{\circ}$}
\end{turn}}} & \begin{turn}{90}
{\footnotesize Decomposition}
\end{turn} &
\includegraphics[width=\linewidth]{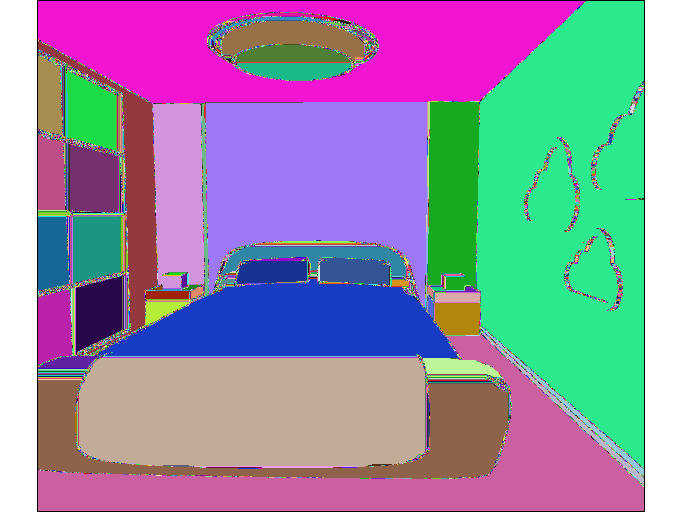} & 
\includegraphics[width=\linewidth]{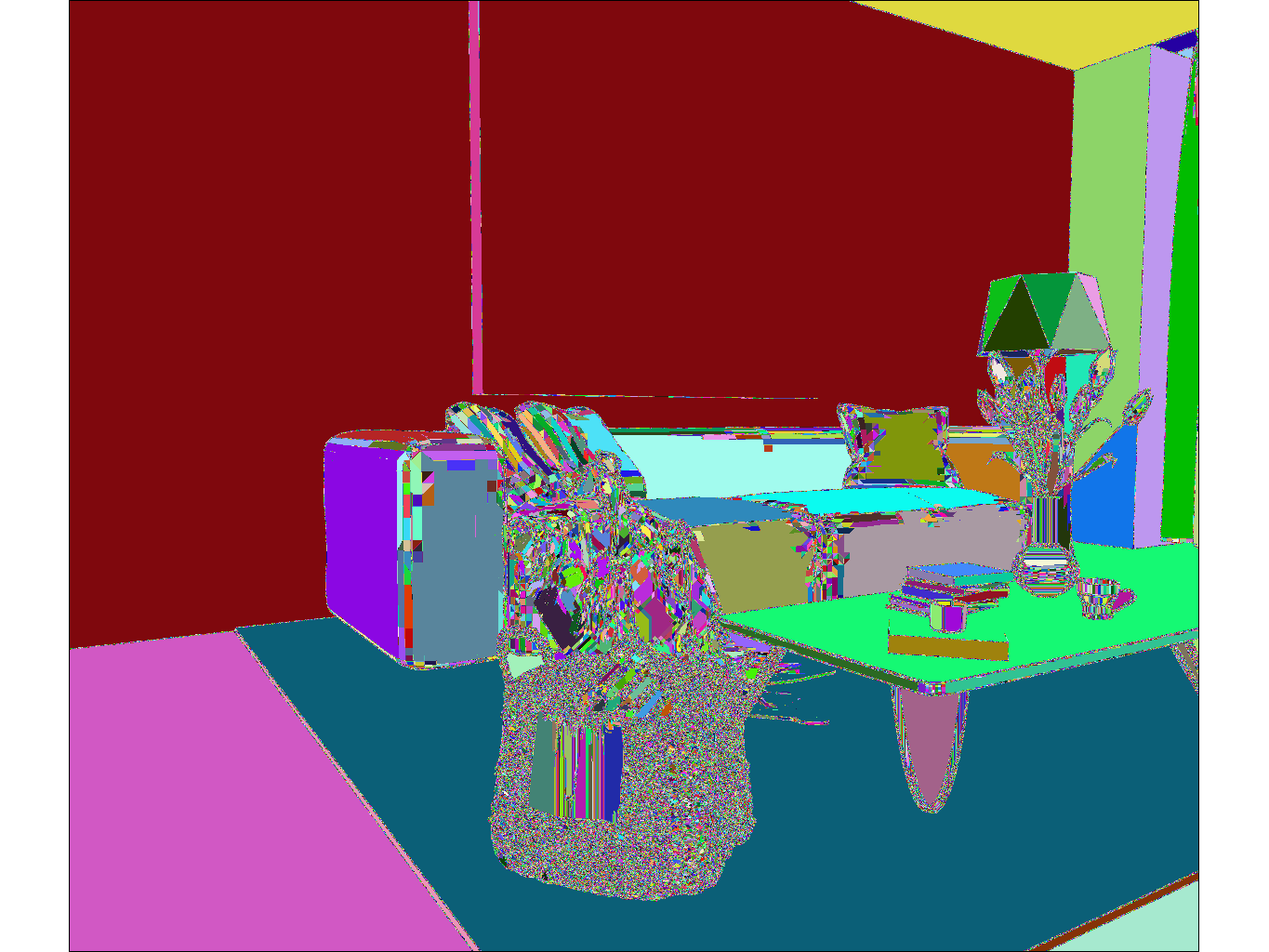} & &
\includegraphics[width=\linewidth]{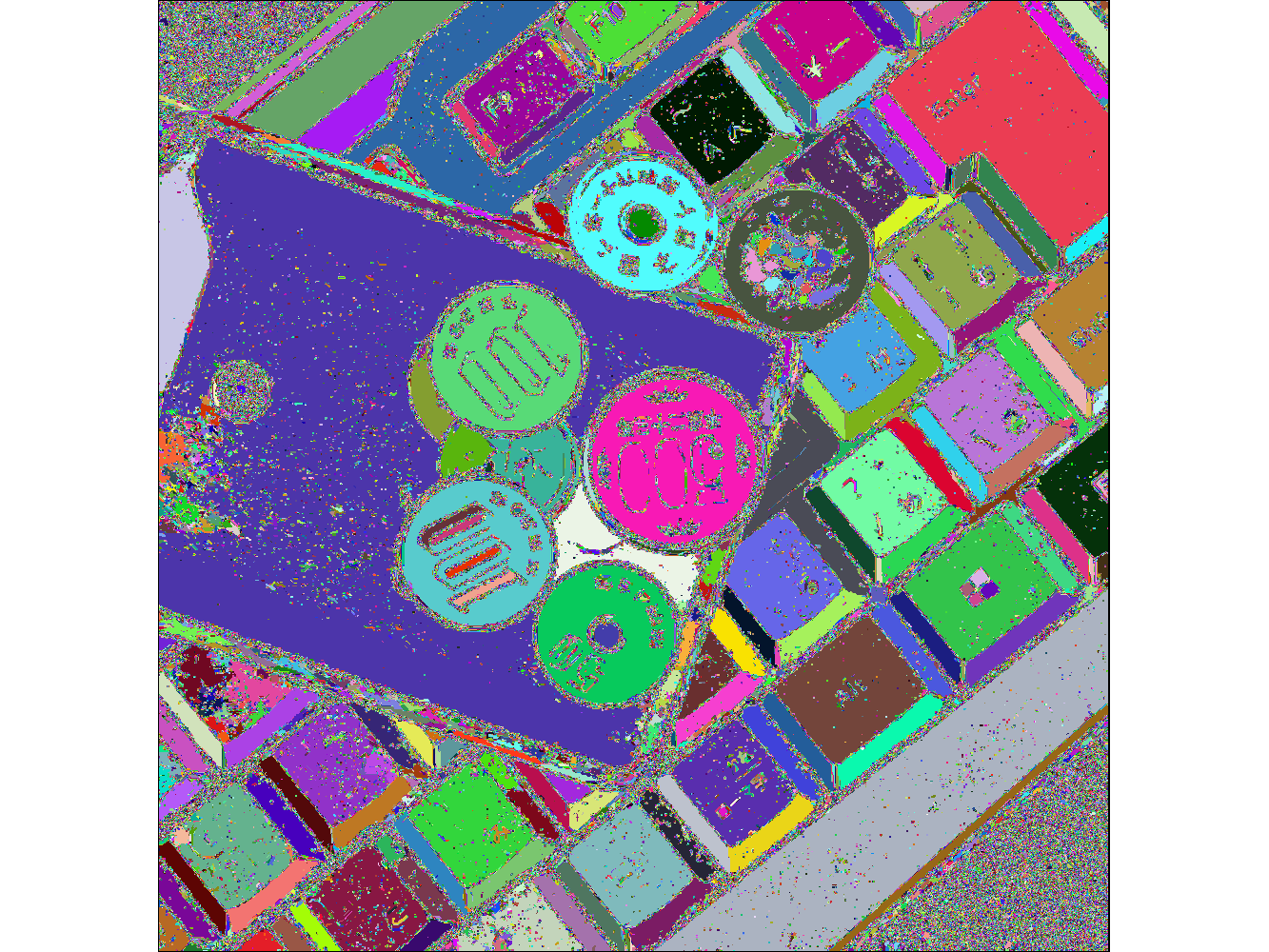} &
\includegraphics[width=\linewidth]{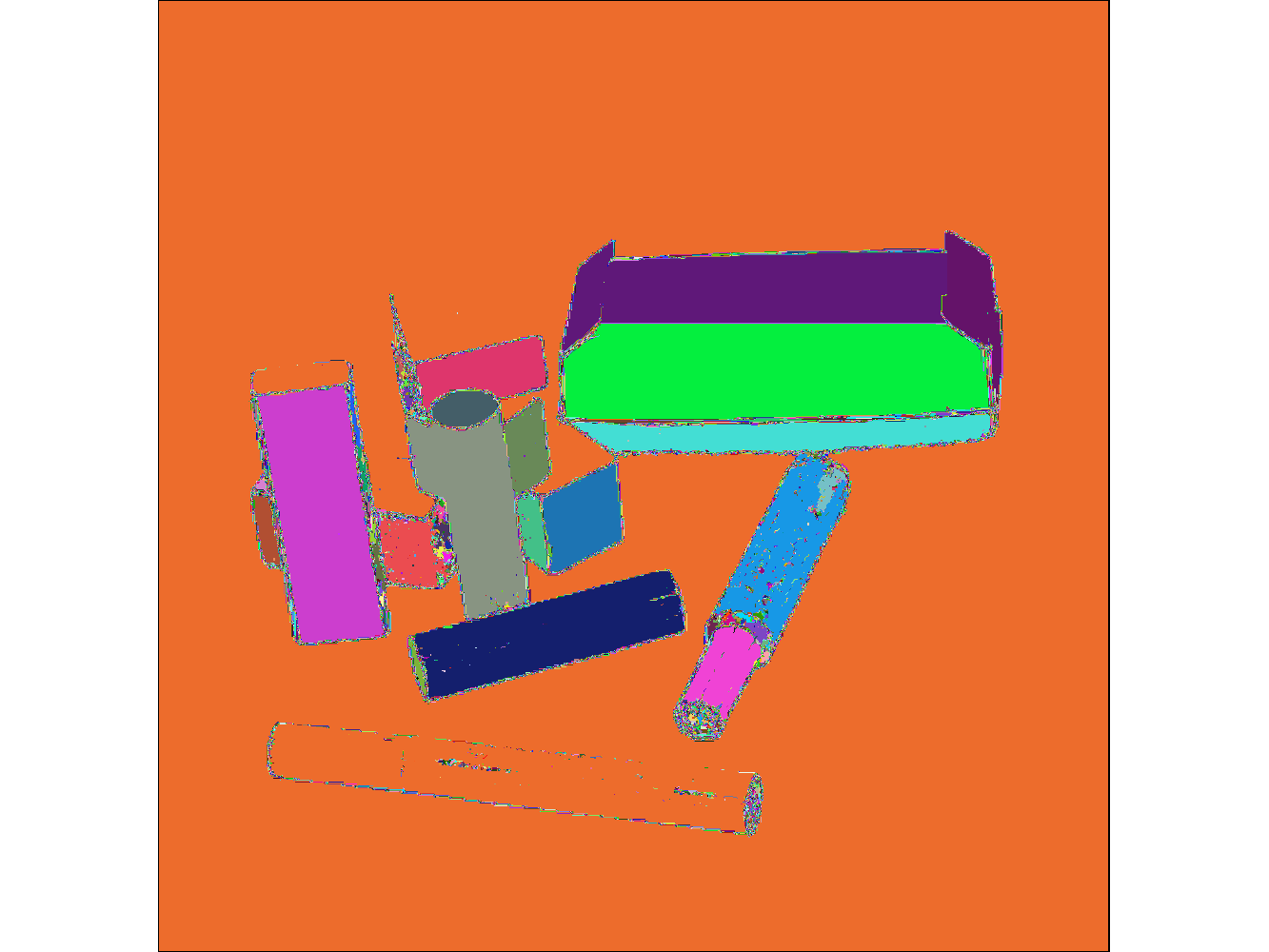} & &
\includegraphics[width=\linewidth]{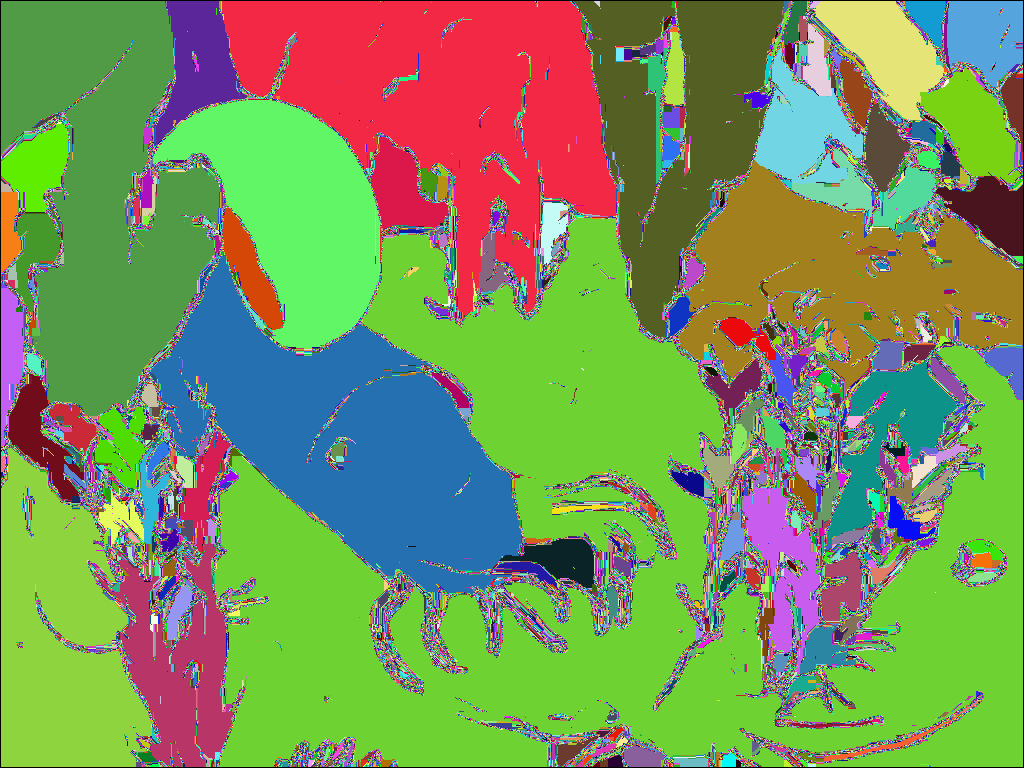} &
\includegraphics[width=\linewidth]{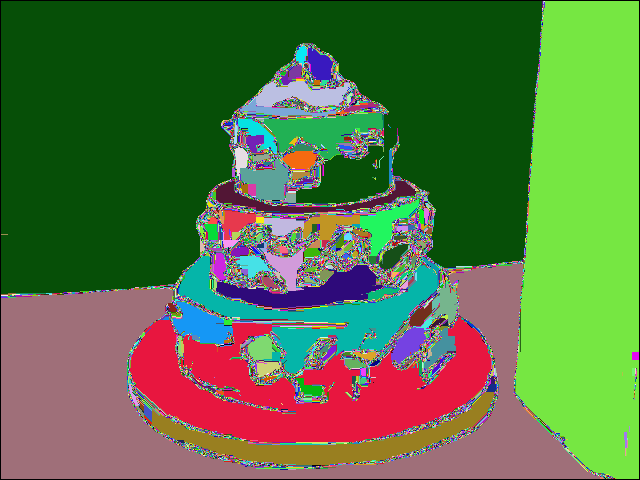}
\\[-5pt]
& & \scriptsize{\#components = $\num{10137}$} & \scriptsize{\#components = $\num{95266}$} & & \scriptsize{\#compon. = $\num{185209}$} & \scriptsize{\#components = $\num{19963}$} & & \scriptsize{\#components = $\num{72828}$} & \scriptsize{\#components = $\num{19567}$}
\\
& \begin{turn}{90}
{\footnotesize Recon. ($\mathrm{M}$)}
\end{turn} &
\includegraphics[width=\linewidth]{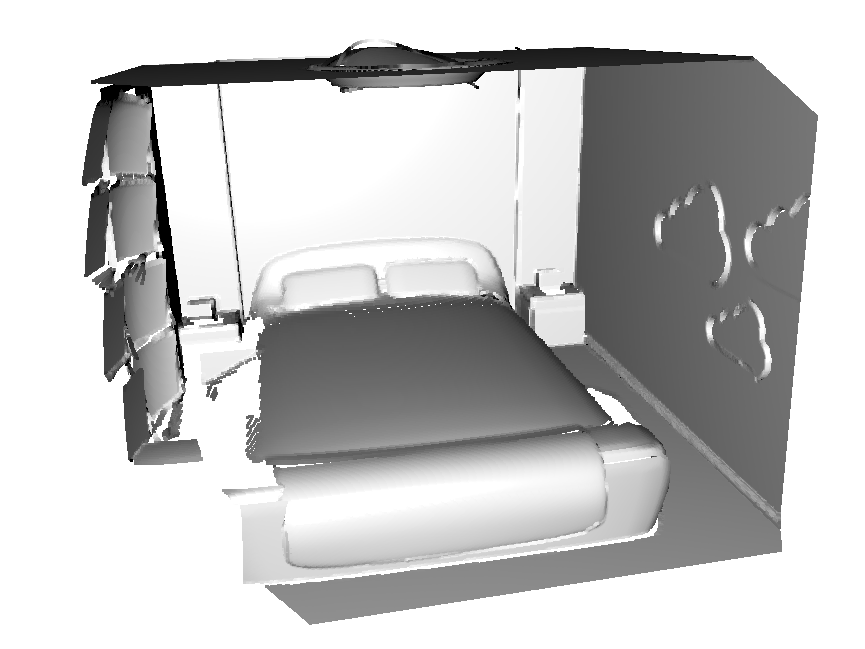} & 
\includegraphics[width=\linewidth]{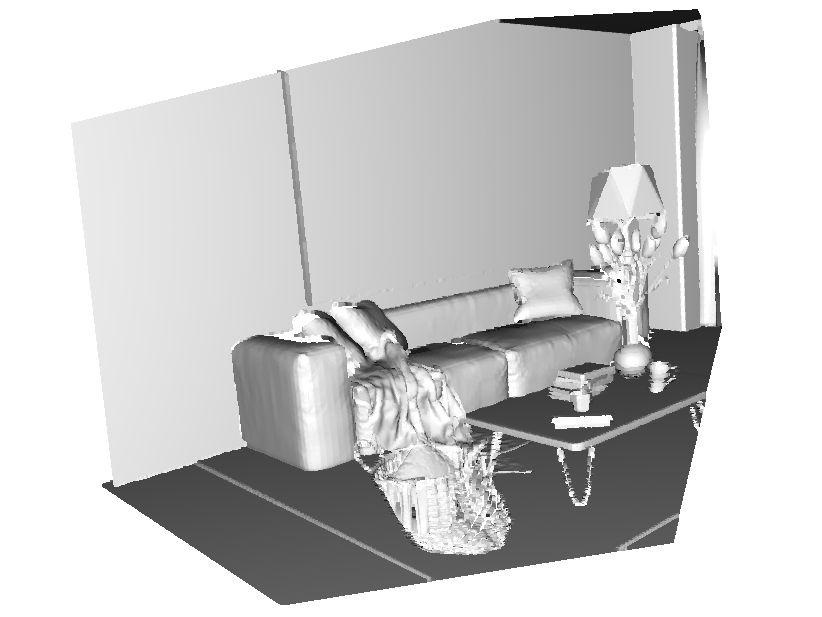} & &
\includegraphics[width=\linewidth]{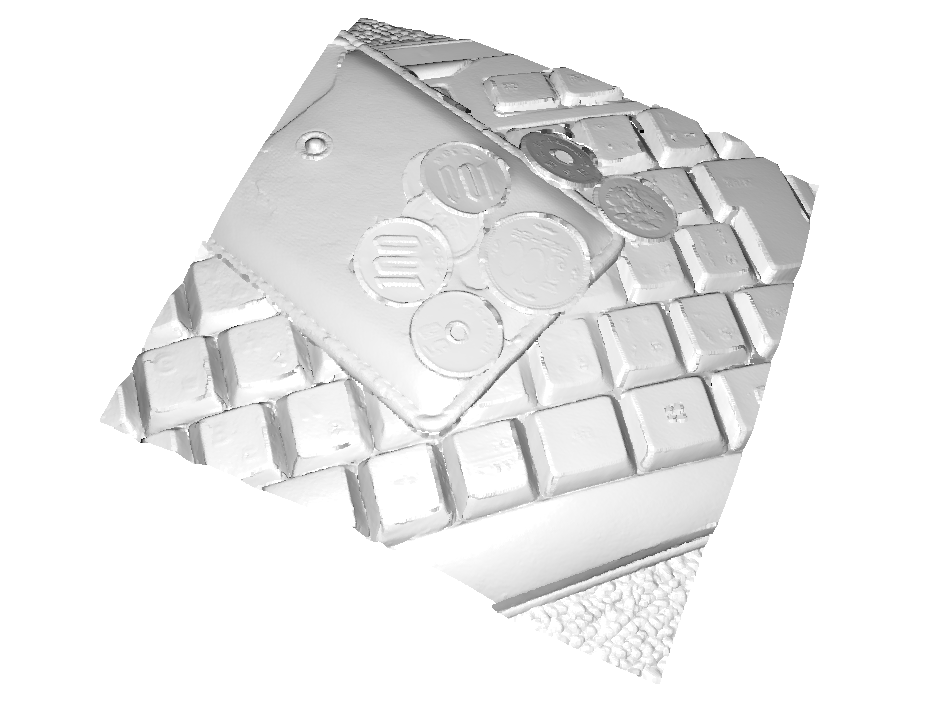} &
\includegraphics[width=\linewidth]{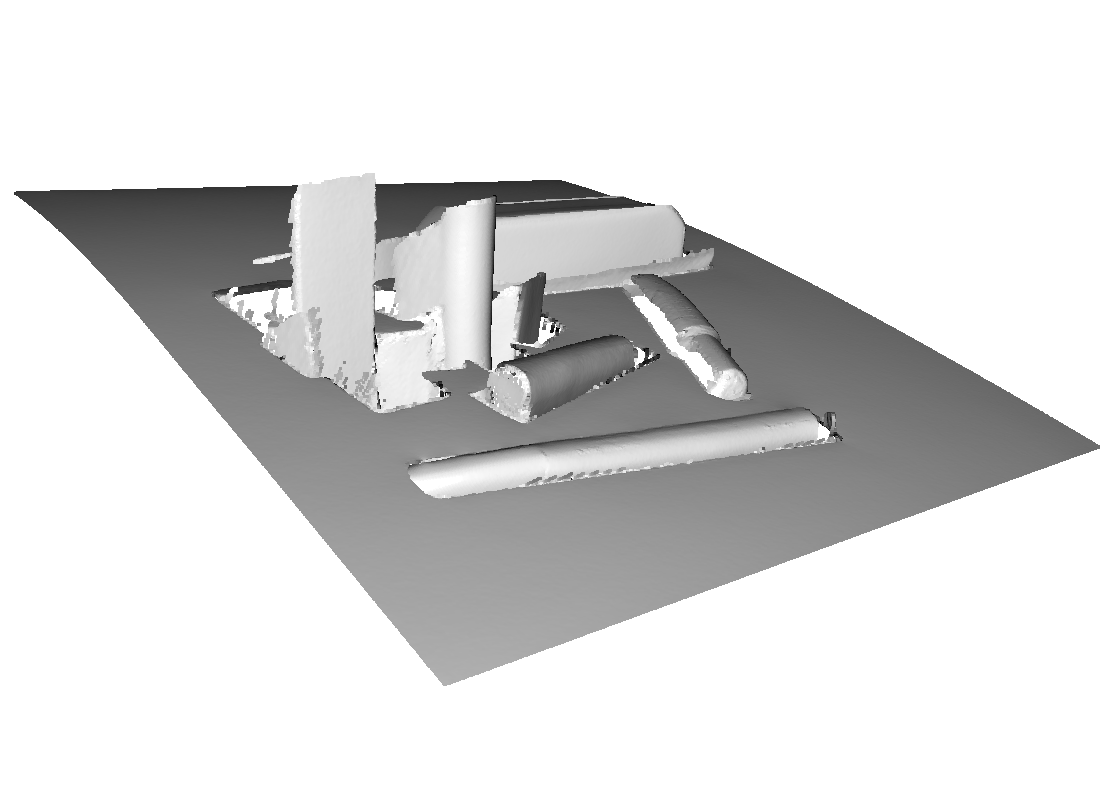} & &
\includegraphics[width=\linewidth]{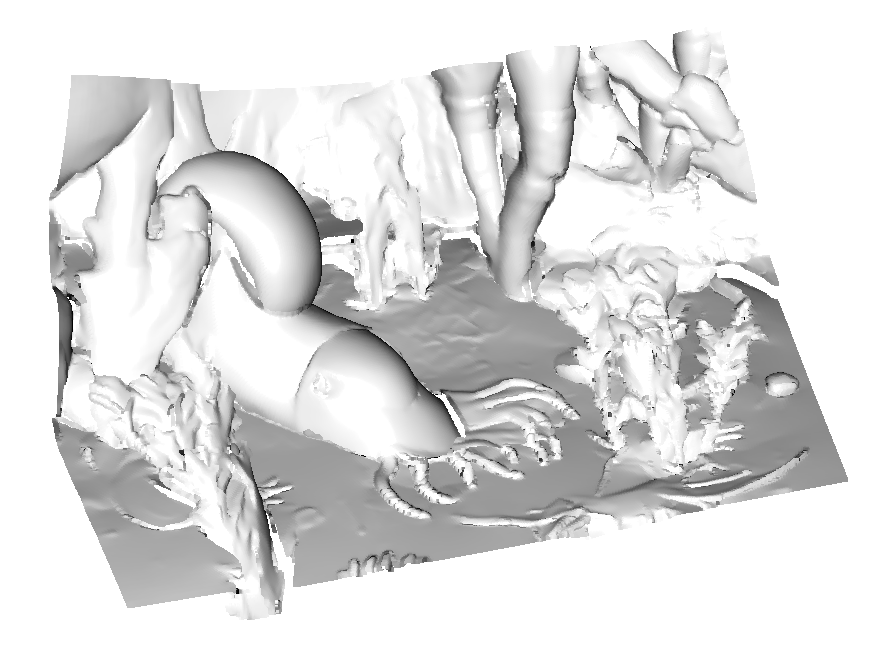} &
\includegraphics[width=\linewidth]{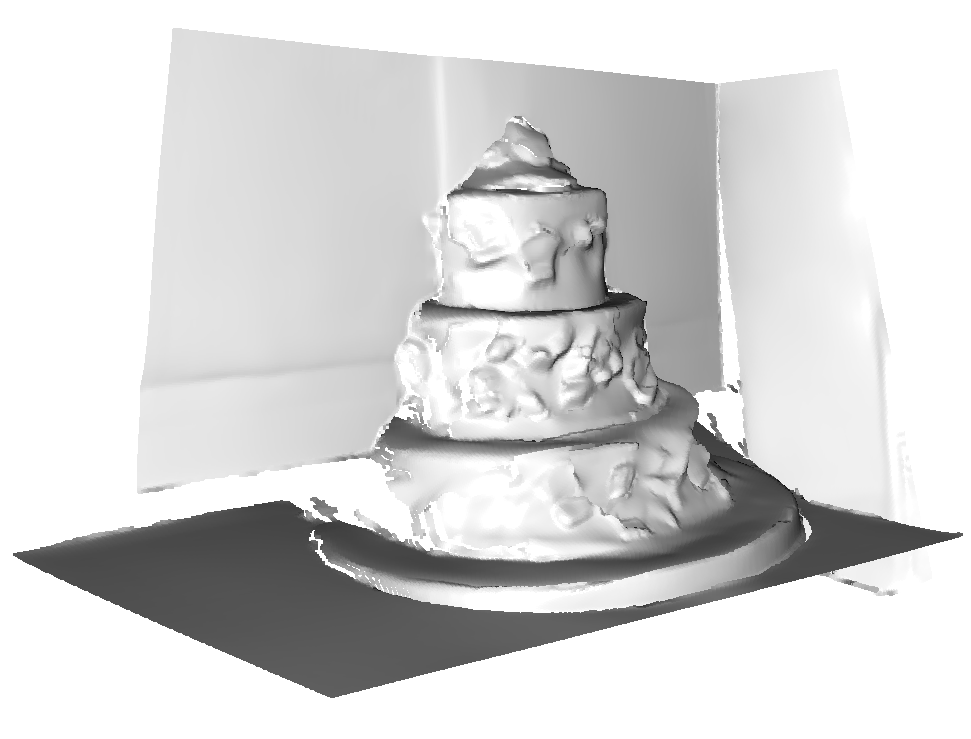}
\\[-5pt]
&
& \scriptsize{$t$: $\SI{6.8}{s}$, \textbf{RE}: $5.6\%$} & 
\scriptsize{$t$: $\SI{105.1}{s}$, \textbf{RE}: $10.6\%$} & 
& \scriptsize{$t$: $\SI{104.6}{s}$} & \scriptsize{$t$: $\SI{76.7}{s}$} & & \scriptsize{$t$: $\SI{58.5}{s}$} & \scriptsize{$t$: $\SI{13.4}{s}$}
\\
\noalign{\vskip -0.2em} 
\cline{3-10}
\noalign{\vskip 0.2em}
\multirow{5}{*}{\parbox[c][10ex][c]{\ht\strutbox}{\begin{turn}{90}
{Ours, pixel-level}
\end{turn}}} & \begin{turn}{90}
{\footnotesize Recon. ($\mathrm{M}$)}
\end{turn} &
\includegraphics[width=\linewidth]{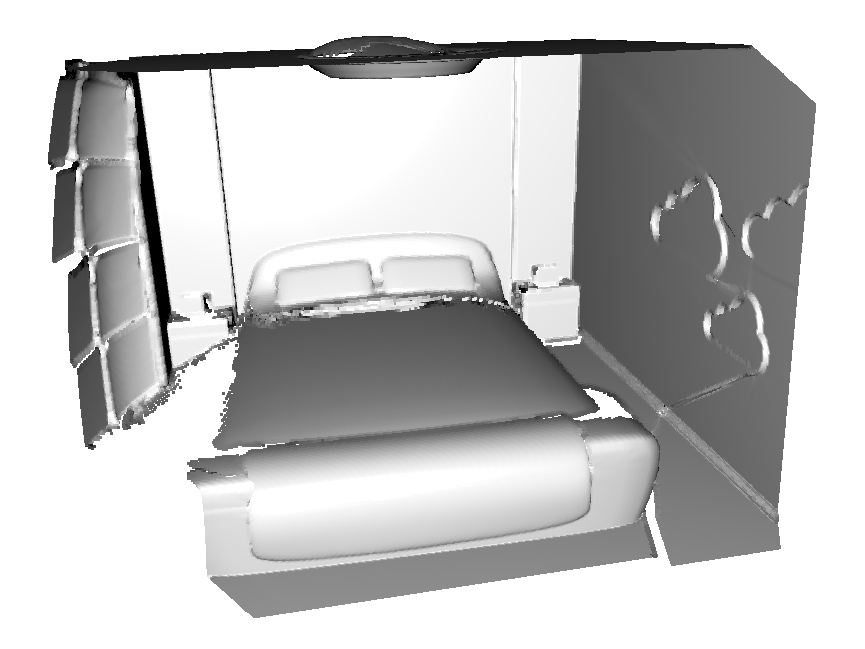} & 
\includegraphics[width=\linewidth]{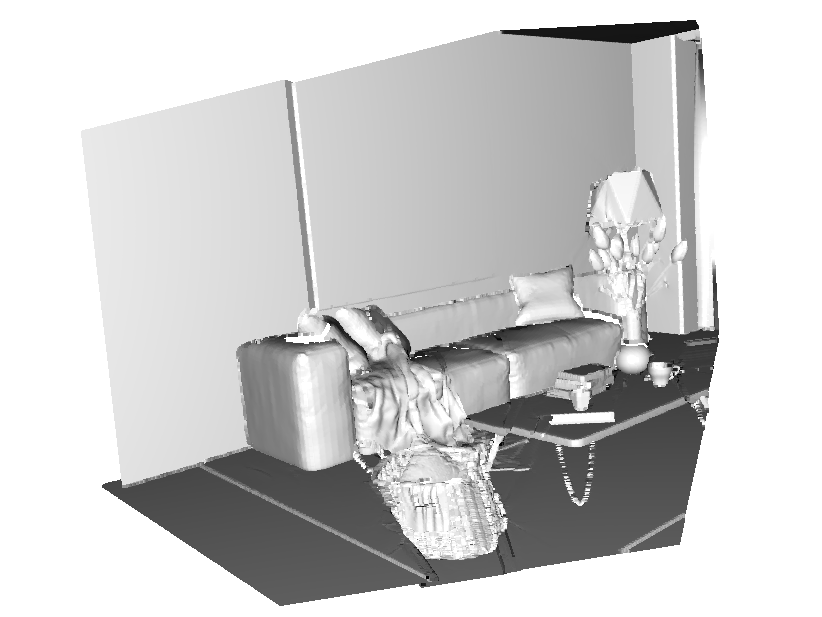} & &
\includegraphics[width=\linewidth]{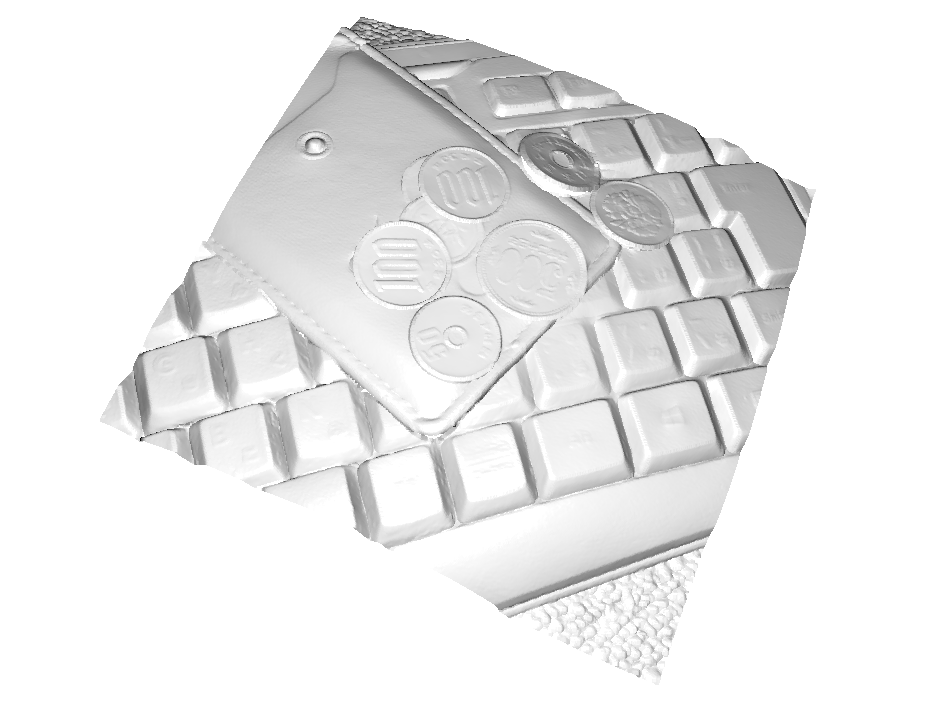} &
\includegraphics[width=\linewidth]{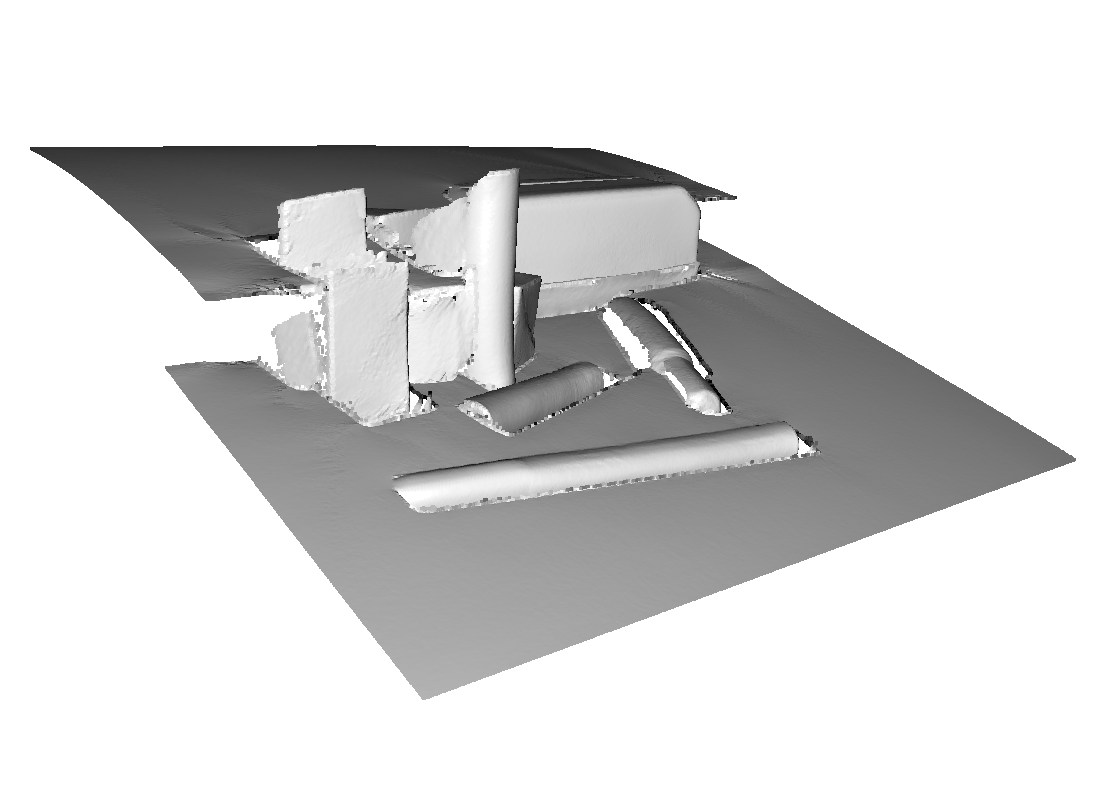} & &
\includegraphics[width=\linewidth]{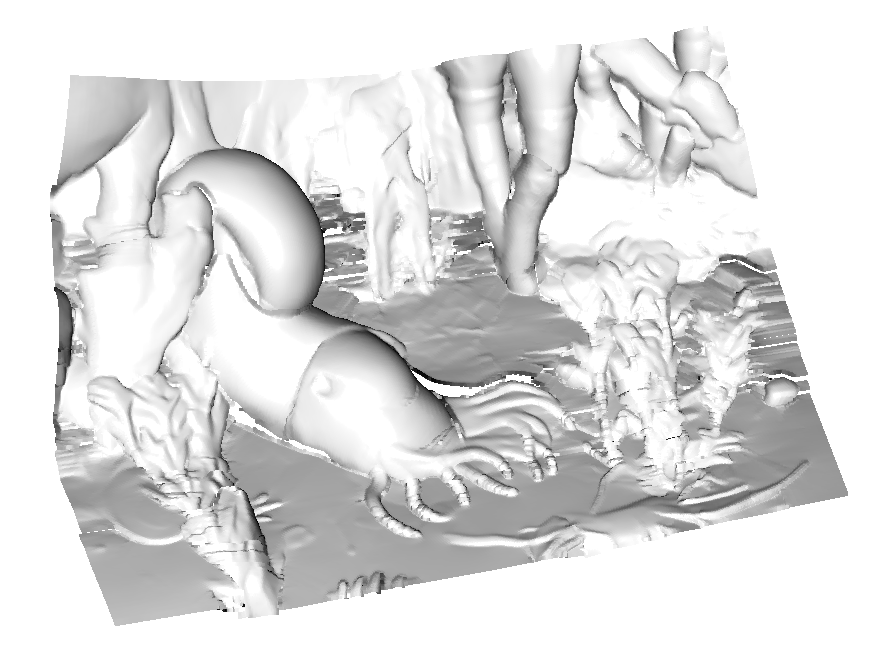} &
\includegraphics[width=\linewidth]{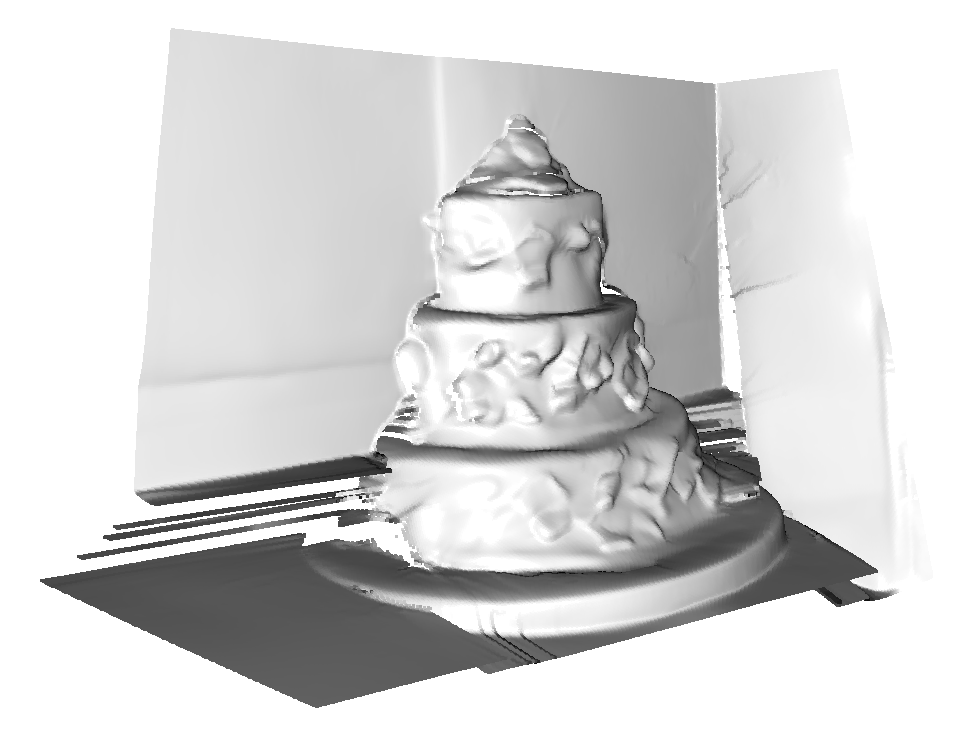}
\\[-5pt]
& & \scriptsize{$t$: $\SI{369.6}{s}$, \textbf{RE}: $8.8\%$} & \scriptsize{$t$:$\SI{1770.5}{s}$, \textbf{RE}: $14.6\%$} & & \scriptsize{$t$: $\SI{633.8}{s}$} & \scriptsize{$t$: $\SI{1016.8}{s}$} & & \scriptsize{$t$: $\SI{646.9}{s}$} & \scriptsize{$t$: $\SI{277.0}{s}$}
\\
& \begin{turn}{90}
{\footnotesize Recon. ($\mathrm{NM}$)}
\end{turn} &
\includegraphics[width=\linewidth]{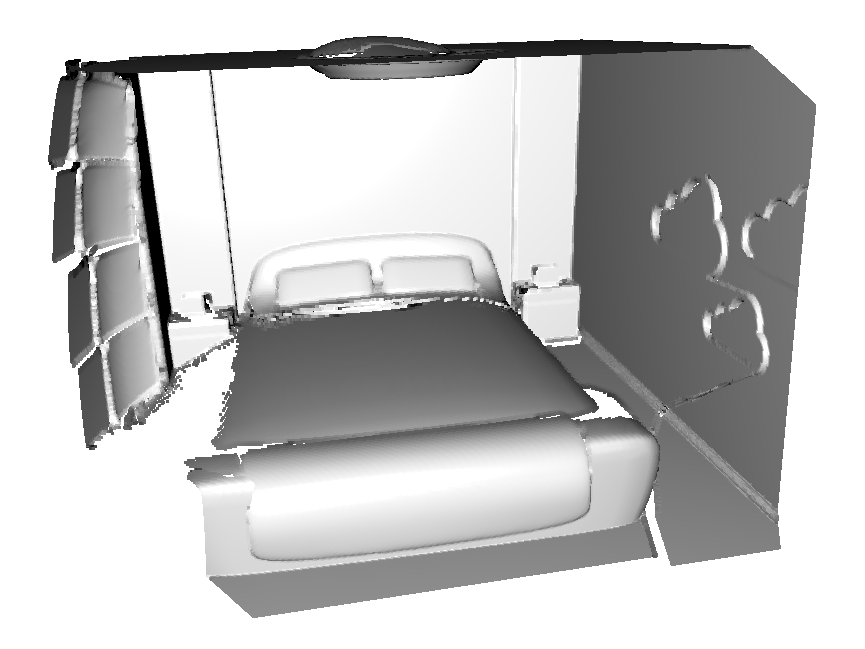} & 
\includegraphics[width=\linewidth]{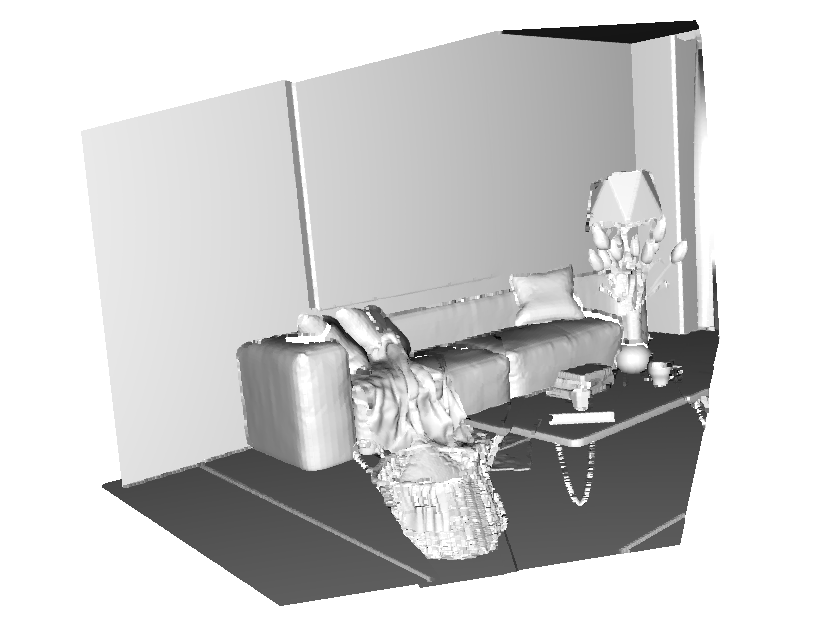} & &
\includegraphics[width=\linewidth]{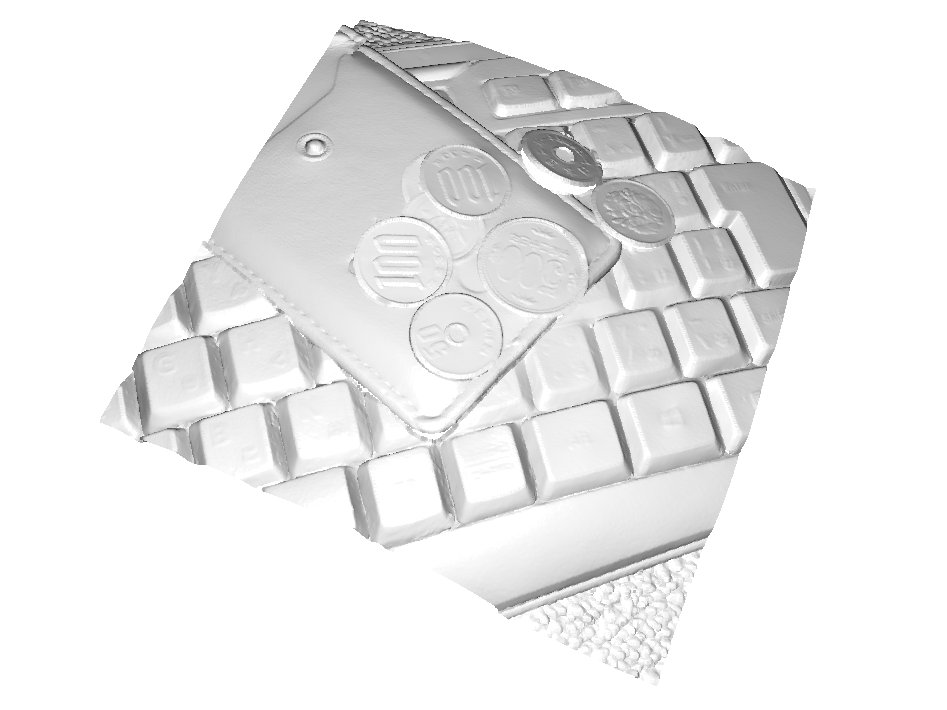} &
\includegraphics[width=\linewidth]{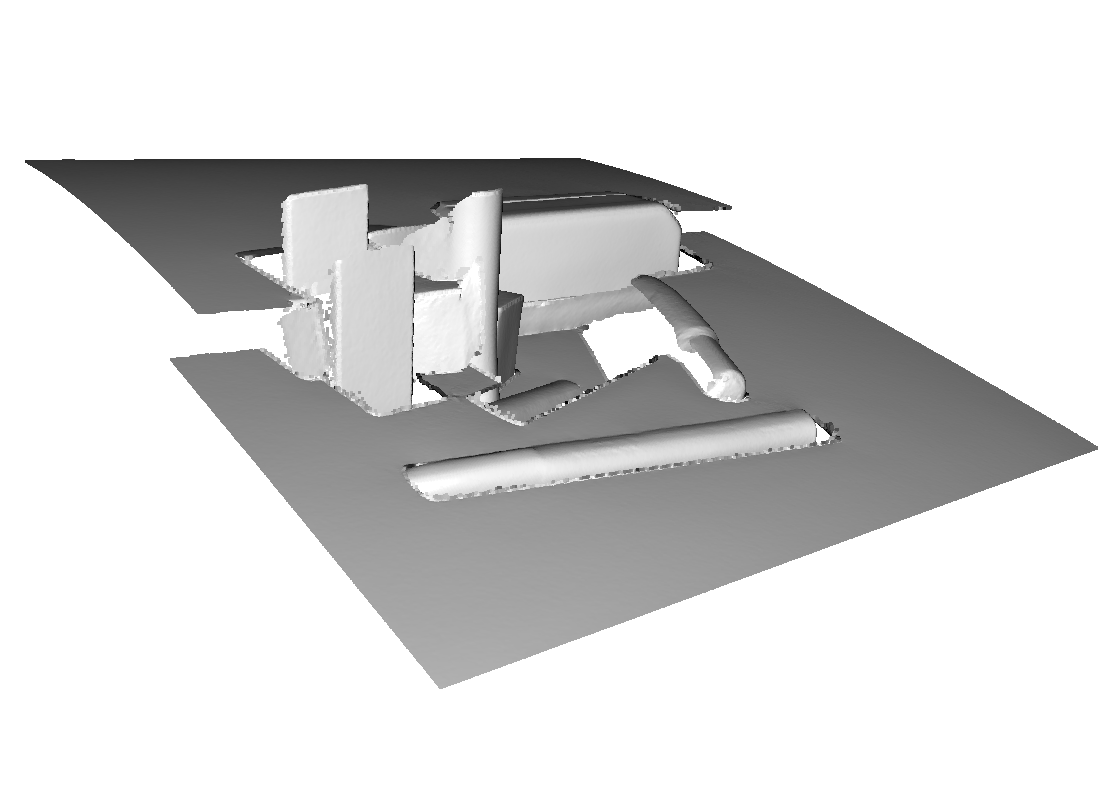} & &
\includegraphics[width=\linewidth]{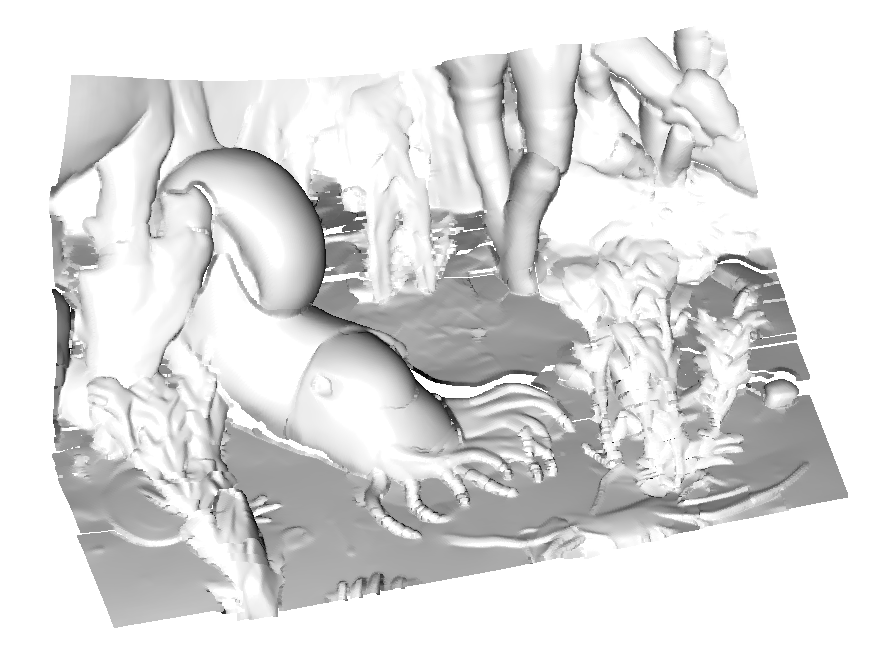} &
\includegraphics[width=\linewidth]{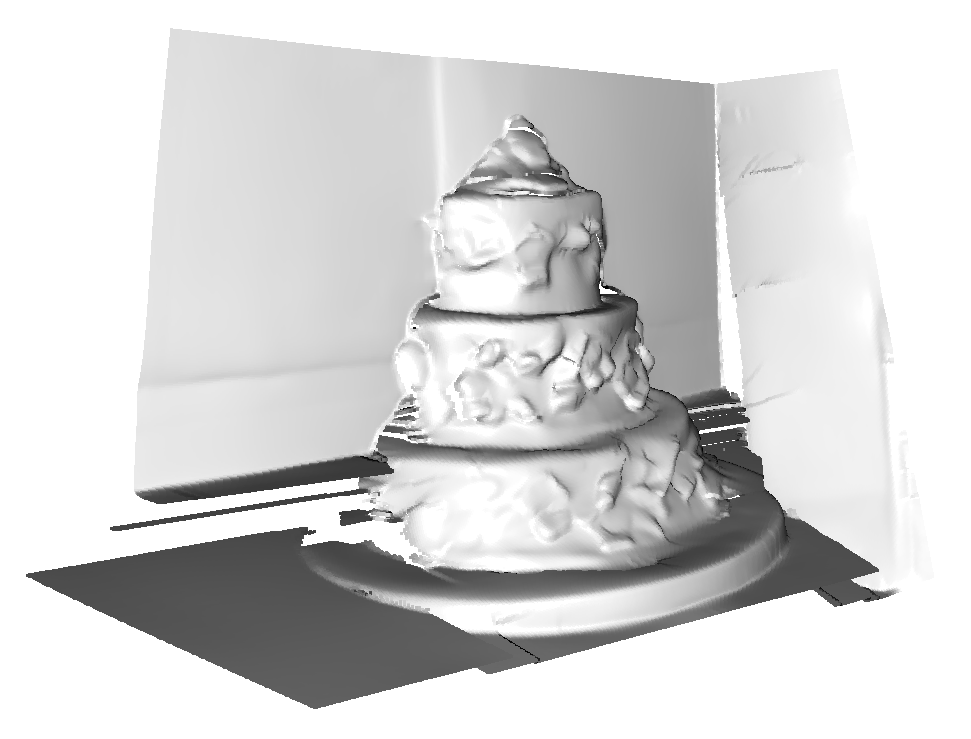}
\\[-5pt]
& & \scriptsize{$t$: $\SI{585.7}{s}$, \textbf{RE}: $8.9\%$} & \scriptsize{$t$:$\SI{2925.0}{s}$, \textbf{RE}: $14.8\%$} & & \scriptsize{$t$: $\SI{1453.3}{s}$} & \scriptsize{$t$: $\SI{2481.0}{s}$} & & \scriptsize{$t$: $\SI{1591.7}{s}$} & \scriptsize{$t$: $\SI{667.1}{s}$}\\
\noalign{\vskip -0.2em} 
\cline{3-10}
\noalign{\vskip 0.2em}
\begin{turn}{90}
{\cite{Milano2025DiscontinuityAwareNormalIntegration}}
\end{turn} & \begin{turn}{90}
{\footnotesize Reconstruction}
\end{turn} & \includegraphics[width=\linewidth]{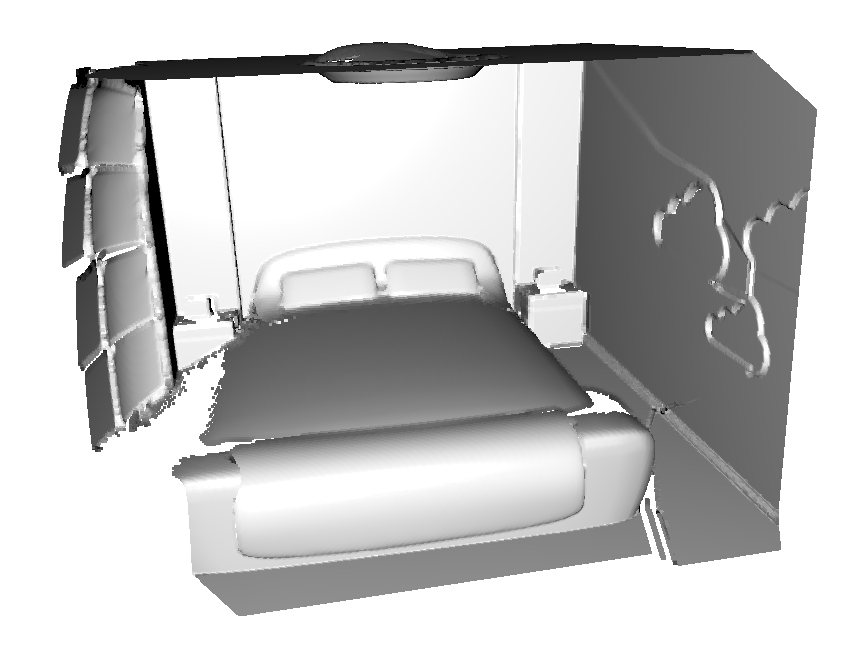} & 
\includegraphics[width=\linewidth]{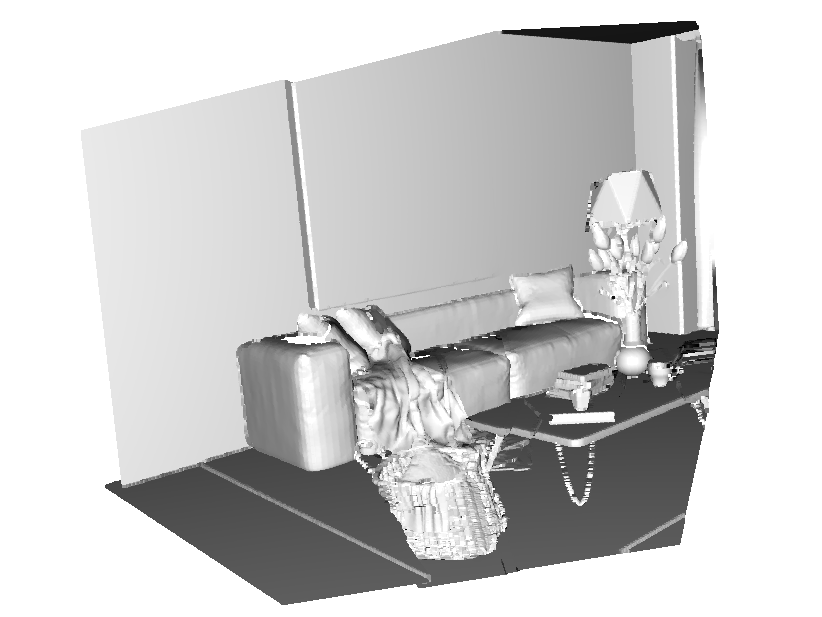} & &
\includegraphics[width=\linewidth]{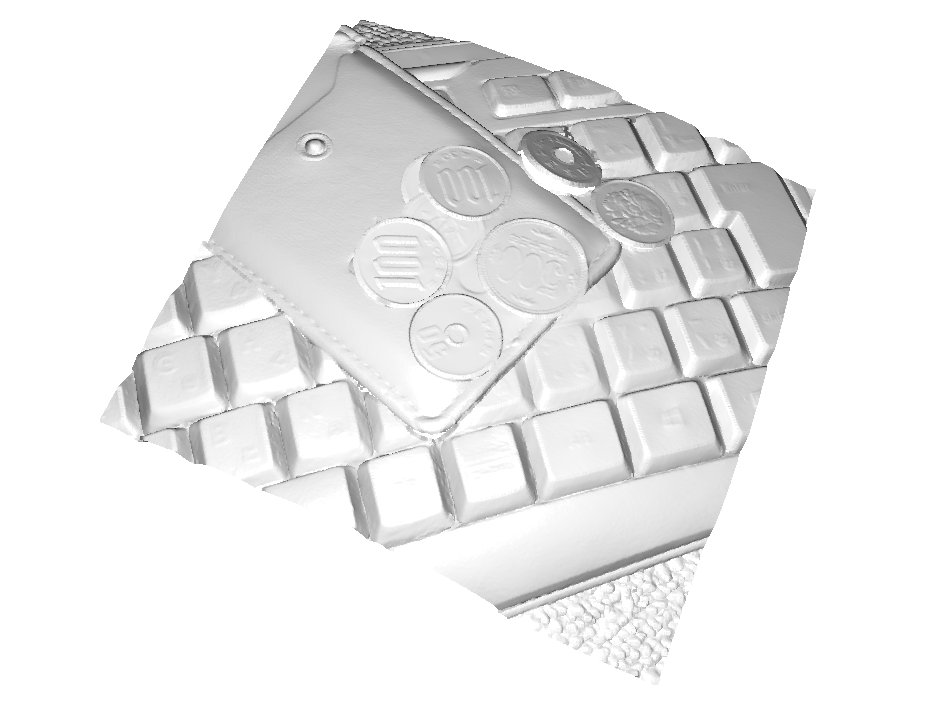} &
\includegraphics[width=\linewidth]{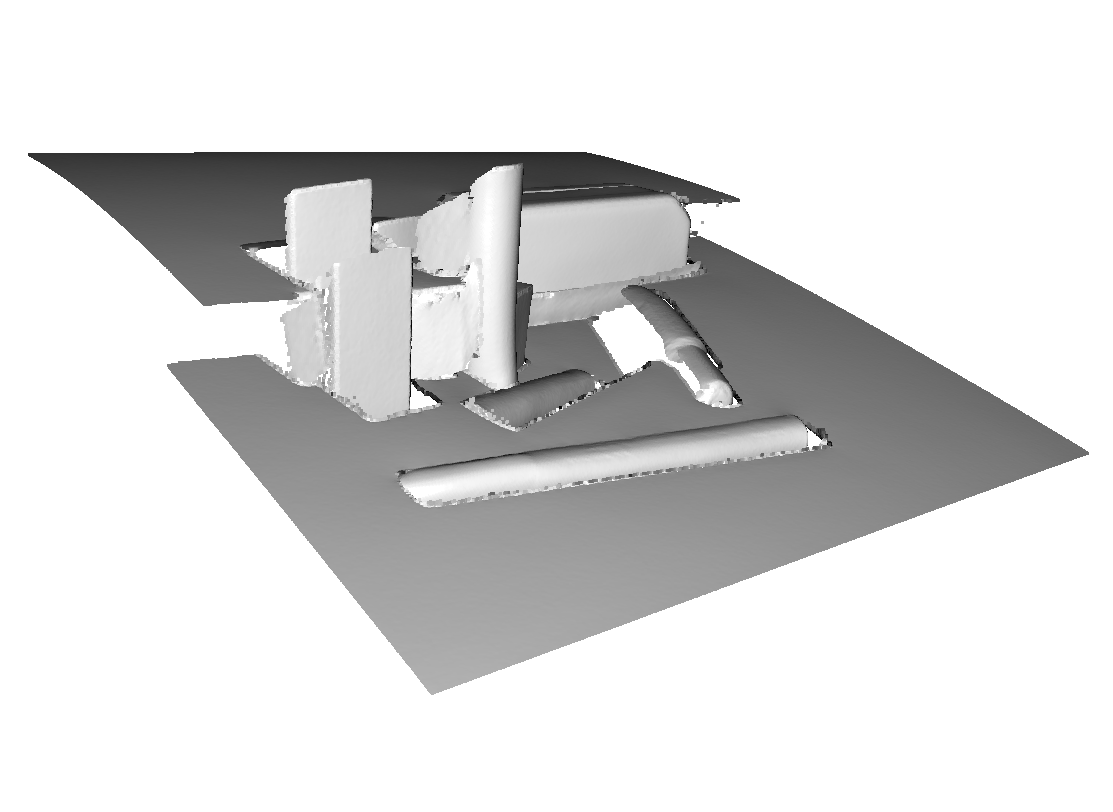} & &
\includegraphics[width=\linewidth]{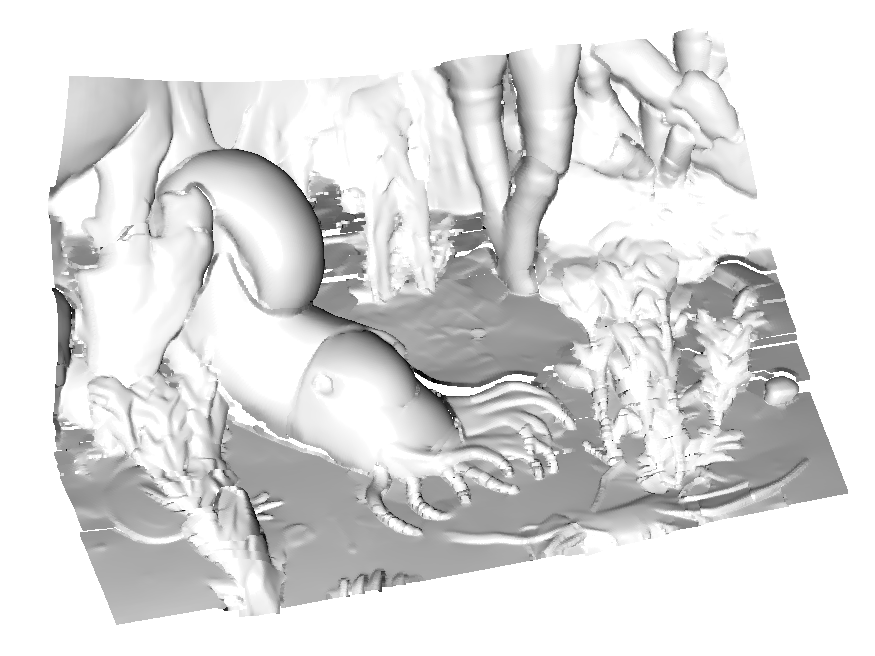} &
\includegraphics[width=\linewidth]{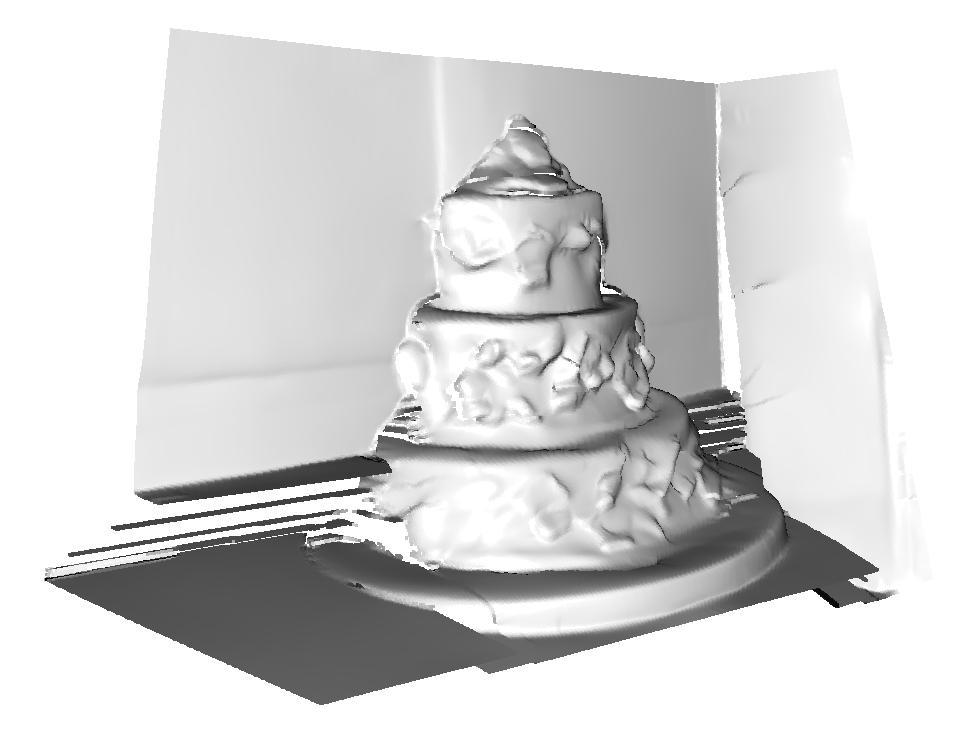}
\\[-5pt]
& & \scriptsize{$t$: $\SI{143.0}{s}$, \textbf{RE}: $11.8\%$} & \scriptsize{$t$: $\SI{992.1}{s}$, \textbf{RE}: $14.4\%$} & & \scriptsize{$t$: $\SI{1076.6}{s}$} & \scriptsize{$t$: $\SI{3068.9}{s}$} & & \scriptsize{$t$: $\SI{685.7}{s}$} & \scriptsize{$t$: $\SI{330.8}{s}$}\\
\end{tabular}
\addtolength{\tabcolsep}{4pt}
\caption{\textbf{Evaluation on large-scale normal maps.}
All experiments
use
$\Delta E_{\mathrm{max}}=10^{-3}$, $T=15$, outlier reweighting, and $\mathrm{freq}_{\mathrm{merging}}=5$ (where applicable). Total execution time $t$ and, where
ground-truth reconstruction is available, mean relative depth error (\textbf{RE}) are reported. 
$\mathrm{M}$ and $\mathrm{NM}$ denote 
our
version with and without merging, respectively. \textit{Sources of the normal maps:}
col.~$1$-$2$: rendered from meshes \cite{MeshCozyModernBedroom2023}, \cite{MeshCozyLivingRoomBaked2022};
col.~$3$-$4$: from photometric stereo~\cite{Ikehata2023SDMUniPS}, data from~\cite{Ikehata2023SDMUniPS};
col.~$5$-$6$: predicted by DSINE~\cite{Bae2024DSINE} on in-the-wild images \cite{Bae2024DSINE}, \cite{ImageWinterWeddingCake2007}.
}
\label{fig:comparison_high_resolution_normal_maps}
\end{figure*}
To assess
the
scalability of
component-based and pixel-level
formulations, we evaluate our method and the baseline on a set of mid-to-high-resolution normal maps that, unlike those from object-level datasets, have no masked-out pixels.
We
include normals rendered from meshes (using BlenderProc~\cite{Denninger2023BlenderProc}), real-world normals from photometric stereo~\cite{Ikehata2023SDMUniPS}, and normals predicted by the state-of-the-art normal estimation method DSINE~\cite{Bae2024DSINE} on in-the-wild images.
The number of pixels with a valid normal vector varies between $\num{311296}$ and $\num{1264640}$.
We run our method with merging enabled (with $\mathrm{freq}_{\mathrm{merging}}=5$) both in its component-based version (with $\theta_c=2.0^{\circ}$) and in its pixel-level variant, including for the latter also the version without merging. For all methods, we use outlier reweighting, with convergence criteria $\Delta E_{\mathrm{max}}=10^{-3}$ and $T=15$.

The results of our evaluation, together with the input normal maps and the output reconstructions, are shown in \cref{fig:comparison_high_resolution_normal_maps}.
Our component-based framework achieves convergence
on average one order of magnitude
faster
than
our pixel-level version and
the pixel-level method of~\cite{Milano2025DiscontinuityAwareNormalIntegration}. This result is a consequence of the smaller scale of
our
optimization problem. Using our normal similarity criterion, our method is able to detect large continuous regions in the input maps and thereby reduce the number of optimization variables by a factor between $5$ and $50$.
It follows that
while pixel-level methods require several minutes for convergence even at the same resolution used in the DiLiGenT dataset (\cf first column in \cref{fig:comparison_high_resolution_normal_maps}), our method converges in a
few seconds for the same
input
and in less than $2$ minutes for the larger resolutions.
While the log-depth-based formulation of~\cite{Milano2025DiscontinuityAwareNormalIntegration} converges on average faster than our \emph{pixel-level} variant under the same conditions, we note that, as opposed to the small-scale DiLiGenT normal maps, enabling merging for our version on
large-scale
normal maps results in significantly reduced runtime with little to no impact on the reconstruction. This highlights that merging can be particularly beneficial in settings where iterations with the full number of pixels are costly and therefore a reduction in the number of variables is desirable.

Our
component-based approach 
achieves
more accurate reconstructions
than pixel-level methods
also for these large-scale normal maps.
Importantly,
we
note
that
our identification and separate optimization of
components
allows more accurately reconstructing particularly
large
continuous
regions,
for which the global optimization of pixel-level methods
tends to introduce
spurious discontinuities
(\cref{fig:comparison_high_resolution_normal_maps}: floor in columns $1, 4$ and wall in column $6$).
We note however that, for
a similar
reason,
due to the model of discontinuity, our method
may occasionally separate continuous surface regions from one another if these are assigned to different components, particularly to background
ones
(\cref{fig:comparison_high_resolution_normal_maps}: tentacles of the octopus in column $5$).
We discuss this and other limitations in more detail in \cref{sec_suppl:limitations}.
\section{Conclusions\label{sec:conclusions}}
We proposed a framework for fast and scalable normal integration by reframing the problem as the optimization of the relative scales of continuous surface components. 
Our approach achieves state-of-the-art accuracy on the standard normal integration benchmark and enables scaling normal integration to large-resolution normal maps, with an order of magnitude reduction in the execution time while preserving discontinuities.

\clearpage
{
    \small
    \bibliographystyle{ieeenat_fullname}
    \bibliography{main}
}

\cleardoublepage
\appendix
\section*{Supplementary Material}
The following Sections constitute the Supplementary Material. \Cref{sec_suppl:pseudocode} provides a pseudocode of our method. In \cref{sec_suppl:detailed_profiling_runtime}, a detailed profiling of our run time is
presented.
\Cref{sec_suppl:outlier_reweighting} investigates the effect of different hyperparameters for our outlier
reweighting.
\Cref{sec_suppl:ablation_convergence_parameters} studies the impact of the convergence
parameters.
\Cref{sec_suppl:visualization_detected_components_diligent} provides visualizations of
our
continuous components
on the DiLiGenT dataset.
In \cref{sec_suppl:ablation_effect_merging_diligent} we analyze the impact of our
merging operation on
reconstruction error on the DiLiGenT dataset. 
\Cref{sec_suppl:impact_connectivity} provides an ablation on the
pixel connectivity used by our method. 
Finally, in \cref{sec_suppl:limitations} we discuss the limitations of our method.

\section{Pseudocode of our method~\label{sec_suppl:pseudocode}}
A pseudocode for our method is shown in \cref{alg:algorithm_pseudocode}.
\begin{algorithm}[H]
  \caption{Pseudocode of our method.}
  \label{alg:algorithm_pseudocode}
  \begin{algorithmic}[1]
  \Require $\theta_c$, $\mathrm{can\ merge}$ (bool), $\mathrm{freq}_{\mathrm{merging}}$, $\Delta E_{\mathrm{max}}$, $T$.
    \State Initialize $\mathbf{\tilde{z}} \gets \mathbf{0}$
    \State Form components
    $\{\mathcal{C}^{(0)}_c\}$
    based on $\theta_{a, b} < \theta_c$
    \State Compute intra-component matrices
    \Statex $\mathbf{A_c}$, $\mathbf{b_c}$, $\mathbf{W_c}=\mathrm{diag}\left(\{W_{b\rightarrow a}\}_{(a, b)\in E(\mathcal{C}^{(0)}_c)}\right)$~\eqref{eq:intracomponent_matrices}
    \State Fill each component $\mathcal{C}_c$ in parallel: 
    \Statex $\mathbf{\tilde{z}_c}^{(0)} \gets \texttt{cg}\left(\mathbf{A_c}^\mathsf{T}\mathbf{W_c}\mathbf{A_c}, \mathbf{A_c}^\mathsf{T}\mathbf{W_c}\mathbf{b_c}\right)$
    \State $\mathrm{converged} \gets \texttt{False}$, $m\gets 0$, $t\gets 0$, $E_0\gets\epsilon$
    \State Form inter-component matrices
        $\mathbf{\overline{A}}_{0}$, $\mathbf{\overline{b}}_{0}$~\eqref{eq:intercomponent_matrices}
    \While{not $\mathrm{converged}$}
    \Comment{Relative-scale optimization}
        \If{$t\le 1$}
            \Comment{Alignment optimization}
            \State Uniform weights:
            $\mathbf{\overline{W}}^{(t)}_{m}\gets \mathrm{diag}(1)$
        \Else
            \Comment{Discontinuity-aware optimization}
            \State BiNI weights with outlier reweighting~\eqref{eq:outlier_reweighting}:
            \Statex \hspace{\algorithmicindent} \hspace{\algorithmicindent} $\mathbf{\overline{W}}^{(t)}_{m}\gets \mathrm{diag}\left(\{W_{b\rightarrow a}\cdot W_{b\rightarrow a}^{\mathrm{out}}\}_{(a, b)\in E^{(m)}_{\mathrm{inter}}}\right)$
        \EndIf
        \State $\mathbf{\tilde{s}}^{(t)}\gets\texttt{cg}\left(\mathbf{\overline{A}}_m^{\mathsf{T}}\mathbf{\overline{W}}^{(t)}_{m}\mathbf{\overline{A}}_m, \mathbf{\overline{A}}_m^{\mathsf{T}}\mathbf{\overline{W}}^{(t)}_{m}\mathbf{\overline{b}}_m\right)$
        \State Scale components (in parallel, via broadcasting):
        \Statex \hspace{\algorithmicindent} $\forall c\in\{0, \dots, |C^{(m)}| -1\}$,\ $\mathbf{\tilde{z}_c}^{(t+1)}\gets\mathbf{\tilde{z}_c}^{(t)} + \tilde{s}_c^{(t)}\mathbf{1}$
        \State $t\gets t +1$
        \If{$\mathrm{can\ merge}\ \land\ (t \equiv 0\ \left(\mathrm{mod }\ \mathrm{freq}_{\mathrm{merging}})\right)$}
            \State $\forall \mathcal{C}^{(m)}_c$, $(\hat{a},\hat{b})^{(m)}_c\gets\arg\min_{(a,b)\in\partial\mathcal{C}^{(m)}_{c}} |\overline{\chi}_{b\rightarrow a}|$
            \State Compute subgraph $\hat{\mathcal{Q}}_m\gets(C^{(m)}, \{(\hat{a},\hat{b})^{(m)}_c\})$
            \State $\{\mathcal{C}^{(m+1)}_c\}\gets\mathrm{connected\ components}(\hat{\mathcal{Q}}_m)$
            \State $m\gets m+1$
            \State Form inter-component matrices
        $\mathbf{\overline{A}}_{m}$, $\mathbf{\overline{b}}_{m}$~\eqref{eq:intercomponent_matrices}
        \EndIf
        \State $E_t\gets (\mathbf{\overline{A}}_m\mathbf{\tilde{s}}^{(t)} - \mathbf{\overline{b}}_m )^{\mathsf{T}}\mathbf{\overline{W}}^{(t)}_{m}(\mathbf{\overline{A}}_m\mathbf{\tilde{s}}^{(t)} - \mathbf{\overline{b}}_m)$
        \State $\mathrm{converged} \gets \frac{|E_t - E_{t-1}|}{E_{t-1}} < \Delta E_{\mathrm{max}} \lor t = T$
    \EndWhile
    \State \Return $\mathbf{\tilde{z}}$
  \end{algorithmic}
\end{algorithm}

\section{Detailed profiling of our run time~\label{sec_suppl:detailed_profiling_runtime}}
\begin{table*}[!ht]
    \centering
    \resizebox{\linewidth}{!}{
    \begin{tabular}{l ccccccccc}
    \toprule
     & \texttt{bear} & \texttt{buddha} & \texttt{cat} & \texttt{cow} & \texttt{harvest} & \texttt{pot1} & \texttt{pot2} & \texttt{reading} & \texttt{goblet} \\
    \midrule
    Formation of $\mathcal{G}_0$ & $0.020$ & $0.018$ & $0.019$ & $0.020$ & $0.019$ & $0.018$ & $0.019$ & $0.019$ & $0.020$ \\
    Computation of $\{\mathcal{C}_c^{(0)}\}$ & $0.092$ & $0.090$ & $0.091$ & $0.087$ & $0.091$ & $0.097$ & $0.089$ & $0.088$ & $0.094$ \\
    Formation of $\mathbf{A_c}$, $\mathbf{b_c}$, $\mathbf{W_c}$ & $0.546$ & $1.857$ & $0.731$ &  $0.758$ & $3.345$ & $1.581$ & $1.384$ & $1.206$ & $0.344$ \\
    Filling of components & $1.168$ & $2.442$ & $1.428$ & $1.360$ & $4.166$ & $2.325$ & $2.103$ & $1.719$ & $0.745$ \\
    Iteration $0$ & $1.174$ & $2.587$ & $1.439$ & $1.368$ & $4.311$ & $2.342$ & $2.118$ & $1.734$ & $0.775$ \\
    Iteration $1$ & $1.183$ & $2.977$ & $1.464$ & $1.378$ & $4.810$ & $2.394$ & $2.148$ & $1.766$ & $0.840$\\
    Iteration $2$ & $1.193$ & $3.490$ & $1.484$ & $1.391$ & $5.747$ & $2.471$ & $2.187$ & $1.822$ & $0.858$ \\
    $\vdots$ & $\vdots$ & $\vdots$ & $\vdots$ & $\vdots$ & $\vdots$ & $\vdots$ & $\vdots$ & $\vdots$ & $\vdots$ \\
    Convergence & $1.270$ [it. $8$] & $8.033$ [it. $11$] & $1.500$ [it. $3$] & $1.525$ [it. $12$] & $18.805$ [it. $17$] & $3.511$ [it. $18$] & $2.655$ [it. $17$] & $2.679$ [it. $20$] & $1.398$ [it. $49$] \\
    \bottomrule
    \end{tabular}
    }
    \caption{\textbf{Profiling of our method on the the DiLiGenT benchmark~\cite{Shi2016DiLiGenT}.} Intermediate execution times from the start, after the completion of each step are reported [$\si{s}$]. $\theta_c=3.5^{\circ}$, $\Delta E_{\mathrm{max}}=10^{-3}$, $T=150$, $8-$connectivity, and outlier reweighting are used, without merging.}
    \label{tab:profiling_diligent}
\end{table*}
\begin{table*}[!ht]
    \centering
    \resizebox{0.8\linewidth}{!}{
    \begin{tabular}{l cccccc}
    \toprule
     & \texttt{bedroom} & \texttt{livingroom} & \texttt{coinskeyboard} & \texttt{schooldesk} & \texttt{seafloor} & \texttt{cake} \\
    \midrule
    Formation of $\mathcal{G}_0$ & $0.019$ & $0.058$ & $0.058$ & $0.062$ & $0.050$ & $0.020$ \\
    Computation of $\{\mathcal{C}_c^{(0)}\}$ & $0.138$ & $0.522$ & $0.375$ & $0.447$ & $0.395$ & $0.147$ \\
    Formation of $\mathbf{A_c}$, $\mathbf{b_c}$, $\mathbf{W_c}$ & $1.554$ & $12.226$ & $42.058$ & $5.759$ & $15.663$ & $4.486$ \\
    Filling of components & $3.804$ & $50.536$ & $51.855$ & $69.922$ & $29.548$ & $8.240$ \\
    Iteration $0$ & $3.955$ & $51.232$ & $54.343$ & $70.075$ & $29.919$ & $8.382$ \\
    Iteration $1$ & $4.290$ & $55.196$ & $60.506$ & $70.448$ & $31.546$ & $8.890$ \\
    $\vdots$ & $\vdots$ & $\vdots$ & $\vdots$ & $\vdots$ & $\vdots$ & $\vdots$ \\
    Merging $m=0$ & $5.876$ & $92.018$ & $96.427$ & $75.205$ & $51.690$ & $12.382$ \\
    Merging $m=1$ & $6.198$ & $102.327$ & $102.591$ & $75.783$ & $57.139$ & $12.883$ \\
    $\vdots$ & $\vdots$ & $\vdots$ & $\vdots$ & $\vdots$ & $\vdots$ & $\vdots$ \\
    Convergence & $6.796$ & $105.081$ & $104.630$ & $76.648$ & $58.524$ & $13.374$ \\
    \bottomrule
    \end{tabular}
    }
    \caption{\textbf{Profiling of our method on the the large-scale normal maps from \cref{fig:comparison_high_resolution_normal_maps}.} Columns $1$ to $6$ in \cref{fig:comparison_high_resolution_normal_maps} are referred to, from left to right, as: \texttt{bedroom}, \texttt{livingroom}, \texttt{coinskeyboard}, \texttt{schooldesk}, \texttt{seafloor}, \texttt{cake}. Intermediate execution times from the start, after the completion of each step are reported [$\si{s}$]. $\theta_c=2.0^{\circ}$, $\Delta E_{\mathrm{max}}=10^{-3}$, $T=15$, $8-$connectivity, $\mathrm{freq}_{\mathrm{merging}}=5$, and outlier reweighting are used.}
    \label{tab:profiling_large_scale_normal_maps}
\end{table*}
We provide a detailed profiling of the run time of our method for the parameter configuration used in our main experiments, both on the DiLiGenT dataset (\cref{tab:profiling_diligent}) and on the large-scale normals maps of \cref{fig:comparison_high_resolution_normal_maps} (\cref{tab:profiling_large_scale_normal_maps}).

It is worth noting that, in both cases, two operations that account for a significant fraction of the total execution time are the two pre-processing steps of forming the intra-component matrices $\mathbf{A_c}$, $\mathbf{b_c}$, and $\mathbf{W_c}$ and filling the components.
For the latter, we use the Python \texttt{joblib} library to parallelly execute multiple instances of per-component conjugate-gradient optimization. While this usually converges in few fractions of a second due to the smaller scale of the per-component optimization compared to the global optimization, larger resolutions might produce larger components, resulting in increased time for this initial step. Additionally, parallelization is capped by the number of processes that are available to the program (we set this to $4$ in our experiments), thereby still requiring
iterative processing. On the other hand, the formation step of the intra-component matrices $\mathbf{A_c}$, $\mathbf{b_c}$, and $\mathbf{W_c}$ is non-optimized in our current implementation, and alone contributes to a factor of up to respectively $39\%$ (\texttt{coinskeyboard}) and $44\%$ (\texttt{cow}) of the total execution time for the large-scale normal maps and the smaller-scale DiLiGenT dataset. The reason for the long time required to perform this step lies in the fact that the matrices $\mathbf{A_c}$ and $\mathbf{W_c}$ are represented in our implementation
as sparse matrices
(\texttt{scipy.sparse.csr\_matrix}) of different shape, and as such cannot benefit from parallel-access optimized broadcasting operations. Implementation improvements in this direction are left to future work.

\section{Ablation on outlier reweighting~\label{sec_suppl:outlier_reweighting}}
\begin{table*}[!ht]
    \centering
    \resizebox{0.8\linewidth}{!}{
    \begin{tabular}{lll c c c c c c c c c}
    \toprule
    \multirow{1}{*}{Reweighting type} & \multirow{1}{*}{$U$}  & \multirow{1}{*}{$\theta_c$} & \multicolumn{1}{c}{\texttt{bear}} & \multicolumn{1}{c}{\texttt{buddha}} & \multicolumn{1}{c}{\texttt{cat}} & \multicolumn{1}{c}{\texttt{cow}} & \multicolumn{1}{c}{\texttt{harvest}} & \multicolumn{1}{c}{\texttt{pot1}} & \multicolumn{1}{c}{\texttt{pot2}} & \multicolumn{1}{c}{\texttt{reading}} & \multicolumn{1}{c}{\texttt{goblet}} \\
    \midrule
    \multirow{4}{*}{Soft} & \multirow{4}{*}{$10^{-5}$} & None & $0.02$ & $0.13$ & $0.03$ & $0.09$ & $1.35$ & $0.40$ & $0.14$ & $0.19$ & $9.37$ \\ 
    & & $2.0^{\circ}$ & $0.02$ & $0.17$ & $0.03$ & $0.09$ & $1.02$ & $0.37$ & $0.14$ & $0.20$ & $9.35$ \\ 
    & & $3.5^{\circ}$ & $0.02$ & $0.10$ & $0.04$ & $0.10$ & $1.10$ & $0.39$ & $0.14$ & $0.10$ & $8.08$ \\ 
    & & $5.0^{\circ}$ & $0.02$ & $0.15$ & $0.51$ & $0.39$ & $1.61$ & $0.55$ & $0.14$ & $0.17$ & $4.45$ \\ 
    \arrayrulecolor{gray!70}\specialrule{0.2pt}{0.2pt}{0.2pt}
    \arrayrulecolor{black}
    \multirow{4}{*}{Soft} & \multirow{4}{*}{$10^{-4}$} & None & $0.02$ & $0.20$ & $0.03$ & $0.09$ & $1.31$ & $0.36$ & $0.13$ & $0.17$ & $9.41$ \\ 
    & & $2.0^{\circ}$  & $0.02$ & $0.17$ & $0.03$ & $0.09$ & $1.04$ & $0.36$ & $0.14$ & $0.10$ & $9.37$ \\ 
    & & $3.5^{\circ}$ & $0.02$ & $0.11$ & ${0.04}$ & $0.09$ & ${1.07}$ & $0.38$ & ${0.14}$ & $0.09$ & $9.49$ \\ 
    & & $5.0^{\circ}$ &  ${0.02}$ & ${0.15}$ & $0.51$ & $0.39$ & $1.75$ & $0.55$ & ${0.13}$ & $0.16$ & $9.62$ \\ 
    \arrayrulecolor{gray!70}\specialrule{0.2pt}{0.2pt}{0.2pt}
    \arrayrulecolor{black}
    \multirow{4}{*}{Soft} & \multirow{4}{*}{$10^{-6}$} & None & $0.02$ & $0.17$ & $0.03$ & $0.09$ & $0.99$ & $0.36$ & $0.13$ & $0.13$ & $9.41$ \\ 
    & & $2.0^{\circ}$ & $0.02$ & $0.13$ & $0.03$ & $0.09$ & $0.88$ & $0.36$ & $0.14$ & $0.10$ & $9.40$ \\ 
    & & $3.5^{\circ}$ & $0.02$ & $0.17$ & $0.04$ & $0.09$ & $1.18$ & $0.38$ & $0.13$ & $0.10$ & $7.53$ \\ 
    & & $5.0^{\circ}$ & $0.02$ & $0.18$ & $0.51$ & $0.39$ & $1.70$ & $0.55$ & $0.13$ & $0.15$ & $9.62$ \\ 
    \arrayrulecolor{gray!70}\specialrule{0.2pt}{0.2pt}{0.2pt}
    \arrayrulecolor{black}
    \multirow{4}{*}{Hard} & \multirow{4}{*}{N/A} & None & $0.03$ & $0.27$ & $0.05$ & $0.09$ & $1.78$ & $0.41$ & $0.13$ & $0.19$ & $9.33$ \\ 
    & & $2.0^{\circ}$ & $0.02$ & $0.42$ & $0.05$ & $0.09$ & $1.13$ & $0.39$ & $0.14$ & $0.22$ & $8.19$ \\ 
    & & $3.5^{\circ}$ & $0.02$ & $0.36$ & $0.05$ & $0.10$ & $1.09$ & $0.40$ & $0.14$ & $0.19$ & $3.54$ \\ 
    & & $5.0^{\circ}$ & $0.02$ & $0.16$ & $0.51$ & $0.39$ & $1.42$ & $0.38$ & $0.14$ & $0.24$ & $4.42$ \\ 
    \bottomrule 
    \end{tabular}
    }
    \caption{\textbf{Ablation on the outlier reweighting mechanism on the DiLiGenT benchmark~\cite{Shi2016DiLiGenT}.} The mean absolute depth error (MADE) [$\si{mm}$] of our method is reported. The upper outlier reweighting threshold $U$ is set to $10^{-3}$, and soft threshold with different lower thresholds $L$ as well as hard thresholding based on $U$ are compared. All experiments use convergence criteria $\Delta E_{\mathrm{max}}=10^{-3}$ and $T=150$, $8-$ pixel connectivity, without merging.}
\label{tab:diligent_ablation_outlier_reweighting}
\end{table*}
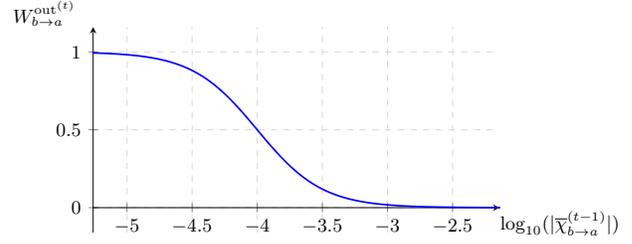
\begin{figure}[t!]
    \centering
    \begin{tikzpicture}[scale=0.8]
        \begin{axis}[
            width=\linewidth,
            height=0.6\linewidth,
            grid=both,
            grid style={dashed, gray!30},
            xlabel={$\log_{10}(|\overline{\chi}^{(t-1)}_{b\rightarrow a}|)$},
            ylabel={$W^{{\mathrm{out}}^{(t)}}_{b\rightarrow a}$},
            xmax=-2.4,
            xmin=-5,
            ymax=1.05,
            ymin=-0.05,
            xlabel style={below right},
            axis y line=left,
            axis x line=middle,
            every axis x label/.style={at={(current axis.right of origin)},right=25mm,below=-1mm},
            every axis y label/.style={at={(current axis.north west)},above=8.5mm,left=0.1mm},
            enlargelimits=true,
            ticklabel style={font=\small},
            label style={font=\small},
            title style={font=\small},
            legend pos=south east
        ]
            \addplot[domain=-5.6:-1.6, samples=200, thick, blue] {1 / (1 + exp(2*(2*x + 8)))};
        \end{axis}
    \end{tikzpicture}
    \caption{\textbf{Outlier reweighting~\eqref{eq:outlier_reweighting} as a function of $\log_{10}(|\overline{\chi}^{(t-1)}_{b\rightarrow a}|)$,  for} $L=10^{-3}$ and $U=10^{-5}$.
    }
    \label{fig:soft_outlier_reweighting}
    \vspace{10pt}
\end{figure}
To complement the definition provided in the main paper, we provide an illustration of our outlier reweighting function in \cref{fig:soft_outlier_reweighting}, using the parameters from the main experiments.
Additionally,
we
ablate on different values for its hyperparameter $L$ and also include a variant of the reweighting which introduces a \emph{hard} outlier thresholding, so that equations with residual magnitude $|\overline{\chi}^{(t-1)}_{b\rightarrow a}|$ are assigned weight $W^{{\mathrm{out}}^{(t)}}_{b\rightarrow a}=0$ if $|\overline{\chi}^{(t-1)}_{b\rightarrow a}| \ge U$ and weight $W^{{\mathrm{out}}^{(t)}}_{b\rightarrow a}=1$ if $|\overline{\chi}^{(t-1)}_{b\rightarrow a}| < U$, where we set $U=10^{-3}$.

\Cref{tab:diligent_ablation_outlier_reweighting} reports the results of the ablation. Overall, the differences across the variants for
outlier reweighting are marginal for most objects. However, for objects with larger discontinuities (\texttt{buddha}, \texttt{harvest}, \texttt{reading}) soft reweighting achieves generally better accuracy, with a small degree of object specificity for the value of $L$, indicating that 
it
might be effective particularly
in controlling
outlier
residuals
that arise due to
discontinuities.

\section{Ablation on the convergence parameters~\label{sec_suppl:ablation_convergence_parameters}}
\begin{table*}[!ht]
    \centering
    \resizebox{\linewidth}{!}{
    \begin{tabular}{lll cc cc cc cc cc cc cc cc cc}
    \toprule
    \multirow{2}{*}{$\Delta E_{\mathrm{max}}$} & \multirow{2}{*}{$T$} & \multirow{2}{*}{$\theta_c$} & \multicolumn{2}{c}{\texttt{bear}} & \multicolumn{2}{c}{\texttt{buddha}} & \multicolumn{2}{c}{\texttt{cat}} & \multicolumn{2}{c}{\texttt{cow}} & \multicolumn{2}{c}{\texttt{harvest}} & \multicolumn{2}{c}{\texttt{pot1}} & \multicolumn{2}{c}{\texttt{pot2}} & \multicolumn{2}{c}{\texttt{reading}} & \multicolumn{2}{c}{\texttt{goblet}\textsuperscript{*}}\\
    \cmidrule(lr){4-5} \cmidrule(lr){6-7} \cmidrule(lr){8-9} \cmidrule(lr){10-11} \cmidrule(lr){12-13} \cmidrule(lr){14-15} \cmidrule(lr){16-17} \cmidrule(lr){18-19} \cmidrule(lr){20-21}
    & & & $\mathrm{Err}$ & $t$ & $\mathrm{Err}$ & $t$ & $\mathrm{Err}$ & $t$ & $\mathrm{Err}$ & $t$ & $\mathrm{Err}$ & $t$ & $\mathrm{Err}$ & $t$ & $\mathrm{Err}$ & $t$ & $\mathrm{Err}$ & $t$ & $\mathrm{Err}$ & $t$\\
    \midrule
    \multirow{4}{*}{$10^{-3}$} & \multirow{4}{*}{$5$} & None & $0.02$ & $4.08$ & $0.51$ & $10.56$ & $0.03$ & $5.17$ & $0.09$ & $2.24$ & $1.77$ & $17.40$ & $0.39$ & $9.49$ & $0.13$ & $4.23$ & $0.17$ & $4.74$ & $9.42$ & $5.54$ \\ 
    & & $2.0^{\circ}$ & $0.02$ & $2.49$ & $0.18$ & $7.92$ & $0.03$ & $2.91$ & $0.09$ & $2.37$ & $1.24$ & $11.65$ & $0.38$ & $5.26$ & $0.13$ & $4.05$ & $0.14$ & $4.75$ & $9.37$ & $1.27$ \\ 
    & & $3.5^{\circ}$ & $0.02$ & $1.22$ & $0.15$ & $4.89$ & $0.04$ & $1.51$ & $0.10$ & $1.45$ & $1.10$ & $7.55$ & $0.41$ & $2.63$ & $0.13$ & $2.26$ & $0.13$ & $1.92$ & $9.47$ & $0.87$ \\ 
    & & $5.0^{\circ}$ & $0.02$ & $0.86$ & $0.18$ & $2.65$ & $0.51$ & $1.11$ & $0.39$ & $0.98$ & $1.78$ & $5.30$ & $0.56$ & $1.96$ & $0.13$ & $1.67$ & $0.18$ & $1.33$ & $9.56$ & $0.75$ \\ 
    \arrayrulecolor{gray!70}\specialrule{0.2pt}{0.2pt}{0.2pt}
    \arrayrulecolor{black}
    \multirow{4}{*}{$10^{-3}$} & \multirow{4}{*}{$15$} & None & $0.02$ & $6.09$ & $0.20$ & $17.69$ & $0.03$ & $9.17$ & $0.09$ & $3.92$ & $1.31$ & $46.33$ & $0.36$ & $22.12$ & $0.13$ & $9.64$ & $0.17$ & $4.31$ & $9.41$ & $5.15$ \\ 
    & & $2.0^{\circ}$ & $0.02$ & $2.55$ & $0.17$ & $17.08$ & $0.03$ & $3.11$ & $0.09$ & $2.50$ & $1.05$ & $28.62$ & $0.36$ & $6.27$ & $0.14$ & $5.40$ & $0.10$ & $7.01$ & $9.37$ & $1.58$ \\ 
    & & $3.5^{\circ}$ & $0.02$ & $1.28$ & $0.11$ & $7.62$ & $0.04$ & $1.52$ & $0.09$ & $1.54$ & $1.06$ & $16.82$ & $0.38$ & $3.27$ & $0.14$ & $2.61$ & $0.09$ & $2.43$ & $9.46$ & $0.99$ \\ 
    & & $5.0^{\circ}$ & $0.02$ & $0.89$ & $0.15$ & $2.75$ & $0.51$ & $1.22$ & $0.39$ & $1.04$ & $1.75$ & $7.87$ & $0.55$ & $2.26$ & $0.13$ & $1.93$ & $0.16$ & $1.49$ & $9.62$ & $0.86$ \\ 
    \arrayrulecolor{gray!70}\specialrule{0.2pt}{0.2pt}{0.2pt}
    \arrayrulecolor{black}
     \multirow{4}{*}{$10^{-3}$} & \multirow{4}{*}{$150$} & None & $0.02$ & $7.39$ & $0.20$ & $19.56$ & $0.03$ & $36.98$ & $0.09$ & $3.35$ & $1.31$ & $44.35$ & $0.36$ & $32.51$ & $0.13$ & $9.83$ & $0.17$ & $3.40$ & $9.41$ & $4.78$ \\ 
    & & $2.0^{\circ}$  & $0.02$ & $2.55$ & $0.17$ & $19.07$ & $0.03$ & $3.11$ & $0.09$ & $2.54$ & $1.04$ & $28.33$ & $0.36$ & $6.26$ & $0.14$ & $5.59$ & $0.10$ & $6.54$ & $9.37$ & $2.33$ \\ 
    & & $3.5^{\circ}$ & $0.02$ & $1.27$ & $0.11$ & $8.03$ & ${0.04}$ & $1.50$ & $0.09$ & $1.53$ & ${1.07}$ & $18.81$ & $0.38$ & $3.51$ & ${0.14}$ & $2.66$ & $0.09$ & $2.68$ & $9.49$ & $1.40$ \\ 
    & & $5.0^{\circ}$ &  ${0.02}$ & ${0.88}$ & ${0.15}$ & ${2.76}$ & $0.51$ & ${1.24}$ & $0.39$ & ${1.04}$ & $1.75$ & $7.29$ & $0.55$ & $5.80$ & ${0.13}$ & ${1.92}$ & $0.16$ & ${1.49}$ & $9.62$ & $0.84$ \\ 
    \arrayrulecolor{gray!70}\specialrule{0.2pt}{0.2pt}{0.2pt}
    \arrayrulecolor{black}
    \multirow{4}{*}{$10^{-5}$} & \multirow{4}{*}{$5$} & None & $0.02$ & $3.43$ & $0.51$ & $7.43$ & $0.03$ & $4.85$ & $0.09$ & $2.54$ & $1.77$ & $17.24$ & $0.39$ & $8.65$ & $0.13$ & $2.93$ & $0.17$ & $4.29$ & $9.42$ & $5.52$ \\ 
    & & $2.0^{\circ}$ & $0.02$ & $2.46$ & $0.18$ & $7.55$ & $0.03$ & $2.94$ & $0.09$ & $2.31$ & $1.24$ & $12.28$ & $0.38$ & $5.18$ & $0.13$ & $3.97$ & $0.14$ & $4.78$ & $9.37$ & $1.28$ \\ 
    & & $3.5^{\circ}$ & $0.02$ & $1.24$ & $0.15$ & $4.93$ & $0.04$ & $1.52$ & $0.10$ & $1.42$ & $1.10$ & $7.98$ & $0.41$ & $2.61$ & $0.13$ & $2.29$ & $0.13$ & $1.95$ & $9.47$ & $0.87$ \\ 
    & & $5.0^{\circ}$ &  $0.02$ & $0.85$ & $0.18$ & $2.65$ & $0.51$ & $1.12$ & $0.39$ & $0.98$ & $1.78$ & $5.18$ & $0.56$ & $1.97$ & $0.13$ & $1.67$ & $0.18$ & $1.35$ & $9.56$ & $0.81$ \\ 
    \arrayrulecolor{gray!70}\specialrule{0.2pt}{0.2pt}{0.2pt}
    \arrayrulecolor{black}
    \multirow{4}{*}{$10^{-5}$} & \multirow{4}{*}{$15$} & None & $0.01$ & $10.17$ & $0.15$ & $25.28$ & $0.03$ & $8.78$ & $0.09$ & $4.99$ & $1.28$ & $53.51$ & $0.36$ & $25.19$ & $0.13$ & $7.16$ & $0.09$ & $14.50$ & $9.40$ & $12.20$ \\ 
    & & $2.0^{\circ}$ & $0.02$ & $2.93$ & $0.17$ & $16.94$ & $0.03$ & $3.49$ & $0.09$ & $2.64$ & $1.05$ & $29.43$ & $0.36$ & $8.36$ & $0.14$ & $5.46$ & $0.09$ & $7.93$ & $9.37$ & $1.59$ \\ 
    & & $3.5^{\circ}$ & $0.02$ & $1.38$ & $0.11$ & $8.88$ & $0.04$ & $1.69$ & $0.09$ & $1.56$ & $1.06$ & $15.89$ & $0.38$ & $3.25$ & $0.14$ & $2.62$ & $0.09$ & $2.46$ & $9.46$ & $1.01$ \\ 
    & & $5.0^{\circ}$ & $0.02$ & $0.95$ & $0.12$ & $3.90$ & $0.51$ & $1.21$ & $0.39$ & $1.05$ & $1.69$ & $9.50$ & $0.55$ & $2.32$ & $0.13$ & $1.93$ & $0.15$ & $1.62$ & $9.62$ & $0.84$ \\ 
    \arrayrulecolor{gray!70}\specialrule{0.2pt}{0.2pt}{0.2pt}
    \arrayrulecolor{black}
    \multirow{4}{*}{$10^{-5}$} & \multirow{4}{*}{$150$} & None & $0.01$ & $11.20$ & $0.19$ & $253.51$ & $0.03$ & $42.47$ & $0.09$ & $12.51$ & $1.32$ & $387.01$ & $0.36$ & $169.22$ & $0.13$ & $80.35$ & $0.10$ & $66.33$ & $9.41$ & $66.55$ \\ 
    & & $2.0^{\circ}$ & $0.02$ & $3.04$ & $0.17$ & $137.36$ & $0.03$ & $4.97$ & $0.09$ & $2.86$ & $1.03$ & $336.65$ & $0.36$ & $78.28$ & $0.14$ & $12.60$ & $0.10$ & $73.13$ & $9.37$ & $6.63$ \\ 
    & & $3.5^{\circ}$ & $0.02$ & $1.60$ & $0.11$ & $52.42$ & $0.04$ & $2.88$ & $0.09$ & $1.93$ & $1.18$ & $98.90$ & $0.38$ & $11.63$ & $0.14$ & $4.24$ & $0.09$ & $10.85$ & $9.49$ & $3.32$ \\ 
    & & $5.0^{\circ}$ & $0.02$ & $1.10$ & $0.12$ & $14.57$ & $0.51$ & $2.08$ & $0.39$ & $1.36$ & $1.69$ & $121.54$ & $0.55$ & $7.42$ & $0.13$ & $2.64$ & $0.16$ & $5.83$ & $9.62$ & $2.44$ \\ 
    \bottomrule
    \end{tabular}
    }
    \caption{\textbf{Ablation on the convergence criteria $\Delta E_{\mathrm{max}}$ and $T$ on the DiLiGenT benchmark~\cite{Shi2016DiLiGenT}.} The mean absolute depth error (MADE, abbreviated as $\mathrm{Err}$) [$\si{mm}$] and the total execution time (abbreviated as $t$) [$\si{s}$] of our method are reported. All experiments use outlier reweighting $W^{\mathrm{out}}_{b\rightarrow a}$~\eqref{eq:outlier_reweighting} with $L=10^{-5}$ and $U=10^{-3}$, without merging.}
    \label{tab:diligent_ablation_delta_energy_and_T}
\end{table*}
\Cref{tab:diligent_ablation_delta_energy_and_T} reports the reconstruction error and run time achieved by our method for different values of the parameters $\Delta E_{\mathrm{max}}$ and $T$ that control the termination of its execution.
We observe that with limited exceptions, mostly restricted to our pixel-level variant ($\theta_c = \mathrm{None}$), enforcing a stricter relative-energy convergence criterion ($\Delta E_{\max} = 10^{-5}$) produces no significant changes in the accuracy of the reconstruction, while requiring a longer run time. Notably, the same conclusion applies to the maximum number of optimization steps, for which we find that our method effectively achieves convergence in $5$ or at most $15$ iterations for all objects. While
earlier termination of the optimization
results
in a slight increase in the running time, this speedup is not particularly significant for the small-scale of the normal maps from DiLiGenT, especially in the case of component-based (rather than pixel-level) optimization. For our main experiments on the DiLiGenT benchmark (\cf main paper), we therefore chose to use $T=150$, to enable convergence for the baselines without outlier reweighting.

\section{Ablation on the threshold for component formation~\label{sec_suppl:visualization_detected_components_diligent}}
\begin{figure*}[!ht]
\centering
\def\colwidth{0.2\textwidth}
\newcolumntype{M}[1]{>{\centering\arraybackslash}m{#1}}
\addtolength{\tabcolsep}{-4pt}
\begin{tabular}{m{0.7em} M{\colwidth} M{\colwidth} M{\colwidth}}
& $\theta_{c} = 2.0^{\circ}$ & $\theta_{c} = 3.5^{\circ}$ & $\theta_{c} = 5.0^{\circ}$ 
\tabularnewline
\cline{2-4}
\noalign{\vskip 0.5em}
\begin{turn}{90}
{\small\texttt{bear}}
\end{turn} & 
\includegraphics[width=\linewidth]{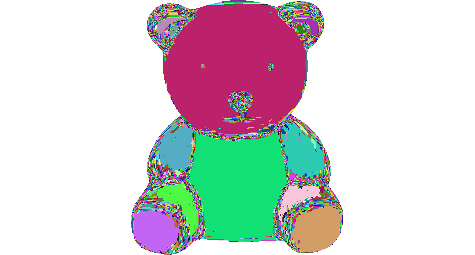} &  
\includegraphics[width=\linewidth]{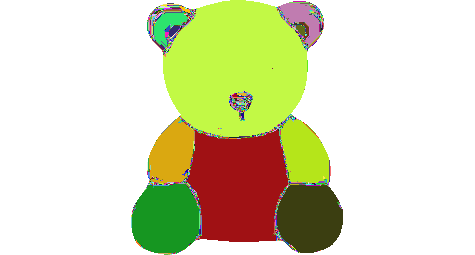} & 
\includegraphics[width=\linewidth]{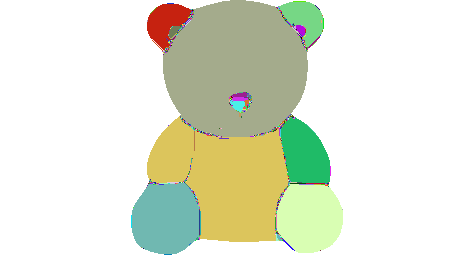}
\\[-5pt]
& \scriptsize{$0.009$,  \#components: $\num{5882}$} & \scriptsize{$0.011$, \#components: $\num{1900}$} & \scriptsize{$0.014$,  \#components: $\num{836}$} \tabularnewline
\begin{turn}{90}
{\small\texttt{buddha}}
\end{turn} & 
\includegraphics[width=\linewidth]{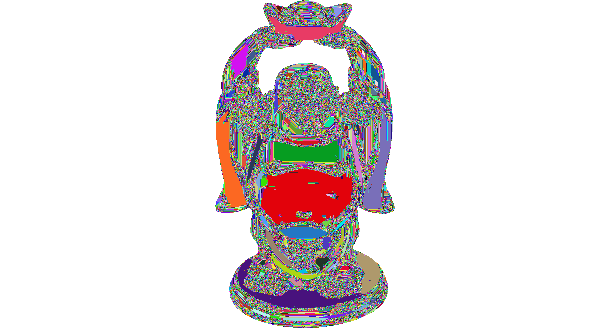} &  
\includegraphics[width=\linewidth]{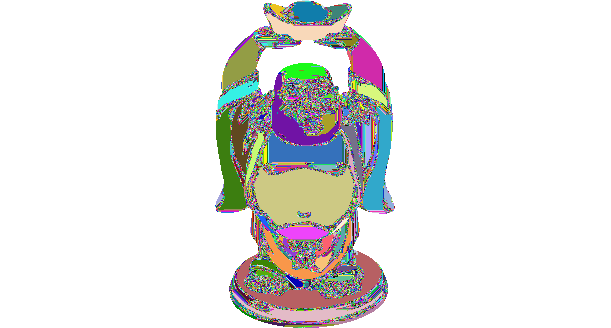} & 
\includegraphics[width=\linewidth]{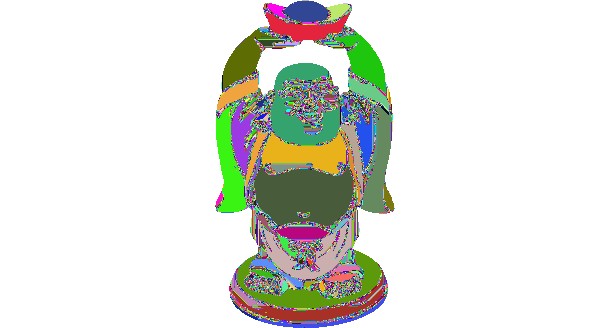}
\\[-5pt]
& \scriptsize{$0.010$,  \#components: $\num{22175}$} & \scriptsize{$0.016$,  \#components: $\num{13883}$} & \scriptsize{$0.025$,  \#components: $\num{9787}$} \tabularnewline
\begin{turn}{90}
{\small\texttt{cat}}
\end{turn} & 
\includegraphics[width=\linewidth]{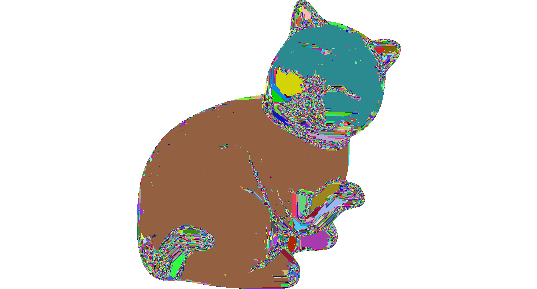} &  
\includegraphics[width=\linewidth]{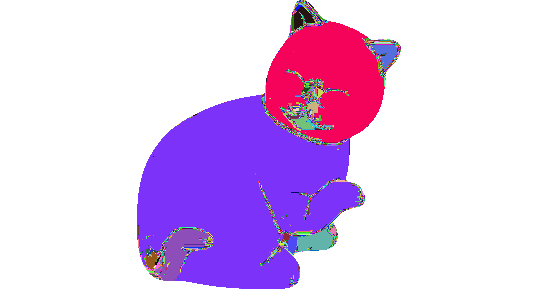} & 
\includegraphics[width=\linewidth]{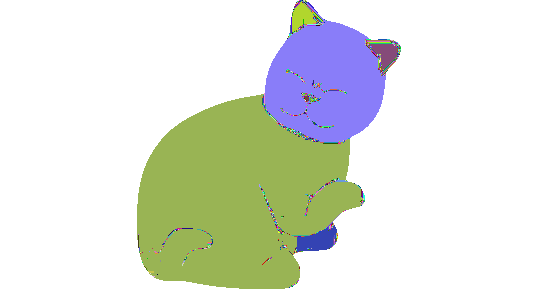}
\\[-5pt]
& \scriptsize{$0.013$,  \#components: $\num{8150}$} & \scriptsize{$0.022$,  \#components: $\num{2896}$} & \scriptsize{$0.285$,  \#components: $\num{1282}$} \tabularnewline
\begin{turn}{90}
{\small\texttt{cow}}
\end{turn} & 
\includegraphics[width=\linewidth]{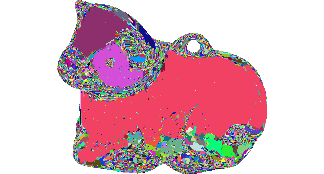} &  
\includegraphics[width=\linewidth]{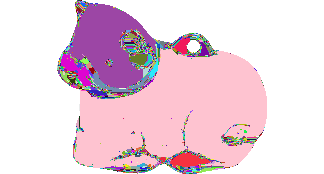} & 
\includegraphics[width=\linewidth]{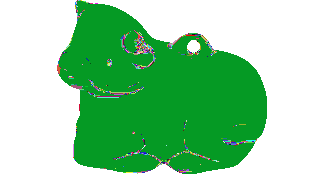}
\\[-5pt]
& \scriptsize{$0.027$,  \#components: $\num{6427}$} & \scriptsize{$0.047$,  \#components: $\num{2416}$} & \scriptsize{$0.355$,  \#components: $\num{997}$} \tabularnewline
\begin{turn}{90}
{\small\texttt{harvest}}
\end{turn} & 
\includegraphics[width=\linewidth]{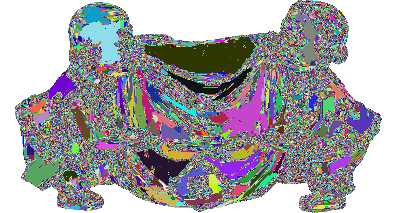} &  
\includegraphics[width=\linewidth]{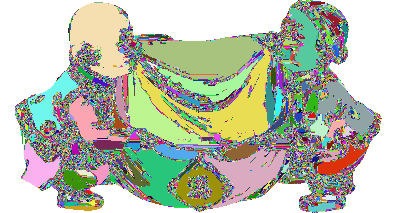} & 
\includegraphics[width=\linewidth]{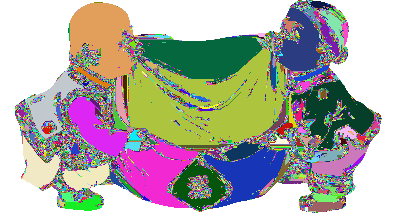}
\\[-5pt]
& \scriptsize{$0.022$,  \#components: $\num{29612}$} & \scriptsize{$0.033$,  \#components: $\num{18145}$} & \scriptsize{$0.121$,  \#components: $\num{12109}$} \tabularnewline
\begin{turn}{90}
{\small\texttt{pot1}}
\end{turn} & 
\includegraphics[width=\linewidth]{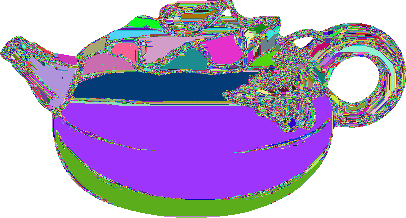} &  
\includegraphics[width=\linewidth]{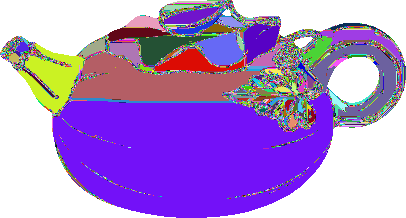} & 
\includegraphics[width=\linewidth]{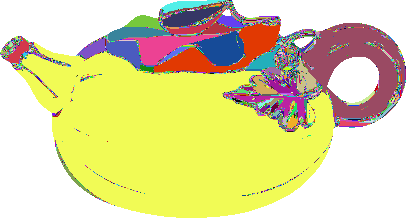}
\\[-5pt]
& \scriptsize{$0.087$,  \#components: $\num{14044}$} & \scriptsize{$0.137$,  \#components: $\num{7443}$} & \scriptsize{$0.190$,  \#components: $\num{4378}$} \tabularnewline
\begin{turn}{90}
{\small\texttt{pot2}}
\end{turn} & 
\includegraphics[width=\linewidth]{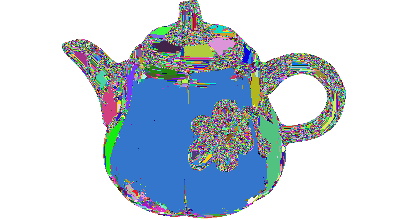} &  
\includegraphics[width=\linewidth]{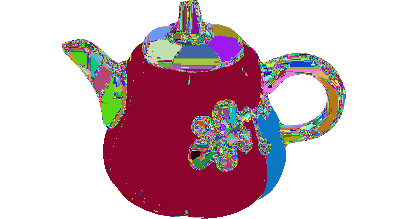} & 
\includegraphics[width=\linewidth]{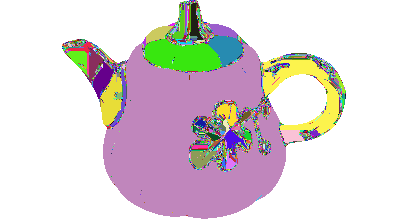}
\\[-5pt]
& \scriptsize{$0.046$,  \#components: $\num{11135}$} & \scriptsize{$0.067$,  \#components: $\num{5904}$} & \scriptsize{$0.079$,  \#components: $\num{3514}$} \tabularnewline
\begin{turn}{90}
{\small\texttt{reading}}
\end{turn} & 
\includegraphics[width=\linewidth]{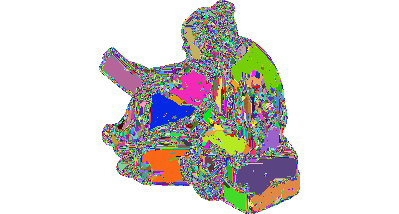} &  
\includegraphics[width=\linewidth]{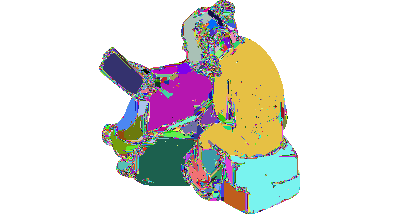} & 
\includegraphics[width=\linewidth]{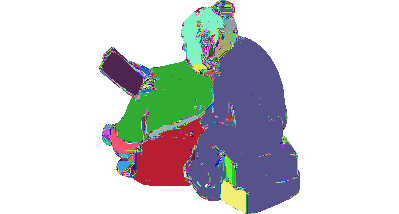}
\\[-5pt]
& \scriptsize{$0.011$,  \#components: $\num{11428}$} & \scriptsize{$0.019$,  \#components: $\num{5040}$} & \scriptsize{$0.084$,  \#components: $\num{2534}$} \tabularnewline
\begin{turn}{90}
{\small\texttt{goblet}}
\end{turn} & 
\includegraphics[width=\linewidth]{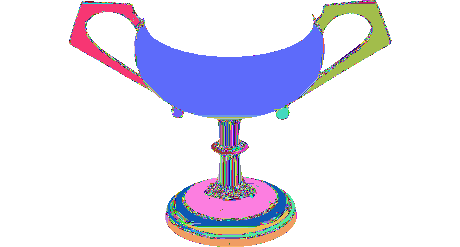} &  
\includegraphics[width=\linewidth]{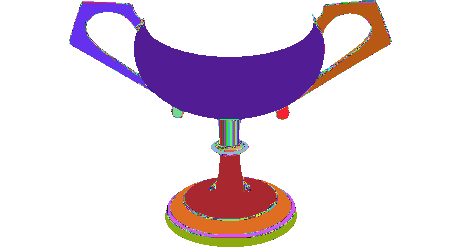} & 
\includegraphics[width=\linewidth]{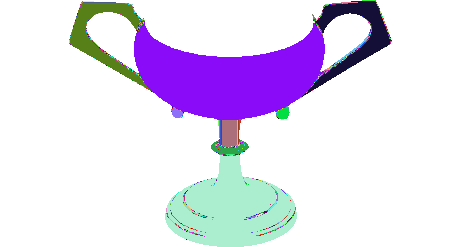}
\\[-5pt]
& \scriptsize{$0.014$,  \#components: $\num{3335}$} & \scriptsize{$0.060$,  \#components: $\num{1512}$} & \scriptsize{$0.113$,  \#components: $\num{933}$} \tabularnewline
\end{tabular}
\addtolength{\tabcolsep}{4pt}
\caption{\textbf{Continuous components identified by our method for different normal similarity thresholds $\theta_{c}$ and $8-$ connectivity, DiLiGenT benchmark~\cite{Shi2016DiLiGenT}.} For each object and threshold, different colors indicate different 
components.
Below each component image are: the minimum mean average depth error (MADE, in $\mathrm{mm}$) that can be theoretically achieved by scaling the continuous components; the number of components.
}
\label{fig:comparison_decompositions_different_thresholds}
\end{figure*}
\Cref{fig:comparison_decompositions_different_thresholds} shows the continuous components identified by our heuristic on the DiLiGenT benchmark, for
the
different values of the normal similarity threshold $\theta_c$ that we use in our experiments.

We note that for each decomposition it is possible to compute the minimum theoretical reconstruction error that could be achieved by the relative scale optimization,
which
coincides with the global mean average depth error (MADE) that would be attained if the optimal combination of scales was applied to the individual reconstructions formed in the component filling stage. In general, each of such per-component reconstructions introduces an error, for however small, in its corresponding surface region. This is due to the approximations inherent in the models of continuity and discontinuity, as well as to the convergence of the optimization. To compute the minimum MADE, it is sufficient to compute the sum of the per-component MADEs, weighted by the number of pixels in each component.

As shown in \cref{fig:comparison_decompositions_different_thresholds}, choosing a smaller threshold for $\theta_c$ results in a smaller minimum theoretical MADE, at the cost of a larger number of components. This is coherent with the fact that the minimum theoretical MADE decreases as the number of components
approaches
the total number of pixels, with a minimum value of $0$ in the limit of per-pixel components, since by definition a perfect reconstruction could be achieved by appropriately scaling the depth of each pixel.

Importantly, we note that for all objects in the benchmark the minimum theoretical MADE achievable by our method is significantly smaller than the accuracy reached by state-of-the-art methods, by different margins depending on the exact value of $\theta_c$. On the one hand, this validates the effectiveness of our component formation and filling, which does not compromise the best quality that the reconstruction can achieve. On the other hand, the remaining gap between the minimum theoretical MADE and the MADE actually achieved by our optimization
presents an opportunity for the future emergence of more accurate models of continuity and
discontinuity,
which
could lead to
even more optimal
relative scale optimization.

\section{Ablation on the effect of merging in the DiLiGenT benchmark~\label{sec_suppl:ablation_effect_merging_diligent}}
\begin{figure*}[!ht]
\centering
\def\colwidth{0.13\textwidth}
\def\colsmallerwidth{1em}
\newcolumntype{M}[1]{>{\centering\arraybackslash}m{#1}}
\addtolength{\tabcolsep}{-4pt}
\begin{tabular}{m{0.7em} M{\colwidth} M{\colwidth} M{\colwidth} M{\colsmallerwidth} M{\colwidth} M{\colsmallerwidth} M{\colwidth} m{0.5em} r}
& $m=0$ & $m=1$ & $m=2$ & & $m=\left\lfloor\frac{M_{\mathrm{tot}}}{2}\right\rfloor$& & $m=M_{\mathrm{tot}} - 1$ & & $M_{\mathrm{tot}}$
\tabularnewline
\noalign{\vskip 0.5em}
\cline{2-8}
\noalign{\vskip 0.5em}
\multirow{2}{*}{\begin{turn}{90}
{\small\texttt{bear}}
\end{turn}} &  
\includegraphics[width=\linewidth]{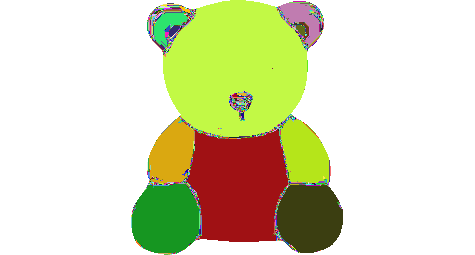} & 
\includegraphics[width=\linewidth]{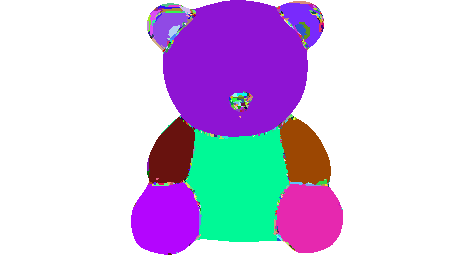} & \includegraphics[width=\linewidth]{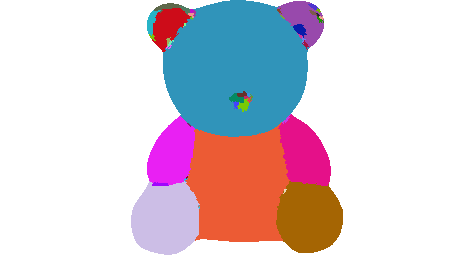} & $\cdots$ & \includegraphics[width=\linewidth]{figures/images/example_components/multiple_steps/bear_component_image_iter_0011_num_components_47.png} & $\cdots$ & \includegraphics[width=\linewidth]{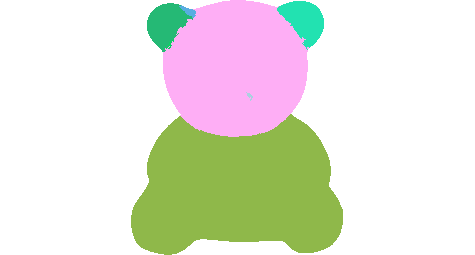} & & \multirow{2}{*}{$4$}
\\[-5pt]
& \scriptsize{$0.011\ \ (\underline{0.018})$} & \scriptsize{$0.013\ \ (\underline{0.018})$} & \scriptsize{$0.014\ \ (\underline{0.018})$} & & \scriptsize{$0.014\ \ (\underline{0.018})$} & & \scriptsize{$0.016\ \ (\underline{0.018})$} \\[-2pt]
& \scriptsize{\#components:\ $\num{1900}$} & \scriptsize{\#components:\ $\num{361}$} & \scriptsize{\#components:\ $\num{47}$} & & \scriptsize{\#components:\ $\num{47}$} &  & \scriptsize{\#components:\ $\num{7}$} \tabularnewline
\cmidrule(lr){2-9}
\multirow{2}{*}{\begin{turn}{90}
{\small\texttt{buddha}}
\end{turn}} &  
\includegraphics[width=\linewidth]{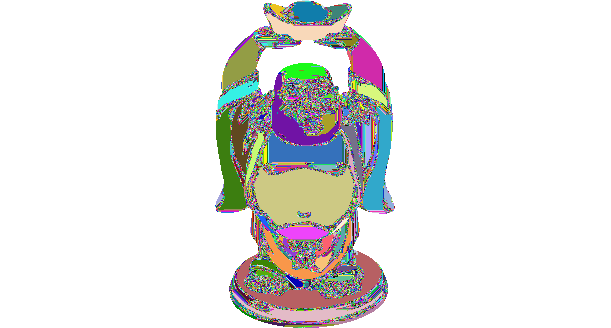} & 
\includegraphics[width=\linewidth]{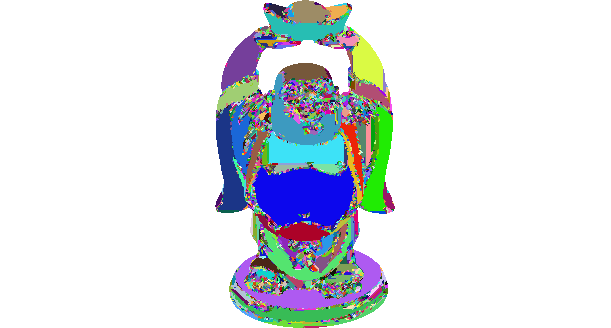} & \includegraphics[width=\linewidth]{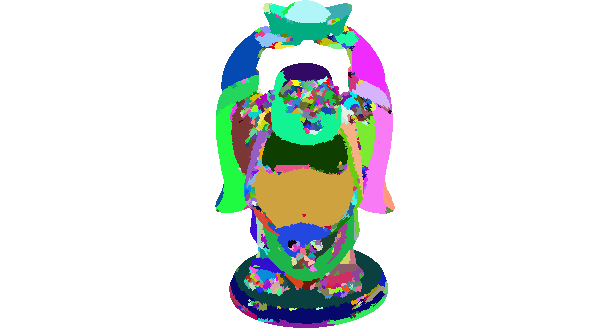} & $\cdots$ & \includegraphics[width=\linewidth]{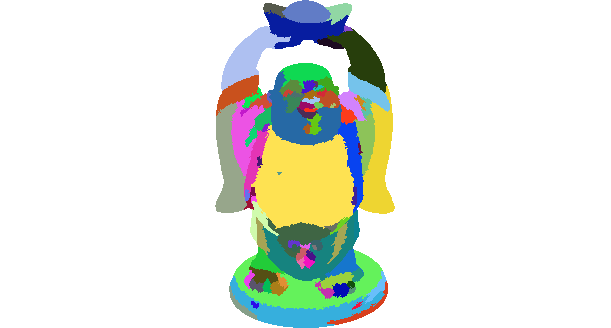} & $\cdots$ & \includegraphics[width=\linewidth]{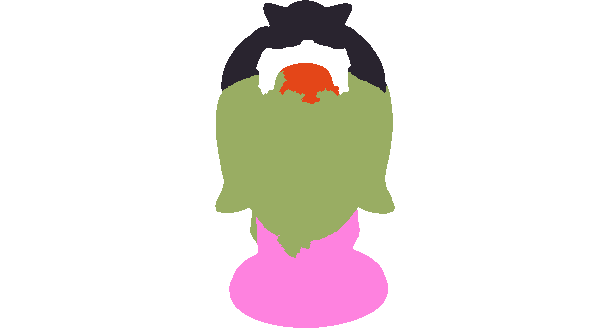} & & \multirow{2}{*}{$6$}
\\[-5pt]
& \scriptsize{$0.016\ \ (\underline{0.148})$} & \scriptsize{$0.049\ \ (\underline{0.116})$} & \scriptsize{$0.063\ \ (\underline{0.115})$} & & \scriptsize{$0.079\ \ (\underline{0.115})$} & & \scriptsize{$0.089\ \ (\underline{0.115})$} \\[-2pt]
& \scriptsize{\#components:\ $\num{13883}$} & \scriptsize{\#components:\ $\num{2904}$} & \scriptsize{\#components:\ $\num{476}$} & & \scriptsize{\#components:\ $\num{95}$} &  & \scriptsize{\#components:\ $\num{4}$} \tabularnewline
\cmidrule(lr){2-9}
\multirow{2}{*}{\begin{turn}{90}
{\small\texttt{cat}}
\end{turn}} &  
\includegraphics[width=\linewidth]{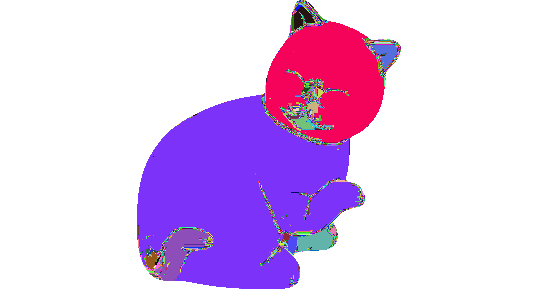} & 
\includegraphics[width=\linewidth]{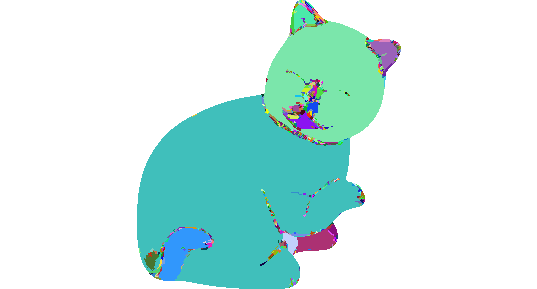} & \includegraphics[width=\linewidth]{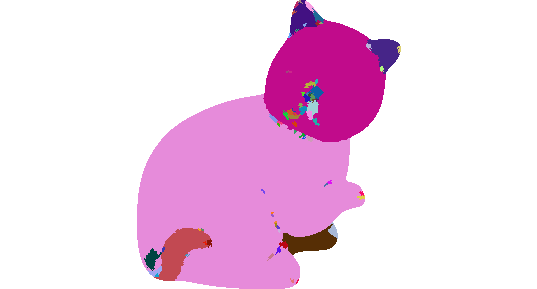} & $\cdots$ & \includegraphics[width=\linewidth]{figures/images/example_components/multiple_steps/cat_component_image_iter_0011_num_components_76} & $\cdots$ & \includegraphics[width=\linewidth]{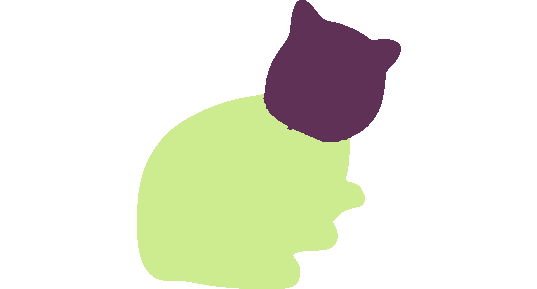} & & \multirow{2}{*}{$5$}
\\[-5pt]
& \scriptsize{$0.022\ \ (\underline{0.044})$} & \scriptsize{$0.029\ \ (\underline{0.044})$} &  \scriptsize{$0.032\ \ (\underline{0.044})$} & & \scriptsize{$0.032\ \ (\underline{0.044})$} & &  \scriptsize{$0.036\ \ (\underline{0.044})$} \\[-2pt]
& \scriptsize{\#components:\ $\num{2896}$} & \scriptsize{\#components:\ $\num{568}$} & \scriptsize{\#components:\ $\num{76}$} & & \scriptsize{\#components:\ $\num{76}$} &  & \scriptsize{\#components:\ $\num{2}$} \tabularnewline
\cmidrule(lr){2-9}
\multirow{2}{*}{\begin{turn}{90}
{\small\texttt{cow}}
\end{turn}} &  
\includegraphics[width=\linewidth]{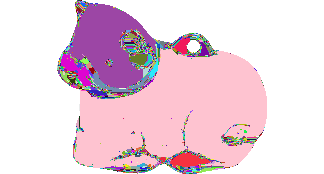} & 
\includegraphics[width=\linewidth]{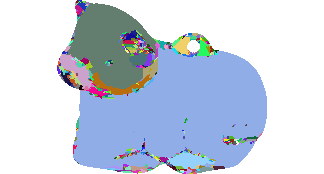} & \includegraphics[width=\linewidth]{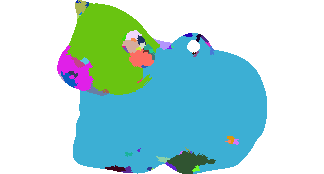} & $\cdots$ & \includegraphics[width=\linewidth]{figures/images/example_components/multiple_steps/cow_component_image_iter_0011_num_components_57} & $\cdots$ & \includegraphics[width=\linewidth]{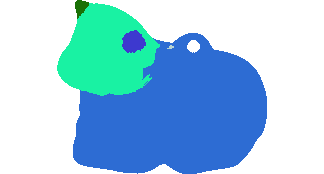} & & \multirow{2}{*}{$4$}
\\[-5pt]
& \scriptsize{$0.047\ \ (\underline{0.096})$} & \scriptsize{$0.054\ \ (\underline{0.095})$} & \scriptsize{$0.062\ \ (\underline{0.095})$} & & \scriptsize{$0.062\ \ (\underline{0.095})$} & & \scriptsize{$0.067\ \ (\underline{0.095})$} \\[-2pt]
& \scriptsize{\#components:\ $\num{2416}$} & \scriptsize{\#components:\ $\num{455}$} & \scriptsize{\#components:\ $\num{57}$} & & \scriptsize{\#components:\ $\num{57}$} &  & \scriptsize{\#components:\ $\num{5}$} \tabularnewline
\cmidrule(lr){2-9}
\multirow{2}{*}{\begin{turn}{90}
{\small\texttt{harvest}}
\end{turn}} &  
\includegraphics[width=\linewidth]{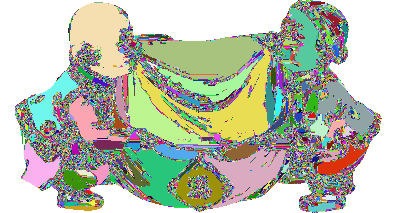} & 
\includegraphics[width=\linewidth]{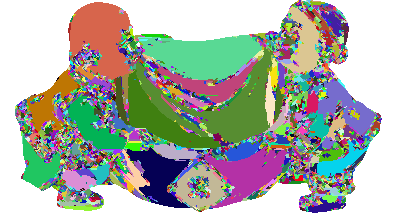} & \includegraphics[width=\linewidth]{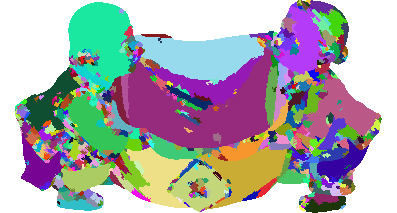} & $\cdots$ & \includegraphics[width=\linewidth]{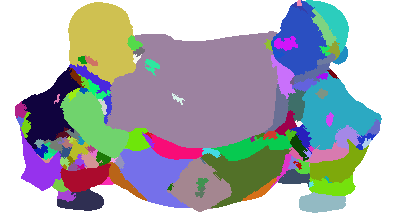} & $\cdots$ & \includegraphics[width=\linewidth]{figures/images/example_components/multiple_steps/harvest_component_image_iter_0016_num_components_122} & & \multirow{2}{*}{$6$}
\\[-5pt]
& \scriptsize{$0.033\ \ (\underline{1.095})$} & \scriptsize{$0.100\ \ (\underline{1.086})$} & \scriptsize{$0.123\ \ (\underline{1.079})$} & & \scriptsize{$0.142\ \ (\underline{1.079})$} & & \scriptsize{$0.641\ \ (\underline{1.079})$} \\[-2pt]
& \scriptsize{\#components:\ $\num{18145}$} & \scriptsize{\#components:\ $\num{3867}$} & \scriptsize{\#components:\ $\num{677}$} & & \scriptsize{\#components:\ $\num{122}$} &  & \scriptsize{\#components:\ $\num{4}$} \tabularnewline
\cmidrule(lr){2-9}
\multirow{2}{*}{\begin{turn}{90}
{\small\texttt{pot1}}
\end{turn}} &  
\includegraphics[width=\linewidth]{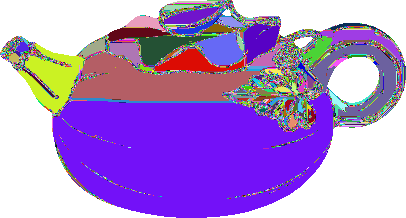} & 
\includegraphics[width=\linewidth]{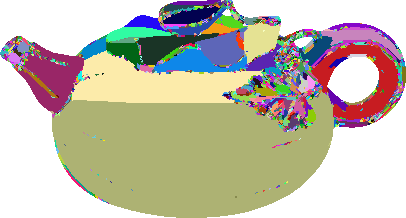} & \includegraphics[width=\linewidth]{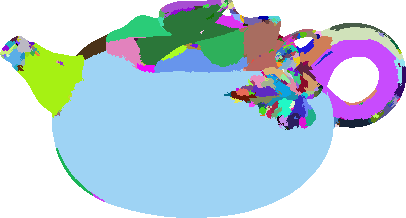} & $\cdots$ & \includegraphics[width=\linewidth]{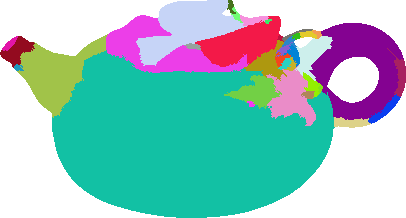} & $\cdots$ & \includegraphics[width=\linewidth]{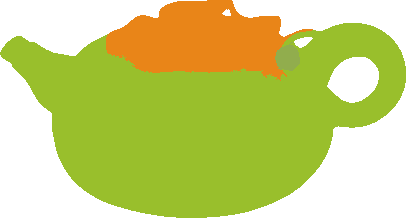} & & \multirow{2}{*}{$6$}
\\[-5pt]
& \scriptsize{$0.137\ \ (\underline{0.409})$} & \scriptsize{$0.178\ \ (\underline{0.395})$} & \scriptsize{$0.211\ \ (\underline{0.394})$} & & \scriptsize{$0.233\ \ (\underline{0.395})$} & & \scriptsize{$0.386\ \ (\underline{0.395})$} \\[-2pt]
& \scriptsize{\#components:\ $\num{7443}$} & \scriptsize{\#components:\ $\num{1408}$} & \scriptsize{\#components:\ $\num{216}$} & & \scriptsize{\#components:\ $\num{36}$} &  & \scriptsize{\#components:\ $\num{3}$} \tabularnewline
\cmidrule(lr){2-9}
\multirow{2}{*}{\begin{turn}{90}
{\small\texttt{pot2}}
\end{turn}} &  
\includegraphics[width=\linewidth]{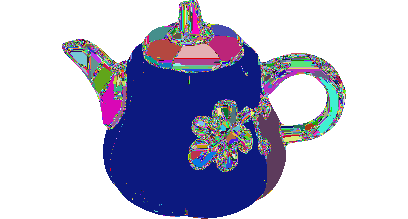} & 
\includegraphics[width=\linewidth]{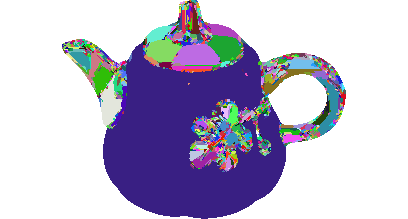} & \includegraphics[width=\linewidth]{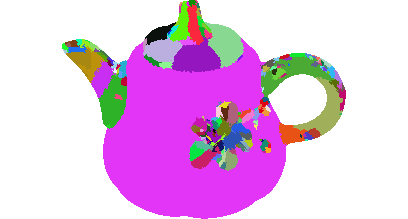} & $\cdots$ & \includegraphics[width=\linewidth]{figures/images/example_components/multiple_steps/pot2_component_image_iter_0011_num_components_181} & $\cdots$ & \includegraphics[width=\linewidth]{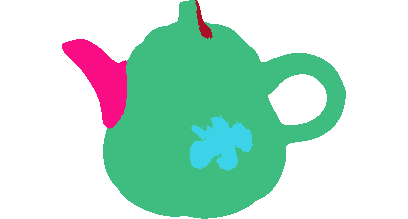} & & \multirow{2}{*}{$5$}
\\[-5pt]
& \scriptsize{$0.067\ \ (\underline{0.135})$} & \scriptsize{$0.084\ \ (\underline{0.135})$} & \scriptsize{$0.094\ \ (\underline{0.135})$} & & \scriptsize{$0.094\ \ (\underline{0.135})$} & & \scriptsize{$0.131\ \ (\underline{0.135})$} \\[-2pt]
& \scriptsize{\#components:\ $\num{5904}$} & \scriptsize{\#components:\ $\num{1224}$} & \scriptsize{\#components:\ $\num{181}$} & & \scriptsize{\#components:\ $\num{181}$} &  & \scriptsize{\#components:\ $\num{4}$} \tabularnewline
\cmidrule(lr){2-9}
\multirow{2}{*}{\begin{turn}{90}
{\small\texttt{reading}}
\end{turn}} &  
\includegraphics[width=\linewidth]{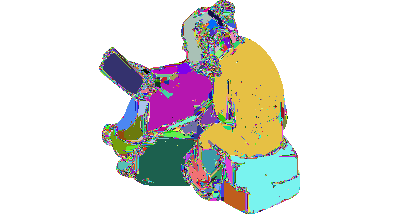} & 
\includegraphics[width=\linewidth]{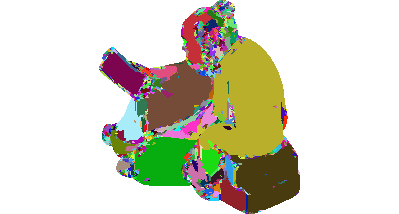} & \includegraphics[width=\linewidth]{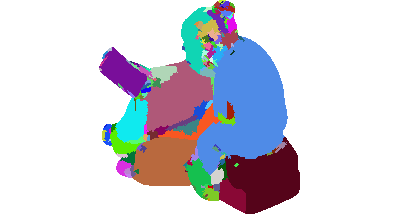} & $\cdots$ & \includegraphics[width=\linewidth]{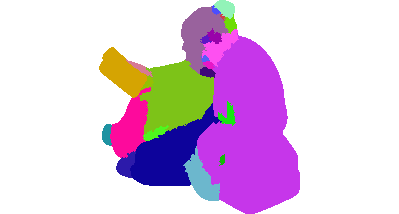} & $\cdots$ & \includegraphics[width=\linewidth]{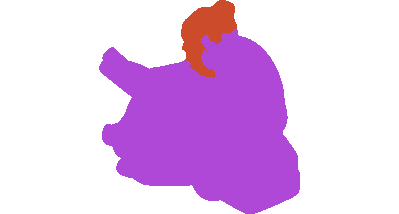} & & \multirow{2}{*}{$6$}
\\[-5pt]
& \scriptsize{$0.019\ \ (\underline{0.132})$} & \scriptsize{$0.044\ \ (\underline{0.111})$} & \scriptsize{$0.055\ \ (\underline{0.106})$} & & \scriptsize{$0.065\ \ (\underline{0.106})$} & & \scriptsize{$0.102\ \ (\underline{0.106})$} \\[-2pt]
& \scriptsize{\#components:\ $\num{5040}$} & \scriptsize{\#components:\ $\num{1006}$} & \scriptsize{\#components:\ $\num{148}$} & & \scriptsize{\#components:\ $\num{24}$} &  & \scriptsize{\#components:\ $\num{2}$} \tabularnewline
\cmidrule(lr){2-9}
\multirow{2}{*}{\begin{turn}{90}
{\small\texttt{goblet}}
\end{turn}} &  
\includegraphics[width=\linewidth]{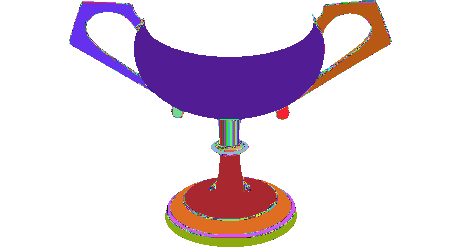} & 
\includegraphics[width=\linewidth]{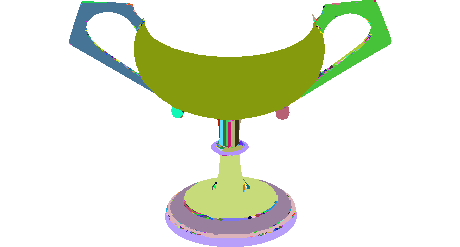} & \includegraphics[width=\linewidth]{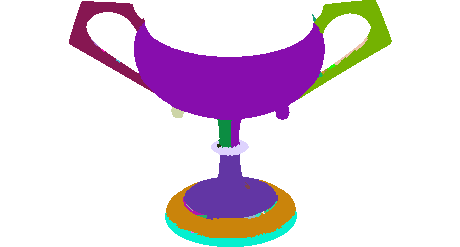} & $\cdots$ & \includegraphics[width=\linewidth]{figures/images/example_components/multiple_steps/goblet_component_image_iter_0016_num_components_28} & $\cdots$ & \includegraphics[width=\linewidth]{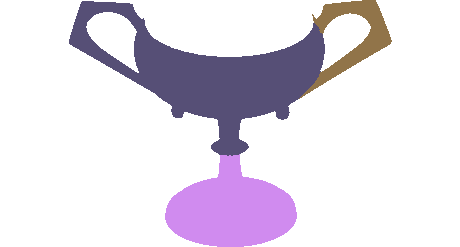} & & \multirow{2}{*}{$5$}
\\[-5pt]
& \scriptsize{$0.060\ \ (\underline{9.471})$} & \scriptsize{$0.086\ \ (\underline{9.499})$} & \scriptsize{$0.386\ \ (\underline{9.499})$} & & \scriptsize{$0.386\ \ (\underline{9.499})$} & & \scriptsize{$1.095\ \ (\underline{9.499})$} \\[-2pt]
& \scriptsize{\#components:\ $\num{1521}$} & \scriptsize{\#components:\ $\num{244}$} & \scriptsize{\#components:\ $\num{28}$} & & \scriptsize{\#components:\ $\num{28}$} &  & \scriptsize{\#components:\ $\num{3}$} \tabularnewline
\cmidrule(lr){2-9}
\end{tabular}
\addtolength{\tabcolsep}{4pt}
\caption{\textbf{Continuous components at different stages $m$ of the merging process, DiLiGenT benchmark~\cite{Shi2016DiLiGenT}.} For each object and threshold, different colors indicate different 
components.
Below each component image are: \textit{first row}: the minimum mean average depth error (MADE, in $\mathrm{mm}$) that can be theoretically achieved by scaling the continuous components and the MADE ($\si{mm}$) achieved by the optimization at the end of the merging stage, the latter in brackets and \underline{underlined}; \textit{second row}: number of components. For each object, $M_{\mathrm{tot}}$ denotes the total number of merging operations required to obtain a single component. For all objects,
normal similarity threshold 
$\theta_{c}=3.5^{\circ}$ and $8-$ connectivity are used, with $\mathrm{freq}_{\mathrm{merging}}=5$ and $\Delta E_{\mathrm{max}} = 10^{-3}$.
}
\label{fig:visualization_merging}
\end{figure*}
\Cref{fig:visualization_merging} shows the progression of the component decomposition, of the minimum theoretical MADE, and of the actual MADE computed from the reconstruction, at different stages of our optional merging process.
The illustrated example uses a merging frequency of $5$ optimization iterations; as previously noted (\cf \cref{sec_suppl:ablation_convergence_parameters} and \cref{tab:diligent_ablation_delta_energy_and_T}), on the small-scale normals from DiLiGenT our method achieves convergence for most object already after this number of iterations. This is
reflected in the fact that the reconstruction error (indicated within brackets in \cref{fig:visualization_merging}) does not change significantly after the first merge operation for most objects. However, small improvements can be observed for objects with larger discontinuities (\texttt{buddha}, \texttt{harvest}, \texttt{reading}). This indicates that in the presence of discontinuities optimization requires a larger number of iterations to converge. For this reason, 
merging can be a viable option to reduce the size of the optimization problem (hence also the execution time of later iterations) while allowing the convergence process to continue.
As already observed (\cref{sec_suppl:ablation_convergence_parameters}), while the computational effect of this operation might be negligible for small-scale normal maps, its impact becomes 
significantly
more important for larger-scale normal maps, for which a reduction in the number of variables can
largely
reduce the run time of the optimization (\cf. \cref{fig:comparison_high_resolution_normal_maps}).

We note, finally, that merging in general increases the minimum theoretical MADE, since it ``fixes" the relative scale of neighboring components to a value that in general does not coincide with its optimal one. However, we stress that the merging operation per se does not increase the reconstruction error with respect to the previous optimization steps. This is because it does not alter the relative scales to which the optimization had converged at the preceding step, but simply relabels pixels in different components so that their scale is jointly optimized in subsequent iterations.

\section{Ablation on pixel connectivity~\label{sec_suppl:impact_connectivity}}

\begin{table*}[!ht]
    \centering
    \resizebox{0.9\linewidth}{!}{
    \begin{tabular}{ll cc cc cc cc cc cc cc cc cc}
    \toprule
    \multirow{2}{*}{$\theta_c$} & \multirow{2}{*}{$\mathrm{Conn}.$} & \multicolumn{2}{c}{\texttt{bear}} & \multicolumn{2}{c}{\texttt{buddha}} & \multicolumn{2}{c}{\texttt{cat}} & \multicolumn{2}{c}{\texttt{cow}} & \multicolumn{2}{c}{\texttt{harvest}} & \multicolumn{2}{c}{\texttt{pot1}} & \multicolumn{2}{c}{\texttt{pot2}} & \multicolumn{2}{c}{\texttt{reading}} & \multicolumn{2}{c}{\texttt{goblet}} \\
    \cmidrule(lr){3-4} \cmidrule(lr){5-6} \cmidrule(lr){7-8} \cmidrule(lr){9-10} \cmidrule(lr){11-12} \cmidrule(lr){13-14} \cmidrule(lr){15-16} \cmidrule(lr){17-18} \cmidrule(lr){19-20}
    & & $\mathrm{Err}$ & $t$ & $\mathrm{Err}$ & $t$ & $\mathrm{Err}$ & $t$ & $\mathrm{Err}$ & $t$ & $\mathrm{Err}$ & $t$ & $\mathrm{Err}$ & $t$ & $\mathrm{Err}$ & $t$ & $\mathrm{Err}$ & $t$ & $\mathrm{Err}$ & $t$\\
    \midrule
    None & $8$ & $0.02$ & $7.39$ & $0.20$ & $19.56$ & $0.03$ & $36.98$ & $0.09$ & $3.35$ & $1.31$ & $44.35$ & $0.36$ & $32.51$ & $0.13$ & $9.83$ & $0.17$ & $3.40$ & $9.41$ & $4.78$ \\ 
    $2.0^{\circ}$ & $8$  & $0.02$ & $2.55$ & $0.17$ & $19.07$ & $0.03$ & $3.11$ & $0.09$ & $2.54$ & $1.04$ & $28.33$ & $0.36$ & $6.26$ & $0.14$ & $5.59$ & $0.10$ & $6.54$ & $9.37$ & $2.33$ \\ 
    $3.5^{\circ}$ & $8$ & $0.02$ & $1.27$ & $0.11$ & $8.03$ & ${0.04}$ & $1.50$ & $0.09$ & $1.53$ & ${1.07}$ & $18.81$ & $0.38$ & $3.51$ & ${0.14}$ & $2.66$ & $0.09$ & $2.68$ & $9.49$ & $1.40$ \\ 
    $5.0^{\circ}$ & $8$ &  ${0.02}$ & ${0.88}$ & ${0.15}$ & ${2.76}$ & $0.51$ & ${1.24}$ & $0.39$ & ${1.04}$ & $1.75$ & $7.29$ & $0.55$ & $5.80$ & ${0.13}$ & ${1.92}$ & $0.16$ & ${1.49}$ & $9.62$ & $0.84$ \\ 
    \arrayrulecolor{gray!70}\specialrule{0.2pt}{0.2pt}{0.2pt}
    \arrayrulecolor{black}
    None & $4$ & $0.03$ & $15.52$ & $0.13$ & $40.19$ & $0.03$ & $16.58$ & $0.08$ & $4.19$ & $1.26$ & $64.00$ & $0.37$ & $209.27$ & $0.13$ & $7.69$ & $0.08$ & $7.73$ & $9.45$ & $111.51$ \\ 
    $2.0^{\circ}$ & $4$ & $0.03$ & $2.98$ & $0.12$ & $26.65$ & $0.03$ & $9.15$ & $0.09$ & $2.97$ & $1.09$ & $29.39$ & $0.38$ & $8.29$ & $0.14$ & $27.20$ & $0.09$ & $17.03$ & $9.27$ & $1.89$ \\ 
    $3.5^{\circ}$ & $4$ & $0.04$ & $1.18$ & $0.14$ & $38.11$ & $0.04$ & $1.65$ & $0.09$ & $2.58$ & $1.09$ & $17.91$ & $0.35$ & $25.06$ & $0.14$ & $2.37$ & $0.08$ & $2.80$ & $9.43$ & $0.96$ \\ 
    $5.0^{\circ}$ & $4$ & $0.02$ & $0.77$ & $0.12$ & $4.46$ & $0.03$ & $1.23$ & $0.09$ & $1.14$ & $0.80$ & $11.11$ & $0.57$ & $7.72$ & $0.13$ & $4.84$ & $0.10$ & $2.21$ & $9.51$ & $0.73$ \\ 
    \bottomrule
    \end{tabular}
    }
    \caption{\textbf{Ablation on the pixel connectivity (abbreviated as $\mathrm{Conn.}$) on the DiLiGenT benchmark~\cite{Shi2016DiLiGenT}.} The mean absolute depth error (MADE, abbreviated as $\mathrm{Err}$) [$\si{mm}$] and the total execution time (abbreviated as $t$) [$\si{s}$] of our method are reported. All experiments use outlier reweighting $W^{\mathrm{out}}_{b\rightarrow a}$~\eqref{eq:outlier_reweighting} with $L=10^{-5}$ and $U=10^{-3}$, convergence criteria $\Delta E_{\mathrm{max}}=10^{-3}$ and $T=150$, without merging.}
    \label{tab:diligent_ablation_connectivity}
\end{table*}
We ablate the impact of the chosen pixel connectivity on the DiLiGenT benchmark, comparing $8-$ connectivity, which we use in our main experiments, with $4-$
connectivity. For a given configuration, we use the connectivity both in the component detection/filling stage and in the subsequent relative scale optimization.
The results of the ablation are shown in \cref{tab:diligent_ablation_connectivity}. Overall, no major differences emerge between the two connectivities, with comparable accuracies and runtimes across all objects. A minor exception is to be found when using normal similarity threshold $\theta_c = 5.0^{\circ}$, for which for some objects (\texttt{cat}, \texttt{harvest}, \texttt{reading}) $4-$ connectivity achieves better accuracy by a significant margin.

\section{Limitations~\label{sec_suppl:limitations}}
\begin{figure*}
    \centering
    \begin{subfigure}[b]{0.18\linewidth}
        \includegraphics[width=\linewidth]{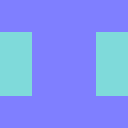}
        \caption{Input normal map}
    \end{subfigure}
    \hfill
    \begin{subfigure}[b]{0.18\linewidth}
        \includegraphics[width=\linewidth]{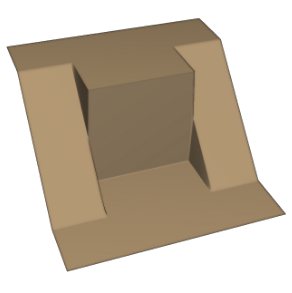}
        \caption{Ground-truth surface}
    \end{subfigure}
    \hfill
    \begin{subfigure}[b]{0.18\linewidth}
        \includegraphics[width=\linewidth]{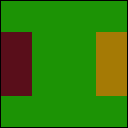}
        \caption{Components}
    \end{subfigure}
    \hfill
    \begin{subfigure}[b]{0.18\linewidth}
        \includegraphics[width=\linewidth]{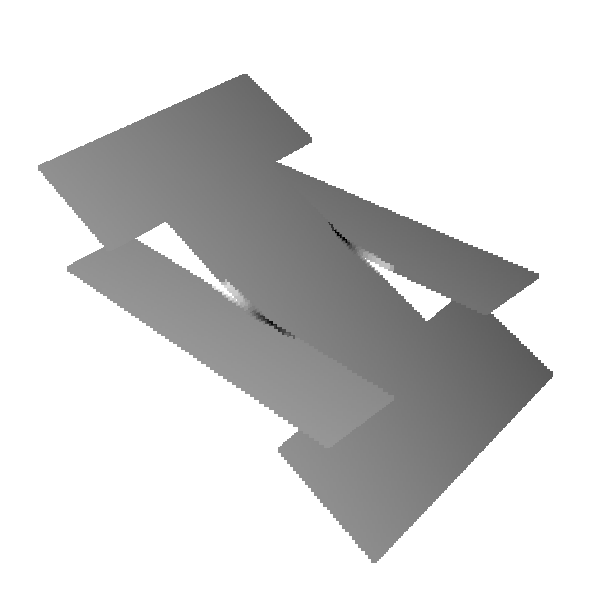}
        \caption{Our reconstruction}
    \end{subfigure}
    \caption{\textbf{Example corner case not handled by our component-formation heuristic.} When the input normals are continuous across a surface discontinuity, our heuristic for component formation cannot detect separate components. Since the model of discontinuity that we adopt~\cite{Cao2022BiNI} cannot recover discontinuities under the same corner case, the discontinuity will not be incorporated during the component filling and the two incorrectly merged pieces of surface will jointly adjust their scale in subsequent steps of the optimization. \textit{Source of the input normal map and ground-truth surface visualization}: \cite{Cao2022BiNI} (Fig.~14, Supplementary).}
    \label{fig:example_corner_case_heuristic}
\end{figure*}
\begin{figure*}[t]
    \centering
    \begin{subfigure}[b]{0.19\linewidth}
        \includegraphics[width=\linewidth]{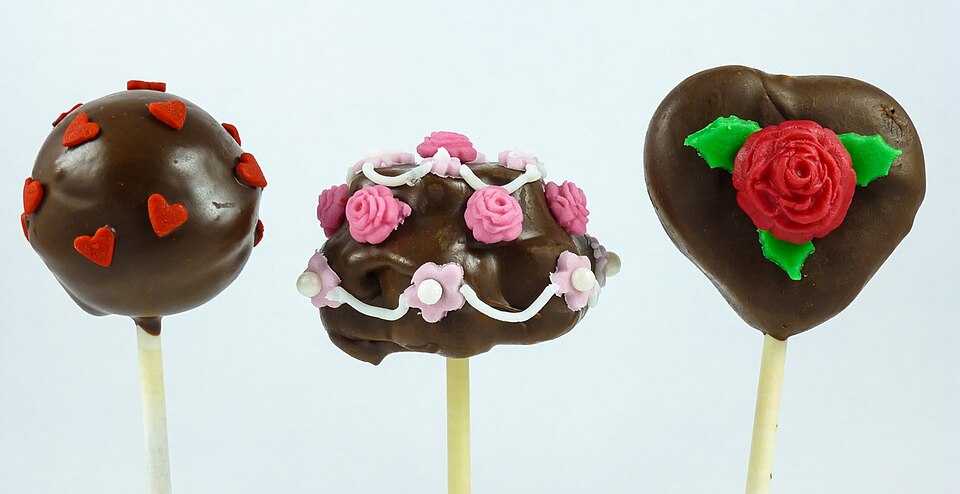}
        \caption{Input color image}
    \end{subfigure}
    \hfill
    \begin{subfigure}[b]{0.19\linewidth}
        \includegraphics[width=\linewidth]{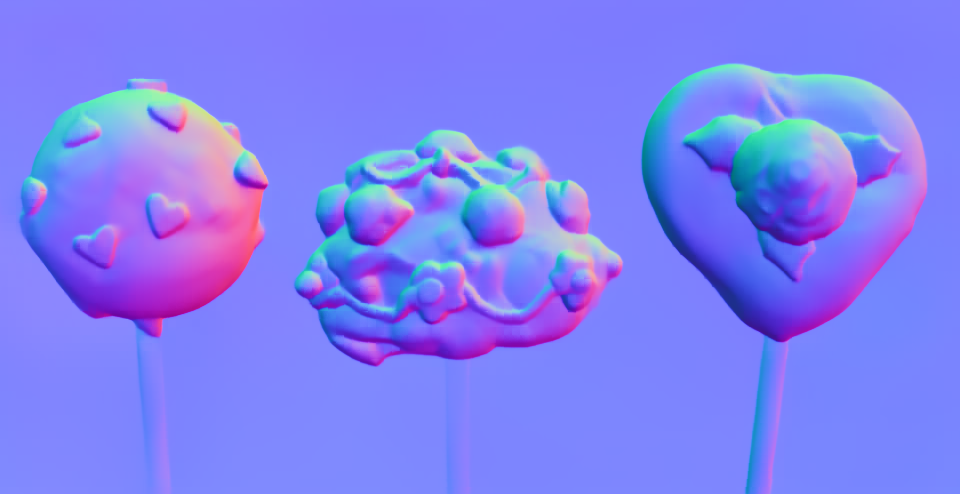}
        \caption{Normals from DSINE}
    \end{subfigure}
    \hfill
    \begin{subfigure}[b]{0.19\linewidth}
        \includegraphics[width=\linewidth]{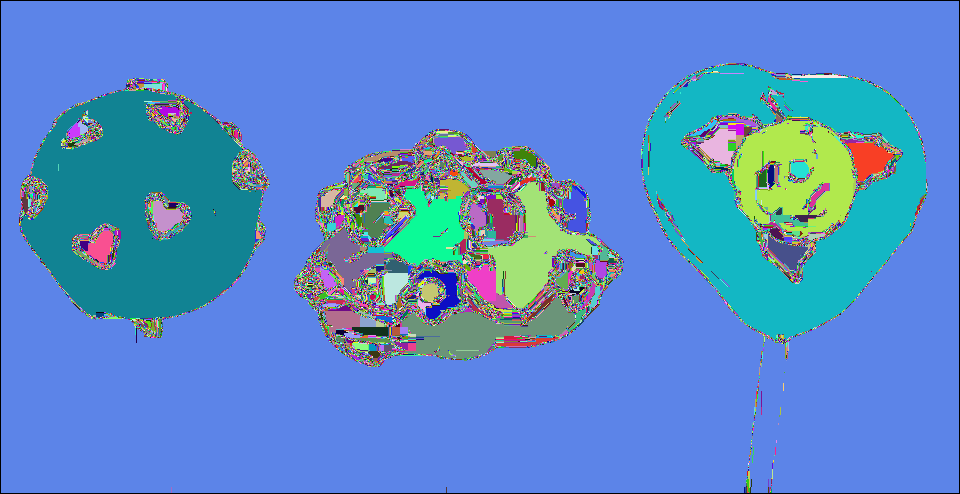}
        \caption{Components}
    \end{subfigure}
    \hfill
    \begin{subfigure}[b]{0.19\linewidth}
        \includegraphics[width=\linewidth]{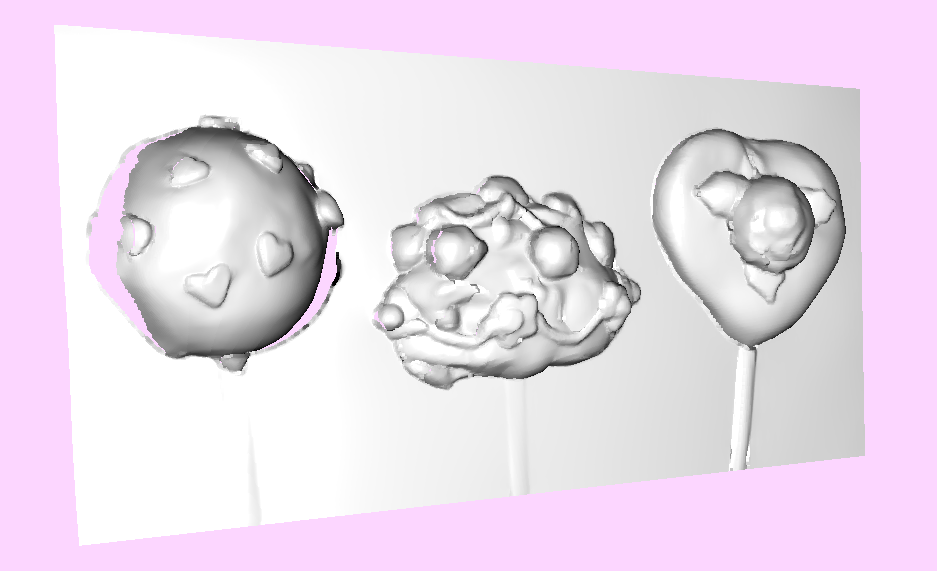}
        \caption{Our reconstruction}
    \end{subfigure}
    \caption{\textbf{Example limitation due to overly-smooth input normal map.} In the case of overly-smooth input normal maps, our heuristic for component formation cannot detect separate components across the non-sharp boundaries, as is the case in this example of the boundary between the sticks in the foreground and the background. As a consequence,
    the foreground object is partly merged into the background.
    \textit{Source of input color image}: \cite{ImageCakePops2013}.
    }
    \label{fig:example_limitation_overly_smooth_normals}
\end{figure*}
\textbf{Component formation and model of discontinuity.} Since our heuristic for component formation relies on the similarity between neighboring normals, if the normals are continuous across a surface discontinuity it cannot detect disconnected surface regions as separate components, as depicted in the example of \cref{fig:example_corner_case_heuristic}.
We note, however, that this
example represents
a corner case more in general
of normal integration methods, as previously noted for instance by~\cite{Cao2022BiNI} (Fig.~14, Supplementary).
In particular, in this case
the integration problem itself is ill-posed,
since in this setting it is not possible to determine
whether a discontinuity is present
from the normal map alone.

A similar issue arises
in the case of overly-smooth input normals, with indistinct boundaries, which
sometimes
occur
in
normal maps
produced as output
by learning-based approaches for normal estimation, such as DSINE~\cite{Bae2024DSINE}. For these normals maps, our heuristic for component formation 
might
sometimes
merge
multiple surfaces into a single component, depending on the similarity threshold $\theta_c$ (\cref{fig:example_limitation_overly_smooth_normals}).
We note, however, that
even
for
an incorrect component decomposition (\ie, that merges surface regions separated by a discontinuity)
the corresponding reconstruction can
in theory still
correctly capture the discontinuity if 
the model of discontinuity is 
able to
describe
it. This is because
in the initial phase each surface patch is
reconstructed
using
the same model of discontinuity
deployed for
relative scale optimization. Viceversa, if the model of discontinuity is unable to capture the discontinuity, as is the case for full discontinuities such as the foreground-background boundaries in \cref{fig:example_limitation_overly_smooth_normals}, the general problem of retrieving the scale of surface regions across such discontinuities cannot be addressed without additional knowledge.

An interesting future direction is to explore alternative, more sophisticated heuristics for component formation, for instance through the use of learning-based methods that could learn priors over the discontinuities from the normal maps.
In the more general case,
additional input or knowledge available depending on the setting might be incorporated in the component formation phase. For instance, if an input color image
was
provided, as is the case in
photometric stereo or surface normal estimation, edges might be extracted from the image and composed with the heuristic based only on surface normals.

\noindent\textbf{Decreased benefit for highly non-smooth normal maps.}
The computational advantage resulting from our method reducing the size of the optimization problem through the use of components is partially reduced when the input normal map is highly non-smooth. In particular, if the normal vectors vary often between neighboring pixels in an area of the input map, as for instance in the case
of
finely-textured regions, many single-pixel components may arise (\cf, \eg, the woven reed basked in the second column of \cref{fig:comparison_high_resolution_normal_maps}).
While our method can still be applied in such a setting,
its exact computational advantage will depend on the overall balance between smooth and non-smooth regions. 
An interesting future direction is to design
heuristics for component formation that could reduce the occurrence of such single-pixel components, for instance by detecting that highly-textured areas belong to the same connected surface region, through learned priors.
Additionally, we note that in many cases the number of components is greatly increased by single-pixel components that occur at the boundaries of
larger
components (\cf., \eg, $m=0$ and $m=1$ in \cref{fig:visualization_merging}).  
For such cases,
postprocessing of the initial
decomposition
could be explored, for instance by merging small components into larger ones without causing the larger ones to collapse into one another.

\end{document}